\documentclass[preprint,5p,times]{elsarticle}

\usepackage{booktabs}
\usepackage{natbib}
\usepackage{multirow}
\usepackage{amsfonts}
\usepackage{amssymb}
\usepackage[usenames,dvipsnames]{pstricks}
\usepackage{lscape}
\usepackage{rotating}
\usepackage{hyperref}
\usepackage{amsmath}
\usepackage{multirow}
\usepackage{color}
\usepackage{lscape}
\usepackage{colortbl}
\usepackage{url}
\usepackage{algorithm}
\usepackage{algorithmic}
\usepackage{lineno}
\usepackage{soul} \usepackage[applemac]{inputenc} 

\renewcommand{\blue}[1]{\textcolor [RGB]{0,0,0}{#1}}

\def\R{\ensuremath{\mathbf{R}}}

\def\W{\ensuremath{\mathbf{W}}}

\def\w{\ensuremath{\mathbf{w}}}

\def\bPhi{\ensuremath{\boldsymbol{\Phi}}}

\def\x{\ensuremath{\mathbf{x}}}
\newcommand{\bb}{\mathbf{b}}
\newcommand{\M}{\mathbf{M}}
\newcommand{\G}{\mathbf{G}}

\newcommand{\mytitle}{Multiclass feature learning for hyperspectral image classification: sparse and hierarchical solutions.}

\journal{ISPRS Journal of Photogrammetry and Remote Sensing	}

\begin{document}

\begin{frontmatter}

\title{\mytitle}
\author[1]{Devis~Tuia}
\cortext[MVgrant]{Corresponding Author: devis.tuia@geo.unizh.ch}
\author[2]{R\'emi Flamary}
\author[3]{Nicolas Courty}

\address[1]{Department of Geography, University of Zurich, Switzerland}
\address[2]{Laboratoire Lagrange UMR CNRS 7293, OCA, Universit\'e de Nice Sofia Antipolis, France}
\address[3]{Universit\'e de Bretagne Sud/IRISA, France}

\linenumbers

\begin{abstract}
In this paper, we tackle the question of discovering an effective set
of spatial filters to solve hyperspectral classification problems.
Instead of fixing \emph{a priori } the filters and their
 parameters using expert knowledge, we let the model find them within random draws in the (possibly infinite) space of possible filters.
 We define an
active set feature learner that includes in the model only features that improve the classifier. To this end, we  consider a fast and
linear classifier, multiclass logistic classification, and show that
with a good representation (the filters discovered), such a simple
classifier can reach at least state of the art performances. We apply the proposed active set learner in four
hyperspectral image classification problems, including agricultural and urban classification
at different resolutions, as well as multimodal data. We also propose
a hierarchical setting, which allows to generate more complex banks of
features that can better describe the nonlinearities present in the
data. \end{abstract}


\end{frontmatter}


\section{Introduction}
Hyperspectral remote sensing allows to obtain a fine description of the materials observed by the sensor: with arrays of sensors focusing on 5-10 nm sections of the electromagnetic spectrum, hyperspectral images (HSI) return a complete description of the response of the surfaces, generally in the visible and infrared range. The use of such data, generally acquired by sensors onboard satellites or aircrafts, allows to monitor the processes occurring at the surface in a non-intrusive way, both at the local and global scale~\citep{Lil08,Ric06}. The reduced revisit time of satellites, in conjunction with the potential for quick deployment of aerial and unmanned systems, makes the usage of hyperspectral systems quite appealing. As a consequence, hyperspectral data is becoming more and more prominent for researchers and public bodies.

Even if the technology is at hand and images can be acquired by
different platforms in a very efficient way, HSI alone are of little
use for end-users and decision makers: in order to be usable, remote
sensing pixel information must be processed and converted into maps
representing a particular facet of the processes occurring at the
surface. Among the different products traditionally available, land
cover maps issued from image classification are the most common (and
probably also the most used). In this paper, we refer to land
cover/use classification as the process of attributing a land cover
(respectively land use) class to every pixel in the image. These maps
can then be used for urban planning~\citep{Tau12,Tau13}, agriculture
surveys~\citep{Alc12} or surveying of
deforestation~\citep{Asn05,Nai12,Lau14}.

The quality of land cover maps is of prime importance. Therefore, a wide panel of research works consider image classification algorithms and their impact on the final maps~\citep{Pla08,Cam11,Mou11,Cam13}. Improving the quality of maps issued from HSI is not trivial, as hyperspectral systems are often high dimensional (number of spectral bands acquired), spatially and spectrally correlated and affected by noise~\citep{Cam13}. 
Among these peculiarities of remote sensing data, spatial relations
among pixels have received particular attention~\citep{Fauvel13}: the
land cover maps are  generally smooth, in the sense that neighboring
pixels tend to belong to the same type of land cover~\citep{Sch12}. On
the contrary, the spectral signatures of pixels of a same type of
cover tend to become more and more variable, especially with the
increase of spatial resolution. Therefore, HSI classification systems
have the delicate task of  describing a smooth land cover using
spectral information with a high within-class variability. Solutions
to this problem have been proposed in the community and mostly recur
to spatial filtering that work at the
level of the input vector~\citep{Ben05,Vai06,Fauvel13}  or to structured models that work by optimization of a context-aware energy function~\citep{Tar10b,Sch12,Mos13}.

In this paper, we start from the first family of methods, those
based 
on the extraction of spatial filters prior to classification. Methods proposed in remote sensing image classification tend to pre-compute a large quantity of spatial filters related to the user's preference and knowledge of the problem: texture~\citep{Pac08}, Gabor~\citep{Li14}, morphological~\citep{Ben05,Mur10} or bilateral filters~\citep{Sch12} are among those used in recent literature and we will use them as buiding blocks for our system. With this static and overcomplete set of filters (or  \emph{filterbank}), a classifier is generally trained.

\begin{figure}
\centering
\includegraphics[width=6cm]{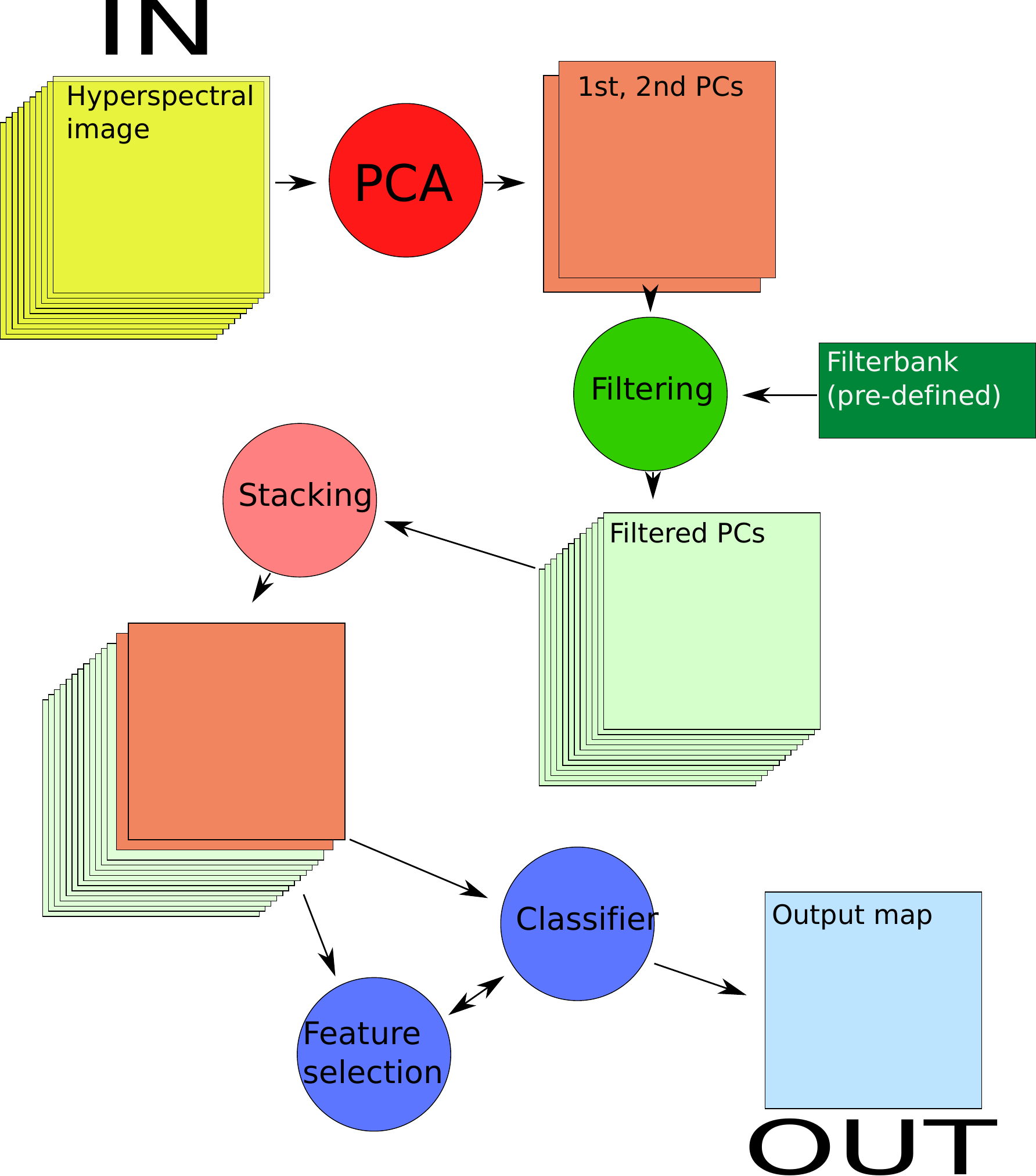}
\caption{Traditional spatio-spectral classification with contextual filters: using pre-defined filterbanks, applied on the first principal component.}
\label{fig:trad}
\end{figure}

Even if successful, these studies still rely on the definition  $a$-priori
of a filterbank. This filterbank depends on the knowledge of the
analyst \blue{ and on the specificities of the image at hand: a pre-defined filterbank} may or may not contain the filters 
leading to the best performances.
A filterbank constructed $a$-priori is also often redundant: as shown in Fig.~\ref{fig:trad}, the filter bank
is generally applied to each band of the image, resulting into a $(f \times
B)$-dimensional filter bank, where $f$ is the number of filters and
$B$ the number of bands. Proceeding this way proved in the past to be unfeasible for high
dimensional datasets, such as hyperspectral data, for which the traditional
way to deal with the problem is to perform a principal components analysis (PCA) and then extract the filters from the $p << B$ principal components related to maximal
variance~\citep{Ben05}. In that case, the final input space becomes $(f \times p)$-dimensional. \blue{ A first problem is related during this dimension reduction phase, for which the choice of the feature extractor and of the number of features $p$ remains arbitrary and may lead to discarding information that is discriminative, but not related to large variance. Therefore, a first objective of our method is to avoid this first data reduction step.}  But independently to the reduction phase, this goes against the desirable property of a model to be compact, i.e., to depend on as few input variables as possible.
Therefore, in most works cited above an additional feature selection step is run to select the most effective subset for classification. This additional step can be a recursive selection~\citep{Tui09b} or be based on kernel combination~\citep{Tui10c}, on the pruning of a neural network~\citep{Pac08} or on discriminative feature extraction~\citep{Ben05}. 

Proceeding this way is suboptimal in two senses: first, one forces to
restrict the number and parameters of filters to be used to a subset,
whose appropriateness only depends on the prior knowledge of the
user. In other words, the features that are relevant to solve the classification problem might not be in the original filterbank. Second, generating  thousands of spatial
filters and use them all together in a classifier, that also might operate with a feature selection strategy, increases the computational cost significantly, and might even deteriorate the classification accuracy because of the curse of dimensionality. Note that, if the spatial filters considered bear continuous parameters (e.g. Gabor or angular features), there is theoretically an infinite number of feature candidates. 

This paper tackles these two problems simultaneously: instead of pre-computing a specific set of filters, we propose to interact with the current model and retrieve only new filters that will make it better. These candidate filters can be of any nature and with parameters unrestricted, thus allowing to explore the (potentially infinite) space of spatial filters. This leads to an integrated approach, where we incrementally build the set of filters from an empty subset and add only the filters improving class discrimination. \blue{ This way of proceeding is of great interest for automatic HSI classification, since the filters are selected automatically among a very large set of possible ones, and are those that best fit the problem at hand.}

Two approaches explored similar concepts in the past: Grafting~\citep{Per03}
and Group Feature Learning~\citep{Rak12}, which incrementally select
the most promising feature among a batch of features extracted from
the universe of all possible features admitted. Since this selection
is based on a heuristic criterion ranking the features by their
informativeness when added to the model, it may be seen as performing
active learning~\citep{Cra12} in the space of possible features (in
this case, the active learning oracle is replaced by the optimality
condition, for which only the features improving the current
classifier are selected).

\begin{figure*}
\begin{tabular}{c|c}
\includegraphics[width=.5\linewidth]{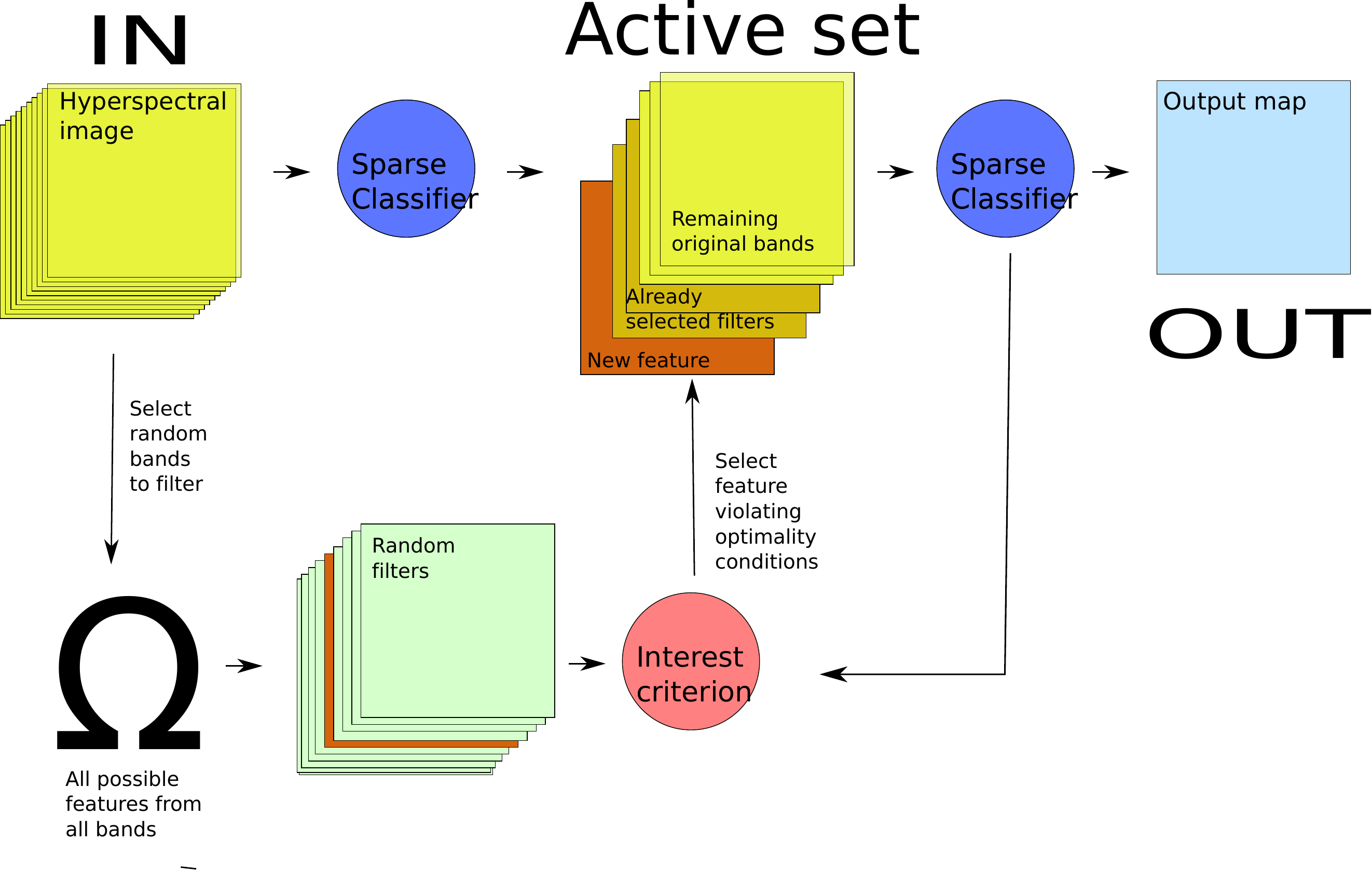}&\includegraphics[width=.5\linewidth]{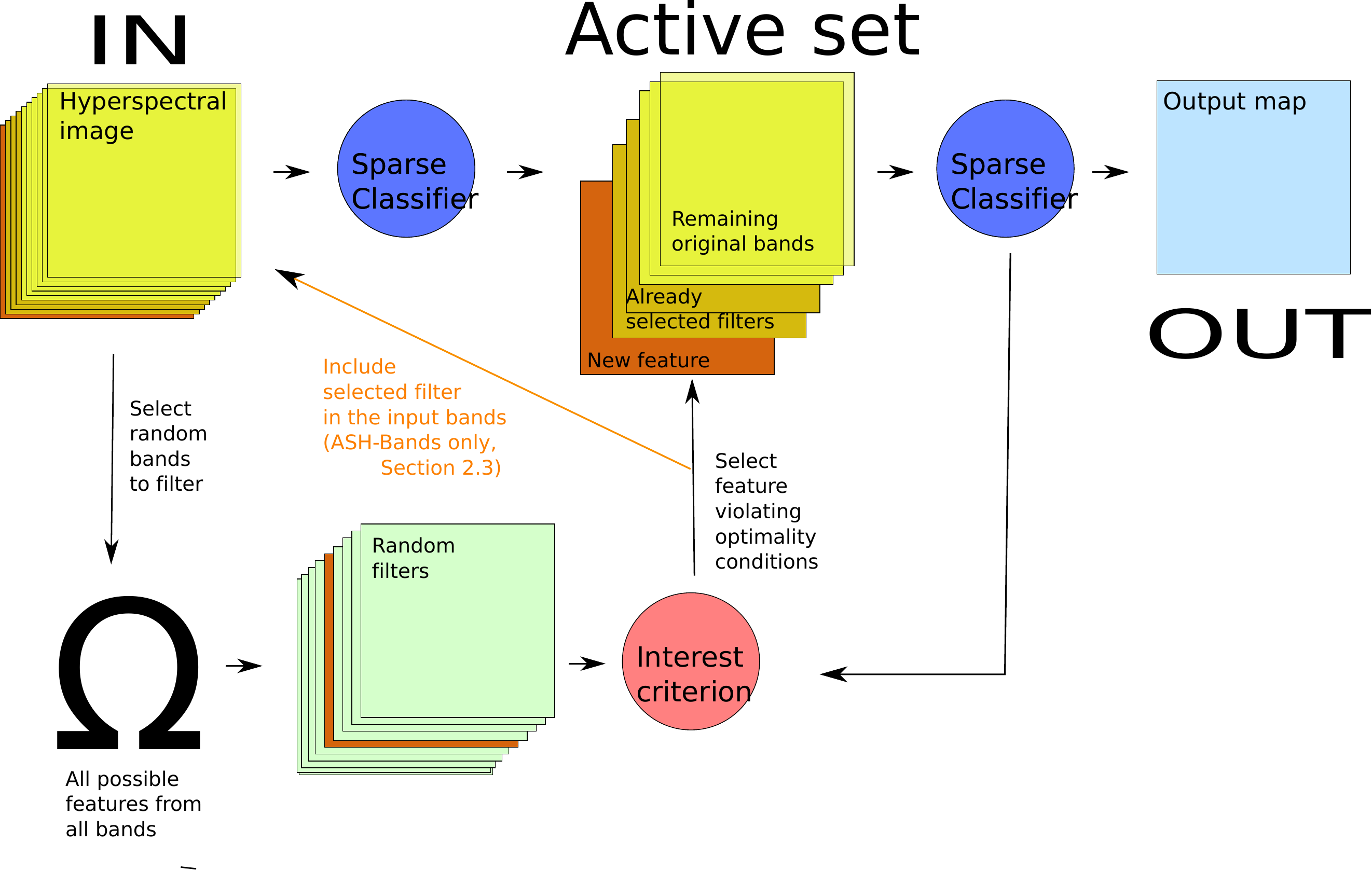} \\
(a)  Proposed \textsc{AS-Bands}& (b) Proposed \textsc{ASH-Bands}\\
\end{tabular}
\caption{Spatio-spectral classification with the proposed active set models. (a) With only the original HSI image as bands input (shallow model, \textsc{AS-Bands}); (b) with the hierarchical feature extraction (deep model, \textsc{ASH-bands}).}
\label{fig:as}
\end{figure*}

In this paper, we propose a new Group Feature Learning model 
based on multiclass logistic regression (also known as multinomial
  regression). The use of a group-lasso regularization~\citep{Yua07} allows to
  jointly select the relevant features and also to derive efficient
  conditions for evaluating the discriminative power of a new
  feature. In~\citet{Rak12}, 
authors propose to use group-lasso for multitask learning by allowing to use an additional sparse average
classifier common to all tasks. Adapting their model in
a multiclass classification setting leads to the use of the sole group-lasso
regularization. Note that one could use a  $\ell_1$ support
vector machine as in \citet{Tui13c} to select the relevant feature in a
One-VS-All setting, but this approach is particularly computationally
intensive, as the incremental problem is solved for each class separately. This
implies the generation of millions of features, that may be useful for
more than one class at a time. To achieve an efficient multiclass strategy, we propose the following three original contributions:

\begin{enumerate}
\item We use here a multiclass logistic classifier (MLC) with a
  softmax loss. MLC allows to natively
  handle several classes without using the One-VS-All approach and has
  the advantage of providing probabilistic prediction scores that can
  more easily be used in structured models (such as Markov
  random fields).

\item We employ a group lasso regularization, which allows to select
  features useful for many classes simultaneously, even if they do not
  show the highest score for a single class.   This means sharing
  information among the classes, similarly to what would happen in a
  multitask setting~\citep{Lei13}. This model, called \textsc{AS-Bands}, is detailed in Fig.~\ref{fig:as}(a).

\item We investigate the automatic selection of complex hierarchical
  spatial filters built as modifications of previously selected
  filters. This leads to a tree- (or graph-) based feature extraction that can
   encode complex nonlinear relationship for each class. Such a hierarchical re-processing of features has connections with deep neural networks~\citep{Lec89,Lec98}, which have recently proven to be able to improve significantly the performance of existing classification methods in computer vision~\citep{Cha14,Gri14}. This model, called \textsc{ASH-bands}, is detailed in Fig.~\ref{fig:as}(b).

\end{enumerate}

We test the proposed method on two landcover classification tasks with
hyperspectral images of agricultural areas and on one landuse
classification example over  an urban area exploiting jointly
hyperspectral and LiDAR images. In all cases, the proposed feature
learning method solves the classification tasks with at least state of
the art numerical performances and  returns compact models including
only features that are discriminative for more than one
class. Among the two method proposed, the hierarchical feature
learning tends to outperform the shallow feature extractor for
traditional classification problems. However, when confronted to
shifting distributions between train and test (i.e. a domain
adaptation problem), it provides slightly worse performances, probably
due to the complexification of the selected features, that overfit the
training examples.

The remainder of this paper is as follows: Section~\ref{sec:method} details the proposed method, as well as the multiclass feature selection using group-lasso. In Section~\ref{sec:data} we present the datasets and the experimental setup. In Section~\ref{sec:res} we present and discuss the experimental results. Section~\ref{sec:conc} concludes the paper.

\section{Multiclass active set feature discovery}
\label{sec:method}

In this section, we first
present the multiclass logistic classification and then derive its
optimality conditions, which are used in the active set algorithm.

\subsection{Multiclass logistic classifier with group-lasso
  regularization}

Consider an image composed of pixels $\x_i \in \mathbb{R}^B$. A subset
of $l_c$ pixels is labeled into one of $C$ classes:
$\{\x_i,y_i\}_{i=1}^{l_c}$, where $y_i$ are integer values $\in \{1,
\dots, C\}$. We consider a (possibly infinite) set of
$\theta$-parametrized functions $\phi_\theta(\cdot)$ mapping each
pixel in the image into the feature space of the filter defined by
$\theta$. As in~\citet{Tui13c}, we define as $\mathcal{F}$ the set of
all possible finite subsets of features and $\varphi$ as an element of
$\mathcal{F}$ composed of $d$ features $\varphi =
\{\phi_{\theta_j}\}_{j=1}^d$. We also define $\Phi_\varphi(\x_i)$ as
the stacked vector of all the values obained by applying the filters
$\varphi$ to pixel  $\x_i$ and
  $\bPhi_\varphi\in\mathbb{R}^{l_c\times d}$ the matrix containing the
 $d$ features in $\varphi$ computed for all the $l_c$ labeled pixels. Note that
  in this work, we suppose that all the features have been normalized
  with each column in matrix $\bPhi_\varphi$ having a unit norm. 

In this paper we consider the classification problem as a multiclass logistic regression problem with group-lasso
  regularization. Learning such a classifier for a fixed
amount of features $\varphi$ corresponds to learning a weight matrix
$\W \in \mathbb{R}^{d \times C}$ and the bias vector $\bb \in
\mathbb{R}^{1 \times C}$ using the softmax loss. In the
following, we refer to $\w_c$ as the weights corresponding to class
$c$, which corresponds to the $c$-th  column of  matrix $\W$. The $k$-th line of matrix
$\W$ is denoted as $W_{k,\cdot}$. The optimization problem for a fixed
feature set $\varphi$ is defined as:
{\small
  \begin{align}
 \min_{\W, \bb}\quad \mathcal{L}(\W, \bb) =\left\{  \frac{1}{l_c}\sum_{i=1}^{l_c}H(y_i,\x_i,\W,\bb)  + \lambda  \Omega(\W)\right\}
    \label{eq:MLC}
\end{align}}
where the first term corresponds to the soft-max loss with $H(\cdots)$
defined as
\begin{equation}
  \label{eq:2}
 H(\cdots)= \log \left( \sum_{c=1}^C \exp\left((\w_c - \w_{y_i})^\top\Phi_\varphi(\x_i) \nonumber+(b_c-b_{y_i}) \right)\right)
\end{equation}
 and the second 
term is a group-lasso regularizer. In this paper, we use the weighted
$\ell_1\ell_2$ mixed norm :
\begin{equation}
 \Omega(\W) = \sum_{j=1}^d\gamma_j ||W_{j,\cdot}||_2
\label{eq:grouplasso}
\end{equation}
where the coefficients $\gamma_j>0$ correspond to the weights used
  for regularizing the $j$th feature. Typically one want all features
  to be regularized similarly by choosing $\gamma_j=1,\;
  \forall j$. However, in the hierarchical feature extraction proposed in  Section~\ref{sec:deep} we will use different weights in order to limit over-fitting when using complex hierarchical features.

This regularization term promotes group sparsity, due to its non
  differentiability at the null vector of each group. In this case we
grouped the coefficients of $\W$ by lines, meaning that the regularization will promote
joint feature selection for all classes. Note that this approach can
be seen as multi-task learning where the tasks corresponds to the
classifier weights of each class 
\citep{obozinski2006multi,rakotomamonjy2011penalty}. As a result, if a
variable (filter) is active, it will be active for all classes.
This is particularly interesting in in a multiclass setting, since a
feature that helps in detecting a given class also helps in ``not
detecting'' the others $C-1$ classes: for this reason a selected feature should be
active for all the classifiers.

The algorithm proposed to solve both the learning problem and
  feature selection is derived from the optimality conditions
  of the optimization problem \blue{ of  Eq.~}\eqref{eq:MLC}. Since the problem
  defined in \blue{ Eq.~}\eqref{eq:MLC} is non-differentiable, we compute the
  sub-differential of its cost function:
\begin{equation}
\partial_\W\mathcal{L}(\W,\bb) = \bPhi_\varphi^\top \R +  \lambda \partial\Omega(\W)
\label{eq:grad}
\end{equation}
where the first term corresponds to the gradient of the softmax
  data fitting and the second term is the sub-differential of the
  weighted group lasso defined in Eq.~\eqref{eq:grouplasso}.
$\R$ is a $l_c \times C$ matrix that, for a given sample $i \in \{1,.,l_c\}$ and a class $c \in \{1,.,C\} $, equals:
\begin{equation}
R_{i,c} =  \frac{\exp(M_{i,c} - M_{i,y_i}) - \delta_{\{y_i-c\}}\sum_{k=1}^C\exp(M_{i,k} - M_{i,y_i})}{l_c\sum_{k=1}^C \exp(M_{i,k} - M_{i,y_i})}
\label{eq:grad}
\end{equation}
where $\M = \bPhi_{\varphi}\W + \mathbf{1}_{l_c} \bb$ and $\delta_{\{y_i-c\}}=
1$ if $c = y_i$ and 0 otherwise. In the following, we define $\G =
\bPhi_\varphi^\top \R$ as a $d\times C$ matrix corresponding to the
gradient of the data fitting term \emph{w.r.t} $\W$. Note that this
gradient can be computed efficiently with multiple scalar product between the
features $\bPhi_\varphi$ and the multiclass residual $\R$. The
optimality conditions can be obtained separately for each $W_{j,\cdot}$,
\emph{i.e.} for each line $j$ of the $\W$ matrix. $\Omega(\W)$
consists in a weighted sum of
non differentiable norm-based regularization \citep{Bach11}.  The
optimality condition for the $\ell_2$ norm consists in
a constraint with its dual norm (namely itself): 
\begin{equation}
||G_{j,\cdot}||_2 \leq \lambda\gamma_j \qquad \forall j \in \varphi
\end{equation}
which in turn breaks down to:
\begin{equation}
\left\{\begin{array}{lll}
||G_{j,\cdot}||_2 = \lambda\gamma_j & \text{if} & W_{j,\cdot} \neq \boldsymbol{0}\\
||G_{j,\cdot}||_2 \leq \lambda\gamma_j & \text{if} & W_{j,\cdot} = \boldsymbol{0}
\label{eq:optinvarphi_w1}
\end{array}
\right.
\end{equation}
These optimality conditions show that the selection
of one variable, \emph{i.e.} one group, can be easily tested with the
second condition of equation \eqref{eq:optinvarphi_w1}. This 
suggests the use of an active set
algorithm. Indeed, if the norm of  correlation of a feature with the
residual matrix is below $\lambda\gamma_j$, it means that this feature is not
useful for classification and its weight will be set to $0$ for all the classes. On the contrary, if
  not, then the group can be defined as ``active'' and its weights
  have to be estimated.

\subsection{Proposed active set criterion (\textsc{AS-bands})}

We want to learn jointly the best set of filters
$\varphi^* \in \mathcal{F}$ and the corresponding MLC classifier. This
is achieved by minimizing Eq.~\eqref{eq:MLC} jointly  on $\varphi$ and
$\W,\mathbf{b}$. As in~\citet{Rak12}, we can extend the optimality conditions in
\eqref{eq:optinvarphi_w1} to all filters with zero weights that are
\emph{not} included in the
current active set $\varphi$:
\begin{equation}
||G_{{\phi_\theta},\cdot}||_2 \leq \lambda\gamma_{\phi_\theta} \qquad \forall \phi_\theta \notin \varphi
\label{eq:optnotinvarphi}
\end{equation}
Indeed, if this constraint holds for a given feature not in the
current active set, then adding this feature to the optimization
problem will lead to a row of zero weights $W_{(d+1),\cdot}$ for this feature.
But this also means that
if we find a feature that violates Eq.~\eqref{eq:optnotinvarphi}, its 
inclusion in $\varphi$ will (after re-optimization)  make
the global MLC cost decrease and provide a feature with non-zero
coefficients for all classes. 

The pseudocode of the proposed algorithm is given in Algorithm~\ref{alg}: we  initialize the active
set $\varphi_0$ with the spectral bands and run a first MLC minimizing
Eq.~\eqref{eq:MLC}. Then we generate a random minibatch of candidate
features, $\Phi_{\theta_j}$, involving spatial filters with random
types and parameters. We then assess the optimality conditions
with~\eqref{eq:optnotinvarphi}: if the feature $\phi_{\theta_j}^*$ with maximal
$||G_{\theta_j,\cdot}||_2$ is greater than $\lambda\gamma_j + \epsilon$,
it  is selected and added to the current
active set $[\phi_{\theta_j}^* \cup \varphi ]$. After one feature is added the MLC classifier is retrained and the process is iterated using the new active set.

\begin{algorithm}[t]
\caption{Multiclass active set selection for MLC (\textsc{AS-Bands})}
\label{alg}
\textbf{Inputs}\\
-  Bands to extract the filters from ($B$)\\
-  Initial active set $\varphi_0 = B$\\
\vspace{-0.4cm}
	\begin{algorithmic}[1] 
	\REPEAT
		\STATE Solve a MLC with current active set $\varphi$
		\STATE Generate a minibatch   $\{\phi_{\theta_j}\}_{j=1}^p \notin \varphi$

		\STATE Compute $G$ as in \eqref{eq:optnotinvarphi}
                $\forall j=1\dots p$
		\STATE Find feature $\phi_{\theta_j}^*$
                maximizing $||G_{\theta_j,\cdot}||_2$
                \IF{$||G_{\theta_j^*,\cdot}||_2 > \lambda\gamma_i+\epsilon$}
 \STATE $\varphi~= \phi_{\theta_j}^*\cup \varphi$
                \ENDIF
	\UNTIL{stopping criterion is met}
	\end{algorithmic}

\end{algorithm}

\subsection{Hierarchical feature learning (\textsc{ASH-bands})}\label{sec:deep}

Algorithm \ref{alg} searches randomly in a possibly
  infinite dimensional space corresponding to all the possible spatial
  filters computed on the input bands. But despite all their differences,  the spatial filters proposed in the remote sensing community (see, as an example, those in Tab.~\ref{tab:filt}) can yield only a limited complexity and
non-linearity. When the classes are not linearly separable, learning
a linear classifier may require  a large number of these relatively
simple features. In this section we investigate the use of hierarchical
feature generation that can yield much more complex data
representation and therefore hopefully decrease the number of features necessary
 for a good classification.

Hierarchical feature extraction is obtained by adding the already
selected features in the pool of images that can be used for filtering at the
next feature generation step. Using a retained filter as a new possible input band leads to more complex filters with higher nonlinearity. This is  somehow related to the methods of deep learning, where deep features are generally obtained by aggregation of convolution operators. In our case, those operators are substituted by spatial filters with known properties, \blue{ which adds up to our approach the appealing property of direct interpretability of the discovered features. In deep learning models, interpretation of the features learned is becoming possible, but at the price of series of deconvolutions~\citep{Zei14}}.
Let $h_j\in\mathbb{N}$ be the depth of a
given feature $\phi_{\theta_j}$, \blue{ with 0 being the depth of original features}: this is the number of filtering steps the original bands has undergone to generate filter $\phi_{\theta_j}$. For example, the band 5 has depth $h_5 = 0$, while the filters that are issued from this band, for example a filter $k$ issued from an opening computed on band 5, will have depth $h_k = 1$. If the opening band is then re-filtered by a texture filter into a new filter $l$, its depth will be $h_l = 2$. 
This leads to a much more complex feature extraction that
{builds upon an hierarchical, tree-shaped, suite of filters}.  
The depth of the feature in the feature generation {tree} is of importance in our case since it is a good proxy
of the complexity of the features. In order to avoid over-fitting, we propose to
regularize the features using their depth in the hierarchy. As a criterion, we use a
regularization weight of the form $\gamma_j=\gamma_0^{h_j}$, with
$\gamma_0\geq1$ being a term penalizing depth in the graph.

The proposed hierarchical feature learning is summarized in
  Algorithm \ref{alg2}.

\begin{algorithm}[t]
\caption{Multiclass active set selection for MLC, hierarchical deep setting (\textsc{ASH-Bands})}
\label{alg2}
\textbf{Inputs}\\
-  Bands to extract the filters from ($B$) with depth $h = 1$\\
-  Initial active set $\varphi_0 = B$\\
\vspace{-0.4cm}
	\begin{algorithmic}[1] 
	\REPEAT
		\STATE Solve a MLC with current active set $\varphi$
		\STATE Generate a minibatch
                $\{\phi_{\theta_j},h_j\}_{j=1}^p \notin \varphi$ using {$B$}                                 as input for filters

                \STATE Compute depth-dependent regularizations as 

\hspace{5mm}$\gamma_j=\gamma_0^{h_j}$

		\STATE Compute $G$ as in \eqref{eq:optnotinvarphi}
                $\forall j=[1\dots p]$
                
                \STATE Compute  optimality conditions violations as

 $\Lambda_j = ||G_{\theta_j,\cdot}||_2 - \lambda\gamma_j-\epsilon$, $\forall j=[1\dots p]$
                
		\STATE Find feature $\phi_{\theta_j}^*$
                maximizing $\Lambda_j$ 
                \IF{$\Lambda_{\theta_j^*} > 0$}
  \STATE $\varphi~= \phi_{\theta_j}^*\cup \varphi$
 \STATE {$B = \phi_{\theta_j}^* \cup B$}
                      \ENDIF
	\UNTIL{stopping criterion is met}
	\end{algorithmic}

\end{algorithm}

\section{Data and setup of experiments}\label{sec:data}
In this section, we present the three datasets used, as well as the setup of the four experiments considered.
\subsection{Datasets}
We studied the proposed active set method on four hyperspectral classification tasks, involving two crops identification datasets and one urban land use dataset (considered in two ways):

\begin{itemize}
\item[a)] Indian Pines 1992 (AVIRIS spectrometer, HS): the first dataset is a 20-m resolution image taken over the Indian Pines (IN) test site in June 1992 (see Fig.~\ref{fig:indpine92}). The image is $145 \times 145$ pixels and contains 220 spectral bands. A ground survey of 10366 pixels, distributed in  16 crop types classes, is available (see Table~\ref{tab:AVIRIS}).
This dataset is a classical benchmark to validate model accuracy. Its challenge resides in the strong
mixture of the classes' signatures, since the image has been acquired
shortly after the crops were planted. As a consequence, all signatures
are contaminated by soil signature, making thus a spectral-spatial
processing compulsory to solve the classification problem.  
As preprocessing, 20 noisy bands covering the region of water
absorption have been removed.

 \begin{table}[!t]
 \caption{Classes and samples ($n_l^c$) of the ground truth of the Indian Pines 1992 dataset (cf. Fig.~\ref{fig:indpine92}).}
 \centering
 \small{
 \setlength{\tabcolsep}{4pt}
\begin{tabular} {c|p{2.45cm}|p{.5cm}||c|p{2cm}|p{.6cm} }
		\toprule
 \hline
& Class & $n_l^c$ &&Class & $n_l^c$  \\
 \hline
 \hline
 \cellcolor[rgb]{1,0.5,0.5} &Alfalfa &54 &\cellcolor[rgb]{1,0,1}& Oats &20 \\
\cellcolor[rgb]{0,1,0}  &Corn-notill &1434 &\cellcolor[rgb]{1,0.5,0}& Soybeans-notill &968\\
\cellcolor[rgb]{0,0,1} &Corn-min & 834 &\cellcolor[rgb]{0,0.8,1}& Soybeans-min &2468\\
\cellcolor[rgb]{0,0.5,0.5} &Corn &234&\cellcolor[rgb]{0.2,0.5,0.5} &Soybeans-clean &614\\
\cellcolor[rgb]{1,0,0} &Grass/Pasture &497&\cellcolor[rgb]{0.4,0.7,0.1} &Wheat &212\\
\cellcolor[rgb]{0.2,0.2,0.9}  &Grass/Trees &747&\cellcolor[rgb]{1,1,1}&Woods &1294\\\
\cellcolor[rgb]{0.3,0,0.1} &Grass/Past.-mowed &26&\cellcolor[rgb]{1,1,0}&Towers &95\\
 &Hay-windrowed &489&\cellcolor[rgb]{0.8,0.7,0}&Other &380\\
 \hline
 \multicolumn{4}{c}{}&Total & 10366\\

 \bottomrule
 \end{tabular}}
  \label{tab:AVIRIS}
 \end{table}

\begin{figure}[t!]
	\centering
	\begin{tabular}{cc}
			\includegraphics[width=3.91cm]{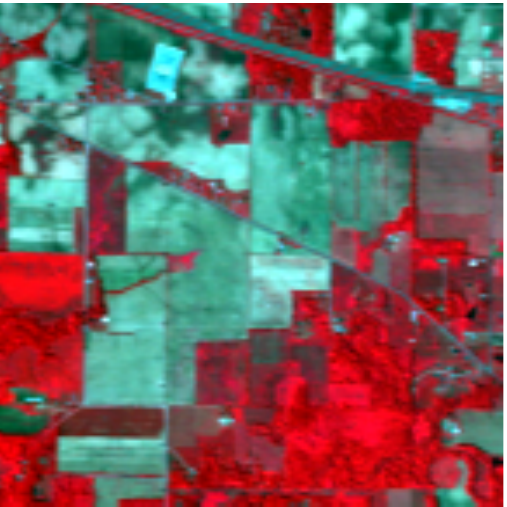}&				\includegraphics[width=3.91cm]{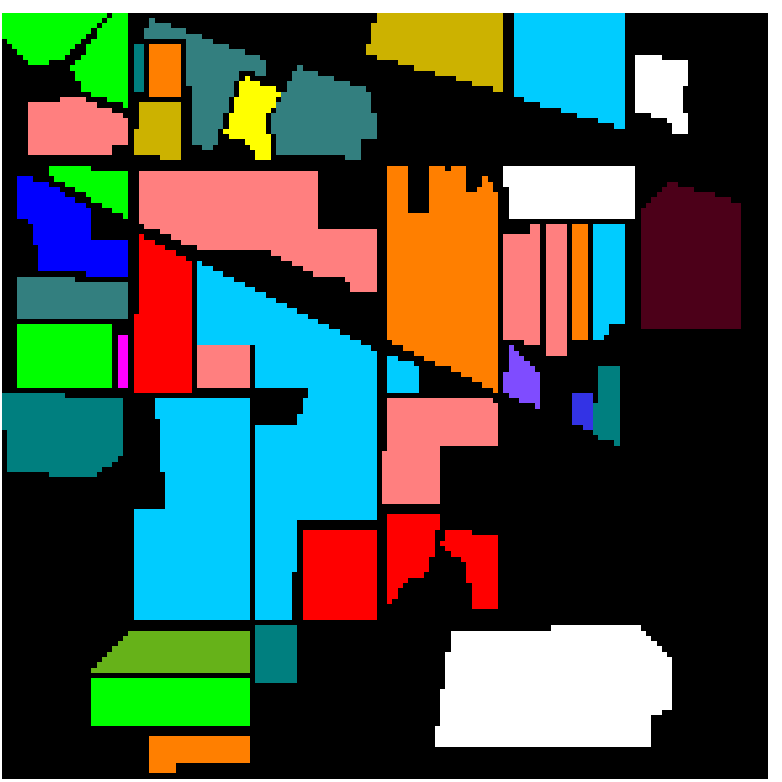}\\
			(a) & (b) \\
			\end{tabular}
	\caption{Indian Pines 1992 AVIRIS data.(a) False color composition and (b) ground truth (for color legend, see Tab.~\ref{tab:AVIRIS}). Unlabeled samples are in black.
		\label{fig:indpine92}
	}
	\end{figure}

\item[b)] Indian Pines 2010 (ProSpecTIR spectrometer, VHR HS): the second dataset considers  multiple flightlines acquired near Purdue University, Indiana, on May 24-25, 2010 by the  ProSpecTIR system  (Fig.~\ref{fig:indpine}). The image subset analyzed in this study contains 445$\times$750 pixels at 2\textit{m} spatial resolution, with 360 spectral bands of 5{\it nm} width. Sixteen land cover classes were identified by field surveys, which included fields of different crop residue, vegetated areas, and man-made structures. Many classes have regular geometry associated with fields, while others are related with roads and isolated man-made structures. Table~\ref{tab:indianPine} shows class labels and number of training samples per class.  

\begin{figure}[t!]
	\centering
	\begin{tabular}{cc}
			\includegraphics[width=.455\linewidth]{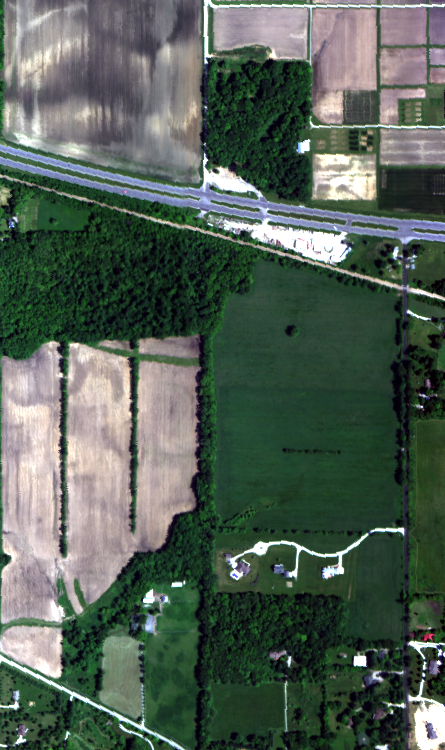}&	
			\includegraphics[width=.46\linewidth]{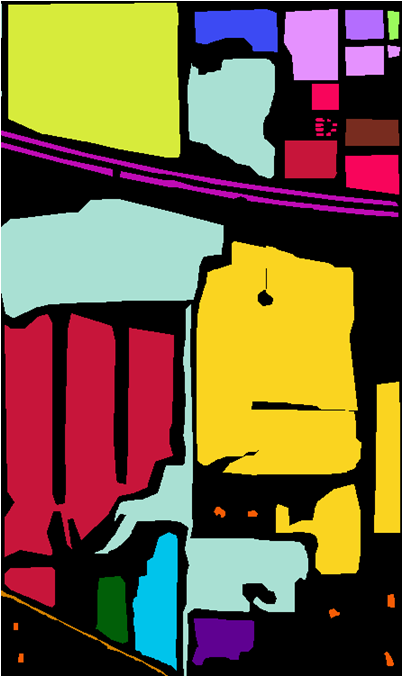}\\
 			(a) & (b)\\
	\end{tabular}
	\caption{Indian Pines 2010 SpecTIR data.(a) RGB composition and (b) ground truth (for color legend, see Tab.~\ref{tab:indianPine}). Unlabeled samples are in black.}
	\label{fig:indpine}
\end{figure}
\begin{table}[t!]
	\centering
	\caption{Classes and samples ($n_l^c$) of the ground truth of the Indian Pines 2010 dataset (cf. Fig.~\ref{fig:indpine}).}
	{\small
	 \setlength{\tabcolsep}{4pt}
		\begin{tabular} {c|p{1.9cm}|p{.8cm}||c|p{1.8cm}|p{.8cm} }
		\toprule
			{ }&Class  &   $n_l^c$ & { } &Class  &  $n_l^c$\\ \hline
			 \cellcolor[rgb]{0.24,0.29,0.95}& Corn-high & 3387  & \cellcolor[rgb]{0.95,0.83,0.13}& Hay & 50045 \\
			 \cellcolor[rgb]{0.47,0.17,0.12}& Corn-mid & 1740  & \cellcolor[rgb]{0,0.77,0.92}& Grass/Pasture & 5544   \\
			 \cellcolor[rgb]{0.6,0.95,0.36} & Corn-low & 356  & \cellcolor[rgb]{0.37,0.01,0.57}& Cover crop 1& 2746  \\
			 \cellcolor[rgb]{0.71,0.42,0.99}& Soy-bean-high & 1365   & \cellcolor[rgb]{0,0.37,0.03}& Cover crop 2& 2164 \\
			 \cellcolor[rgb]{0.78,0.08,0.23}& Soy-bean-mid & 37865  & \cellcolor[rgb]{0.66,0.88,0.83}& Woodlands & 48559 \\
			 \cellcolor[rgb]{0.84,0.92,0.23}& Soy-bean-low & 29210 & \cellcolor[rgb]{0.75,0.04,0.705}& Highway & 4863 \\
			 \cellcolor[rgb]{0.89,0.57,0.99}& Residues & 5795 & \cellcolor[rgb]{0.82,0.50,0}& Local road & 502 \\
			 \cellcolor[rgb]{0.94,0.02,0.36}& Wheat & 3387 & \cellcolor[rgb]{0.96,0.36,0.02}& Buildings & 546 \\\hline
			 \multicolumn{4}{c}{}&Total& 198074 \\
			\bottomrule
		\end{tabular}	
		}
	\label{tab:indianPine}
\end{table}

\item[c)] Houston 2013 (CASI spectrometer VHR HS + LiDAR data). The third dataset depicts an urban area nearby the campus of the University of Houston~(see Fig.~\ref{fig:DFC}). The dataset was proposed as the challenge of the IEEE IADF Data Fusion Contest 2013~\citep{grss_dataset}. The hyperspectral image was acquired by the CASI sensor (144 spectral bands at 2.5m resolution).  An aerial LiDAR scan was also available: a digital surface model (DSM) at the same resolution as the hyperspectral image was extracted, coregistered and used as an additional band in the input space. Fifteen urban land-use classes are to be classified (Tab.~\ref{tab:DFC}). Two preprocessing steps have been performed: 1)  histogram matching has been applied to the large shadowed area in the right part of the image (cf. Fig~\ref{fig:DFC}), in order to reduce domain adaptation problems~\citep{Cam13}, which are not the topic of this study: the shadowed area has been extracted by segmenting a near-infrared band and the matching with the rest of the image has been applied; 2) A height trend has been removed from the DSM, by applying a linear detrending of 3m from the West along the x-axis. Two classification experiments were performed with this data:

\begin{itemize}
\item \emph{Houston 2013A}: we consider the left part of the image, which is unaffected by the cloud shadow. This corresponds to an image of size ($349 \times 1100$) pixels. The same subsampling was applied to the LiDAR DSM. The whole ground truth within the red box in Figure~\ref{fig:DFC}c was used to extract the train and test samples.
\item \emph{Houston 2013B}: the whole image was considered. Separate training and test set (in green and red in Fig.~\ref{fig:DFC}d, respectively), are considered instead of a random extraction. In this case, even though the projected shadow has been partially corrected by the local histogram matching, some spectral drift remains between the test samples (some of which are under the shadow) and the training ones (which are only in the illuminated areas). This was the setting of the IEEE IADF Data Fusion Contest 2013 and aimed at classification under dataset shift~\citep{Cam13}. This problem is much more challenging than \textsc{Houston 2013A} and we use it as a benchmark against the state of the art, i.e. the results of the contest. However, remind that our method is not designed to solve domain adaptation problems explicitly. 
\end{itemize}

\begin{figure*}[t!]
	\centering
	\begin{tabular}{c}
			(a) CASI image after local histogram matching\\
			\includegraphics[width=.9\textwidth]{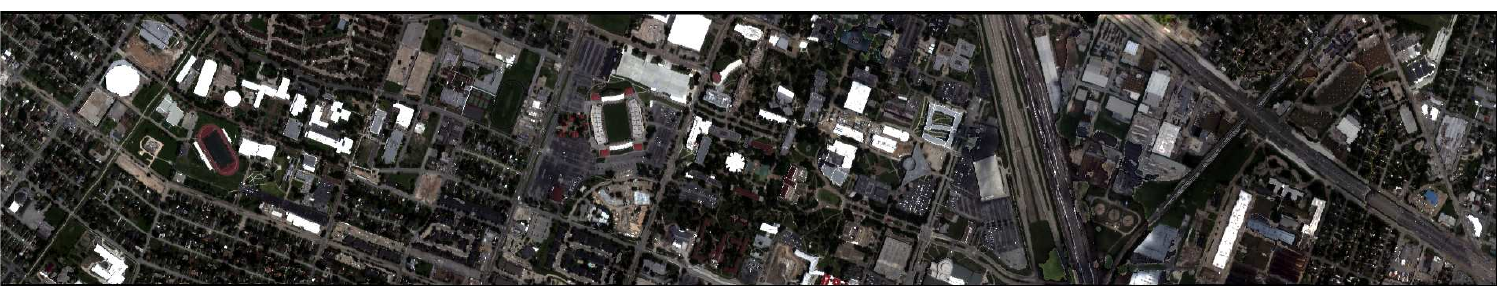}\\	
 			(a) Detrended LiDAR DSM [m]\\ 			
			\includegraphics[width=.9\textwidth]{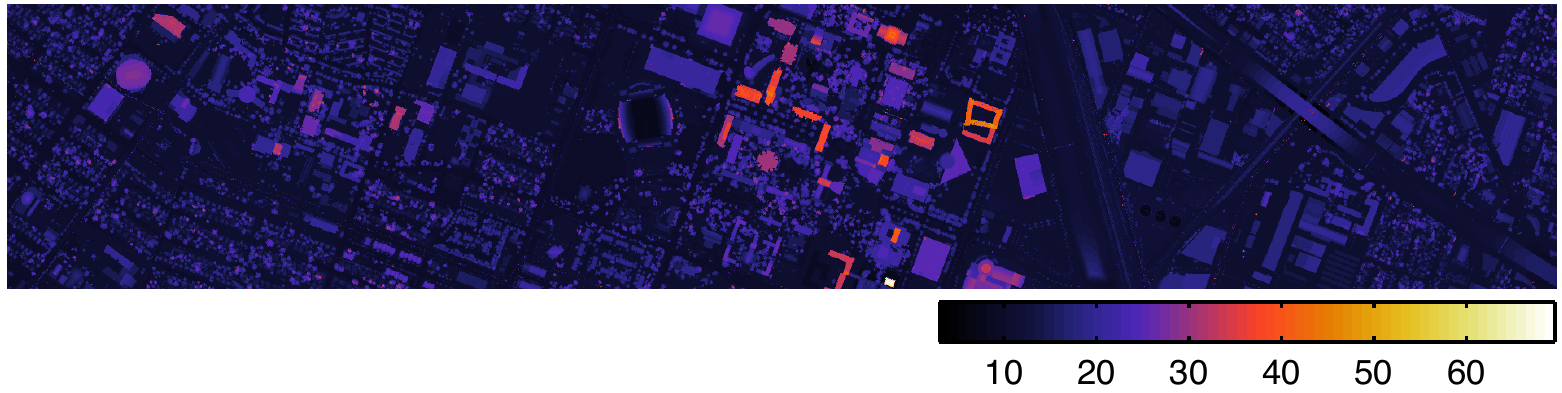}\\	
 			(c) Ground truth\\
			\includegraphics[width=.9\textwidth]{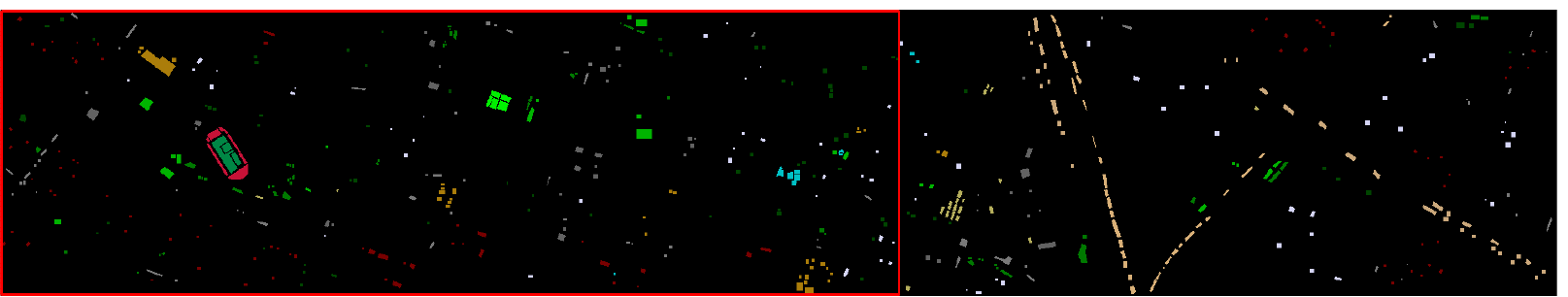}\\
			 (d) Training samples (green) vs test samples (red)\\
			\includegraphics[width=.9\textwidth]{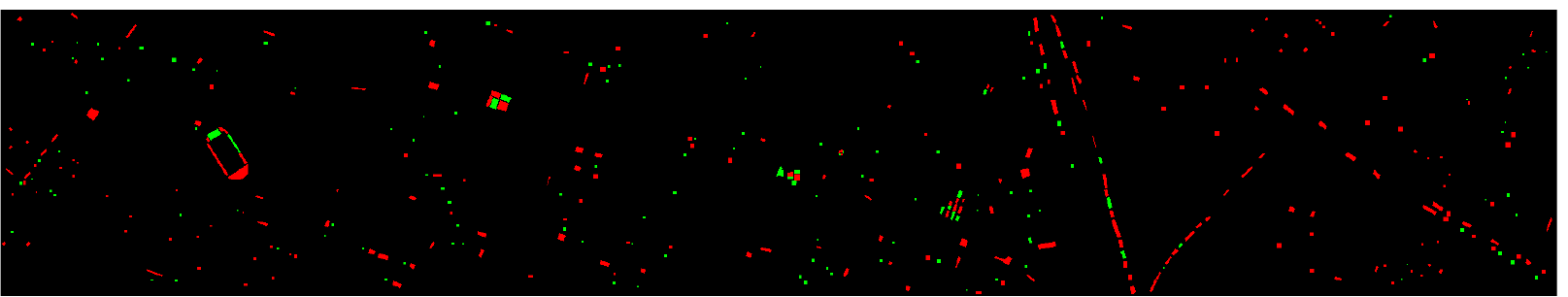}\\		

	\end{tabular}
	\caption{Houston 2013.(a) RGB composition of the CASI data, (b) DSM issued from the LiDAR point cloud and (c) train and test ground truths.
 (for color legend, see Tab.~\ref{tab:indianPine}). The area in the red box of the (c) panel has been used in the \textsc{Houston2013A} experiment, while the whole area has been used in the \textsc{Houston2013B} experiment, with (d) a training/test separation shown in the last panel (green: training, red: test). Unlabeled samples are in black.}
	\label{fig:DFC}
\end{figure*}

\begin{table}[t!]
	\centering
	\caption{Classes and samples ($n_l^c$) of the ground truth of the Houston  2013 dataset (cf. Fig.~\ref{fig:DFC}).}
	{\small
	 \setlength{\tabcolsep}{4pt}
		\begin{tabular} {c|p{1.95cm}|p{.7cm}||c|p{1.9cm}|p{.7cm} }
		\toprule
			{ }&Class  &   $n_l^c$ & { } &Class  &  $n_l^c$\\ \hline
  \cellcolor[rgb]{0,0.7098,0}&  Healthy grass &1231&   \cellcolor[rgb]{0.4745,0.4745,0.4745}&  Road&1219 \\
  \cellcolor[rgb]{0,0.4784,0.0039}&  Stressed grass &1196&  \cellcolor[rgb]{0.7961,0.6627,0.4824}&  Highway&1224 \\
  \cellcolor[rgb]{0.0039,0.5333,0.2706} & Synthetic grass&697 &  \cellcolor[rgb]{0.8549,0.6941,0.4745}&  Railway&1162 \\
  \cellcolor[rgb]{0,0.2745,0}&  Trees&1239&  \cellcolor[rgb]{0.3882,0.3882,0.3882}& Parking Lot 1&1233 \\
  \cellcolor[rgb]{0.6745,0.4902,0.0431}&  Soil&1152 & \cellcolor[rgb]{0.7137,0.6824,0.3647}& Parking Lot 2&458 \\
  \cellcolor[rgb]{0,0.7569,0.7765} & Water&325 &  \cellcolor[rgb]{0.0039,0.9412,0}&  Tennis Court&428 \\
  \cellcolor[rgb]{0.4706,0,0.0039}&  Residential&1260& \cellcolor[rgb]{0.7843,0.0706,0.2196} & Running Track&660 \\
  \cline{4-6}
  \cellcolor[rgb]{0.8353,0.8353,0.9686}&  Commercial &1219& \multicolumn{1}{l}{ }& Total & 14703 \\
			\bottomrule
		\end{tabular}	
		}
	\label{tab:DFC}
\end{table}

\end{itemize}

\subsection{Setup of experiments}
For every dataset, all the features have been mean-centered and
normalized to unit norm. This normalization is mandatory due to  the
optimality conditions, which is based on a scalar product (thus depending linearly on
 the norm of the feature).

In all the experiments, we use the multiclass logistic  classifier (MLC) with $\ell_1\ell_2$ norm implemented in the \textsc{SPAMS} package\footnote{\url{http://spams-devel.gforge.inria.fr/}}. We start by training a model with all available bands (plus the DSM in the \textsc{Houston2013A/B}  case) and use its result as the first active set. \blue{ Therefore, we do not reduce the dimensionality of the data prior to the feature generation. Regarding the active set itself}, we used the following parameters:

\begin{itemize}
\item[-] \blue{ The stopping criterion is a number of iterations: $150$ in the \textsc{Pines 1992, 2010} and \textsc{Houston 2013 B} and $100$ in the \textsc{Houston 2013A} case (the difference explained by faster convergence in the last dataset).}
\item[-] A minibatch is composed of filters extracted from $20$ bands, randomly selected. In the \textsc{Houston 2013A/B} case, the DSM is added to each minibatch.
\item[-] The possible filters are listed in Tab.~\ref{tab:filt}. Structuring elements ($SE$) can be disks, diamonds, squares or lines. If a linear structuring elements is selected, an additional orientation parameter is also generated ($\alpha \in [-\pi/2, \dots\pi/2]$). These filters are among those generally used in remote sensing hyperspectral classification literature (see~\citet{Fauvel13}), but any type of spatial or frequency filter, descriptor or convolution can be used in the process.
\item[-] A single minibatch can be used twice (i.e. once a first filter has been selected, it is removed and  Eq.~\eqref{eq:optnotinvarphi} is re-evaluated on the remaining filters after re-optimization of the MLC classifier).
\end{itemize}

\begin{table}[!t]
\caption{Filters considered in the experiments ($B_i, B_j$: input bands indices ($i, j \in [1, \ldots b]$); $s$: size of moving window, $SE$ : type of structuring element; $\alpha$: angle).}
\label{tab:filt}
\begin{tabular}{lp{3.7cm}|p{2.9cm}}
\toprule
\multicolumn{2}{l}{Filter} & $\theta$ \\
\hline\hline
\multicolumn{2}{l}{ Morphological}\\
- & Opening / closing & $B_i$, $s$, $\alpha$ \\
-& Top-hat opening / closing & $B_i$, $s$, $SE$, $\alpha$\\
-& Opening / closing by reconstruction & $B_i$, $s$, $SE$, $\alpha$\\
-& Opening / closing by reconstruction top-hat& $B_i$, $s$, $SE$, $\alpha$\\

\hline
\multicolumn{2}{l}{Texture}\\
-& Average & $B_i$, $s$ \\
-& Entropy & $B_i$, $s$ \\
-& Standard deviation & $B_i$, $s$ \\
-& Range & $B_i$, $s$ \\

\hline
\multicolumn{2}{l}{Attribute} \\ 
-& Area & $B_i$, Area  threshold \\
-& Bounding box diagonal  & $B_i$,  Diagonal threshold \\

\hline
\multicolumn{2}{l}{Band combinations}\\
- & Simple ratio& $B_i / B_j$ \\
-& Normalized ratio &  $(B_i-B_j) / (B_i + B_j)$ \\
-& Sum &  $B_i + B_j$ \\
-& Product &  $B_i * B_j$ \\

\bottomrule

\end{tabular}
\end{table}

In each experiment, we start by selecting an equal number of labeled pixels per class $l_c$: we extracted 30 random pixels per class in the \textsc{Indian Pines 1992} case,  $60$ in the \textsc{Indian Pines 2010} and in the \textsc{Houston 2013A/B} case\footnote{\blue{ When the number of pixels available was smaller than $l_c$, we extracted 80\% for training and left the rest for testing}}. The difference in the amount of labeled pixels per class is related to  i) the amount of labeled pixels available per task and ii) the complexity of the problem at hand. As test set, we considered all remaining labeled pixels, but disregard those in the spatial vicinity of the pixels used for training. In the \textsc{Indian Pines 1992} case, we consider all labeled pixels out of a $3 \times 3$ window around the training pixels, in the \textsc{Indian Pines 2010} case a $7 \times 7$ window and in the \textsc{Houston 2013A} case a $5 \times 5$ window. The difference is basically related to the images spatial resolution. In the \textsc{Houston 2013B} case, a spatially disjoint test set was provided in a separate file and was therefore used for testing purposes without spatial windowing.

When considering the hierarchical model \textsc{ASH-bands}, every feature
that is added to the active set is also added to the input bands $B$ 
(see line $10$ of
Algorithm~\ref{alg2}). In order to penalize overcomplex deep features,
we considered $\gamma = 1.1^h$, where $h$ is the depth of the feature
defined in Section~\ref{sec:deep}.                                                                                                 When adding filters issued from two inputs (as, for example, band
ratios) $h = max(h_{B_i},h_{B_j})+1$.

Each experiment was repeated $5$ times, by random sampling of the initial training set (the test set also varies in the \textsc{Indian Pines 1992/2010} and \textsc{Houston 2013A} datasets, since it depends on the specific location of the training samples). Average performances, along with their standard deviations, are reported.

\section{Results and discussion}\label{sec:res}
In this section, we present and discuss both the numerical results obtained and the feature selected in the \textsc{AS-Bands} (shallow) and \textsc{ASH-Bands} (deep) algorithms.

\subsection{Performances along the iterations} 
\begin{figure}[!t]
\begin{tabular}{cc}
\multicolumn{2}{c}{\textsc{Indian Pines 1992}}\\
\includegraphics[width=.42\linewidth]{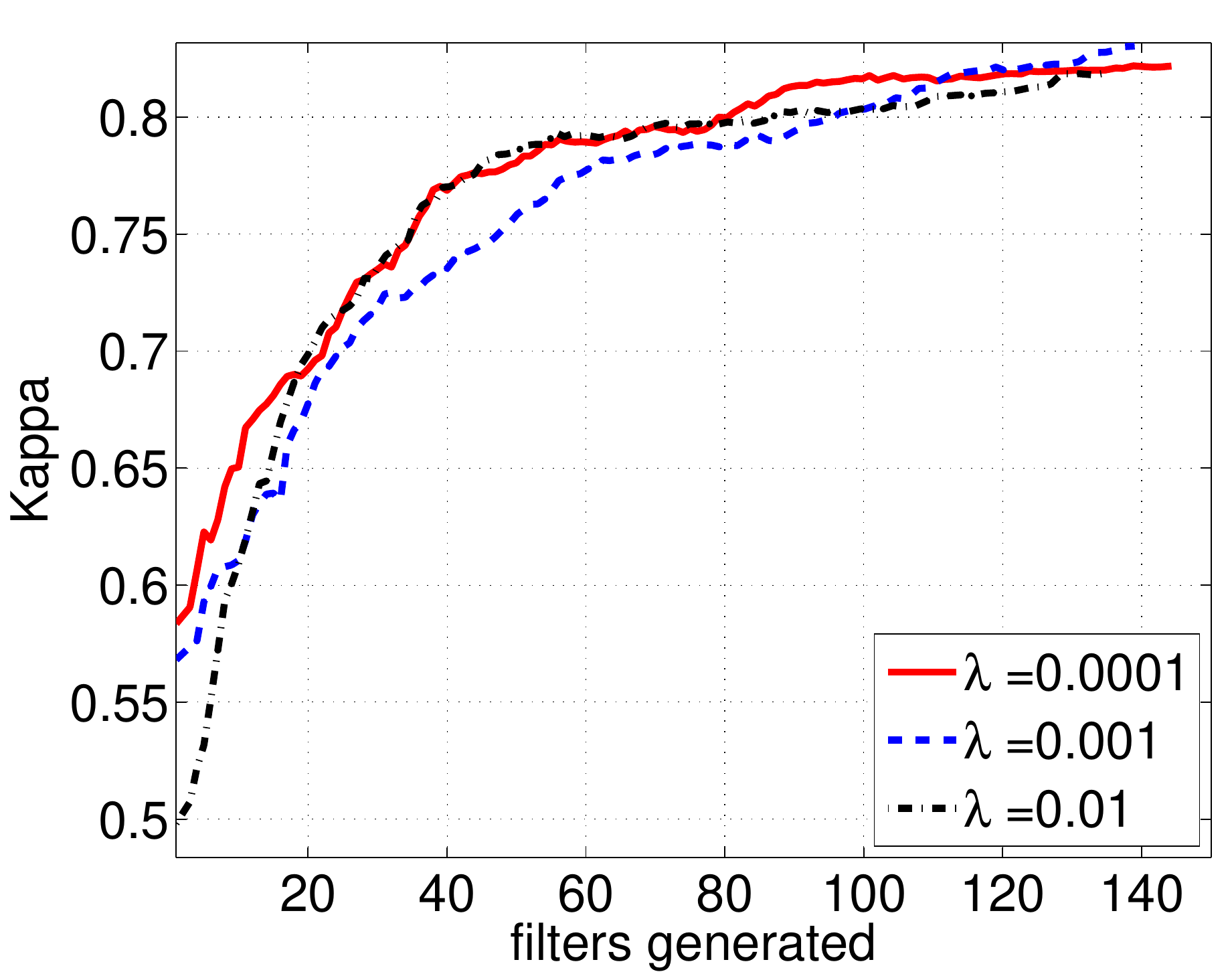}&
\includegraphics[width=.42\linewidth]{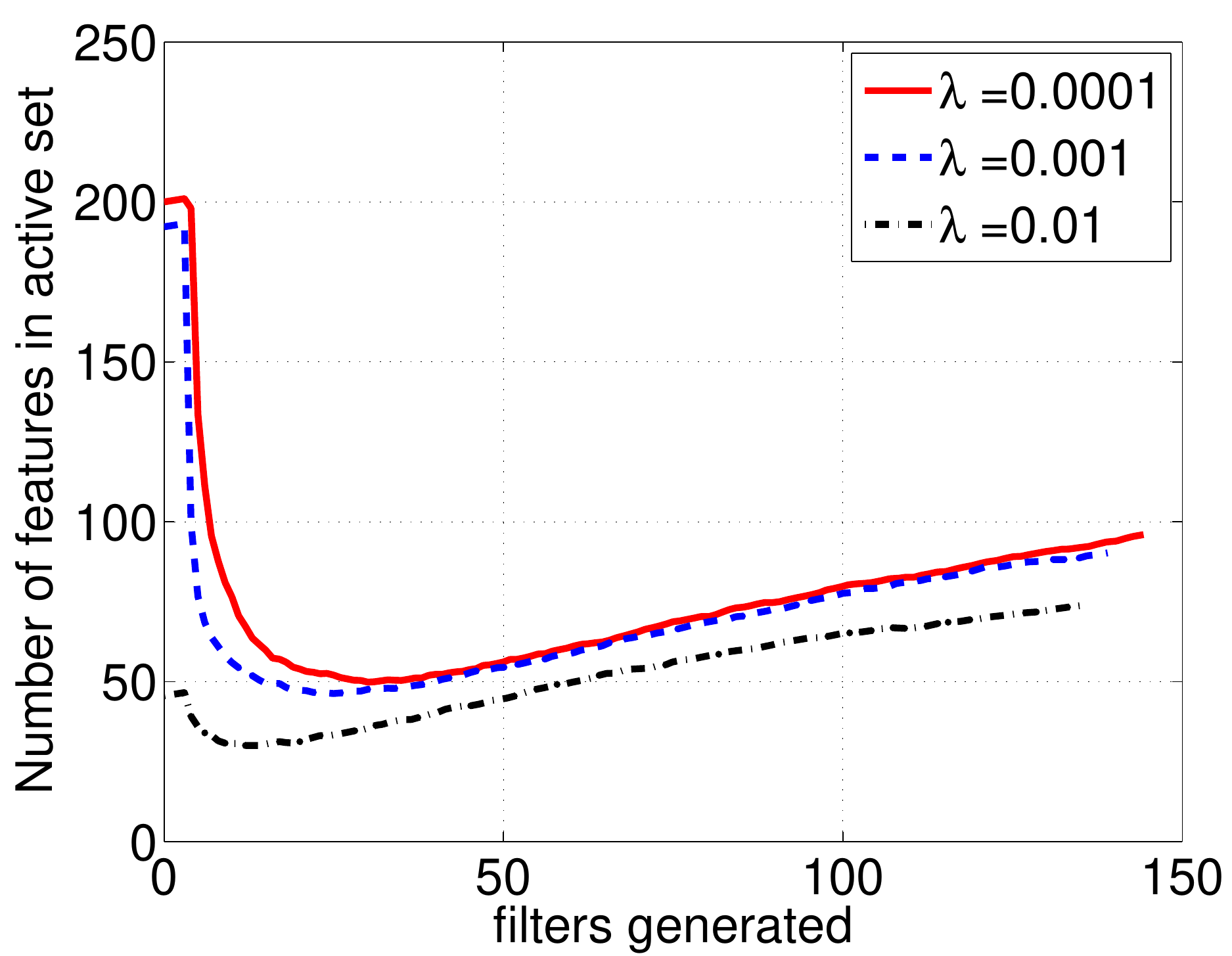}\\
\multicolumn{2}{c}{\textsc{Indian Pines 2010}}\\
\includegraphics[width=.42\linewidth]{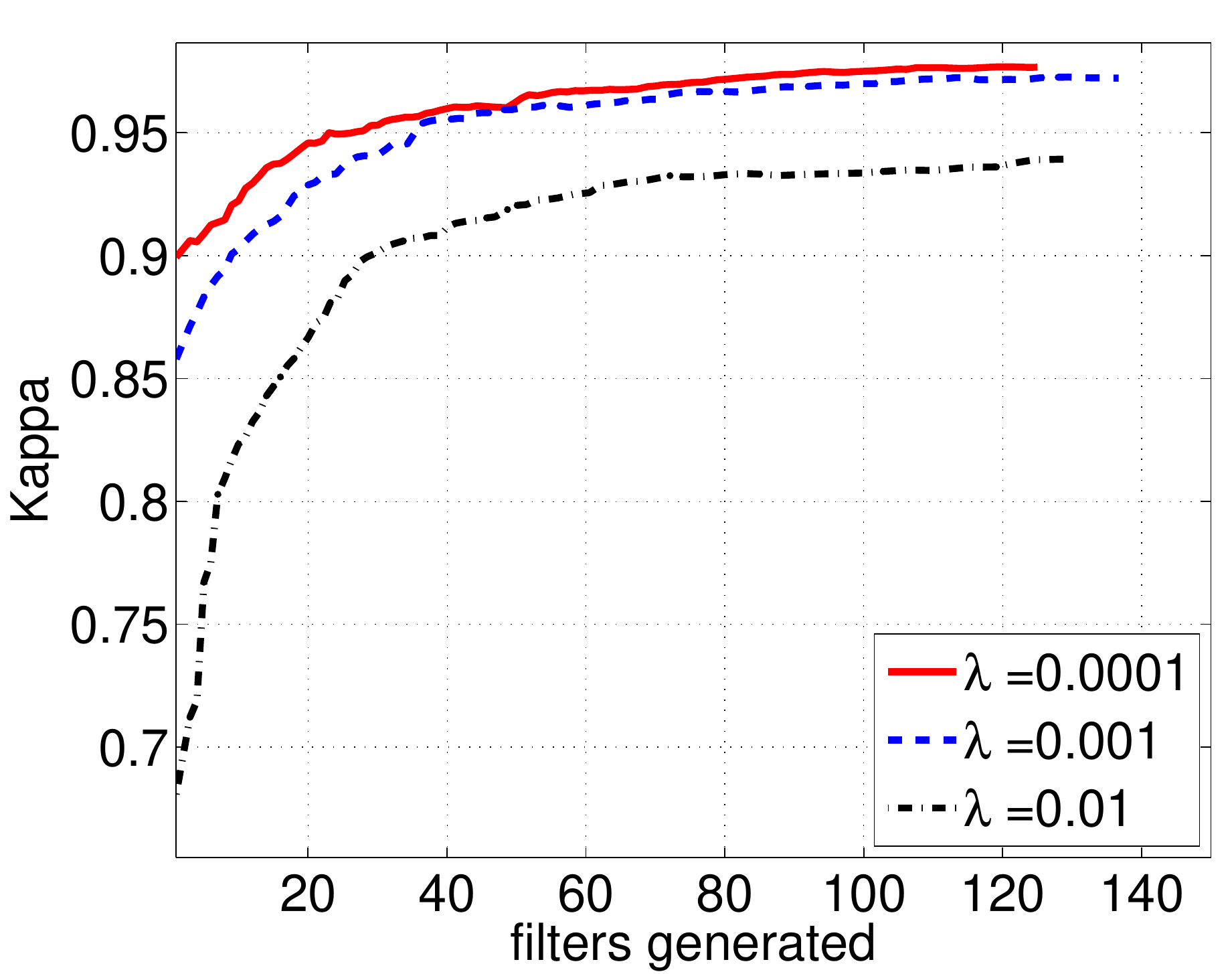}&
\includegraphics[width=.42\linewidth]{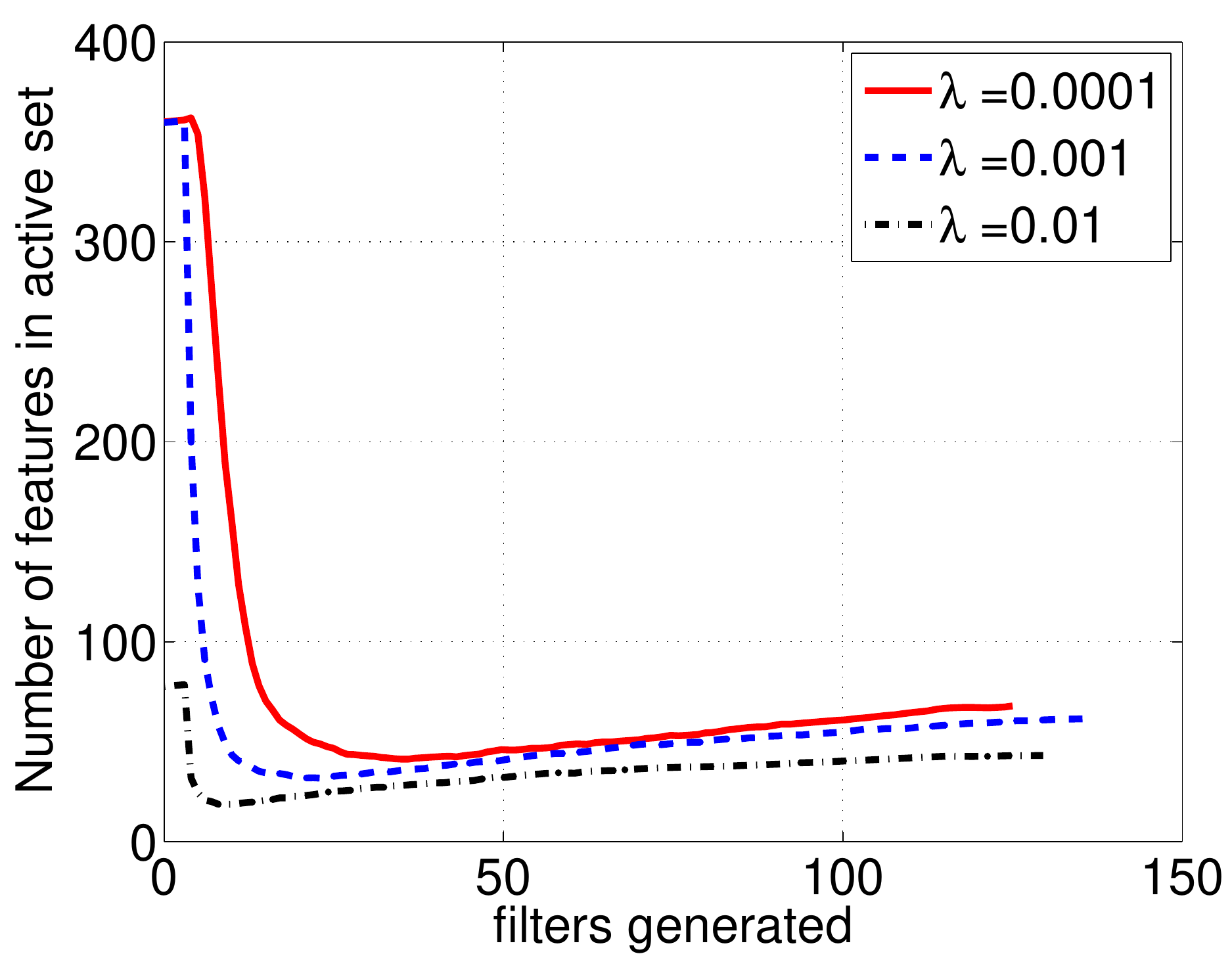}\\
\multicolumn{2}{c}{\textsc{Houston 2013A}}\\
\includegraphics[width=.42\linewidth]{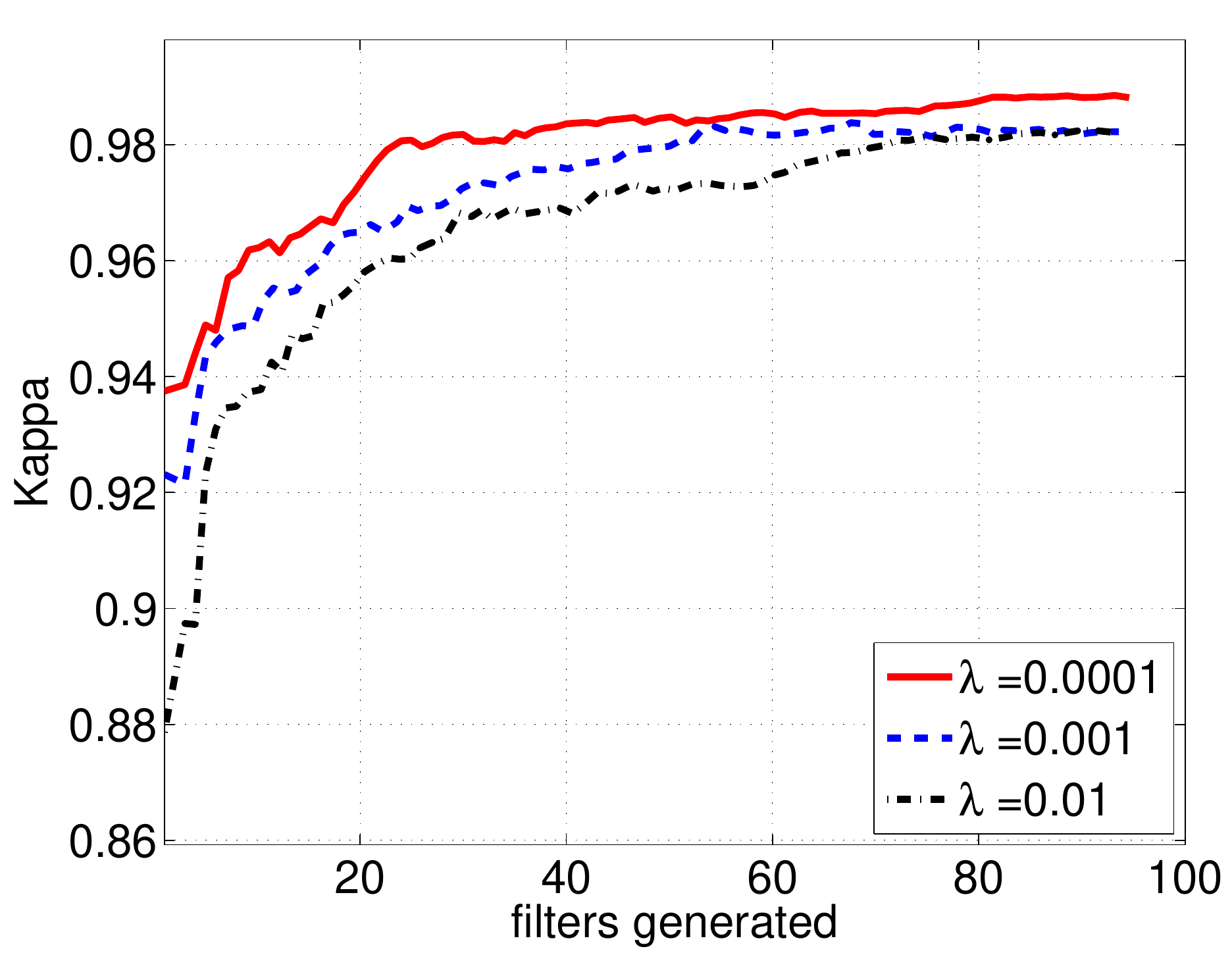}&
\includegraphics[width=.42\linewidth]{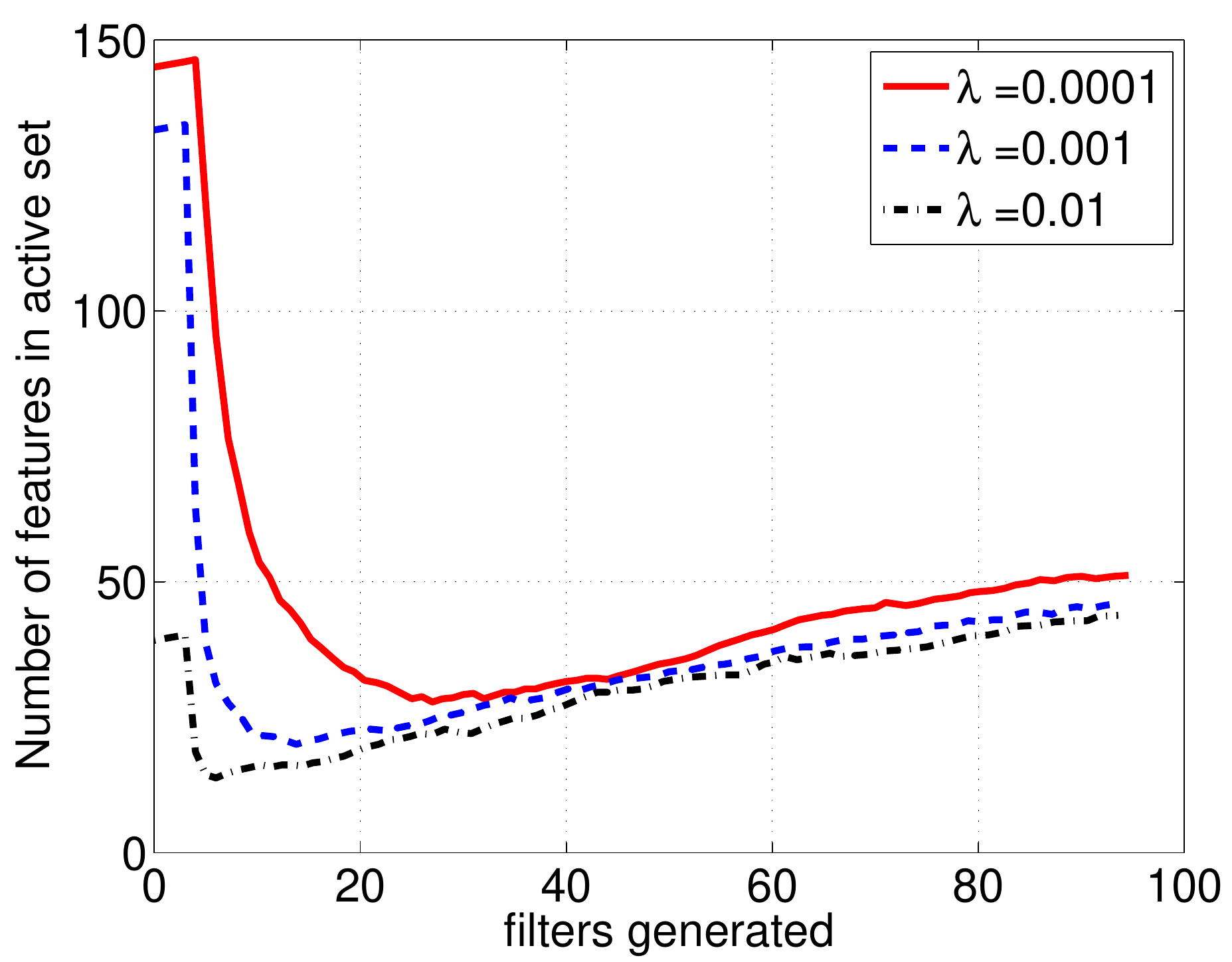}\\
\multicolumn{2}{c}{\textsc{Houston 2013B}}\\
\includegraphics[width=.42\linewidth]{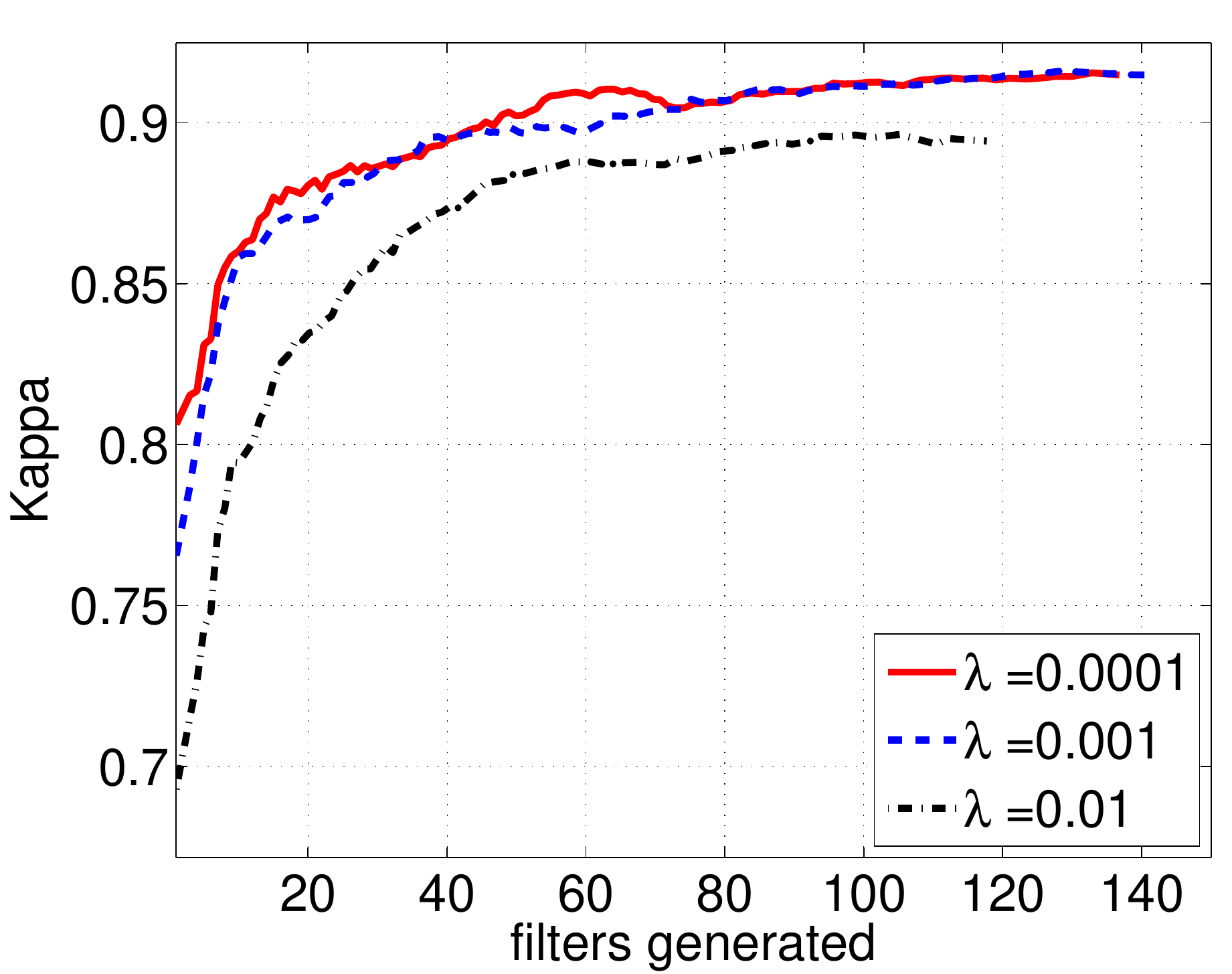}&
\includegraphics[width=.42\linewidth]{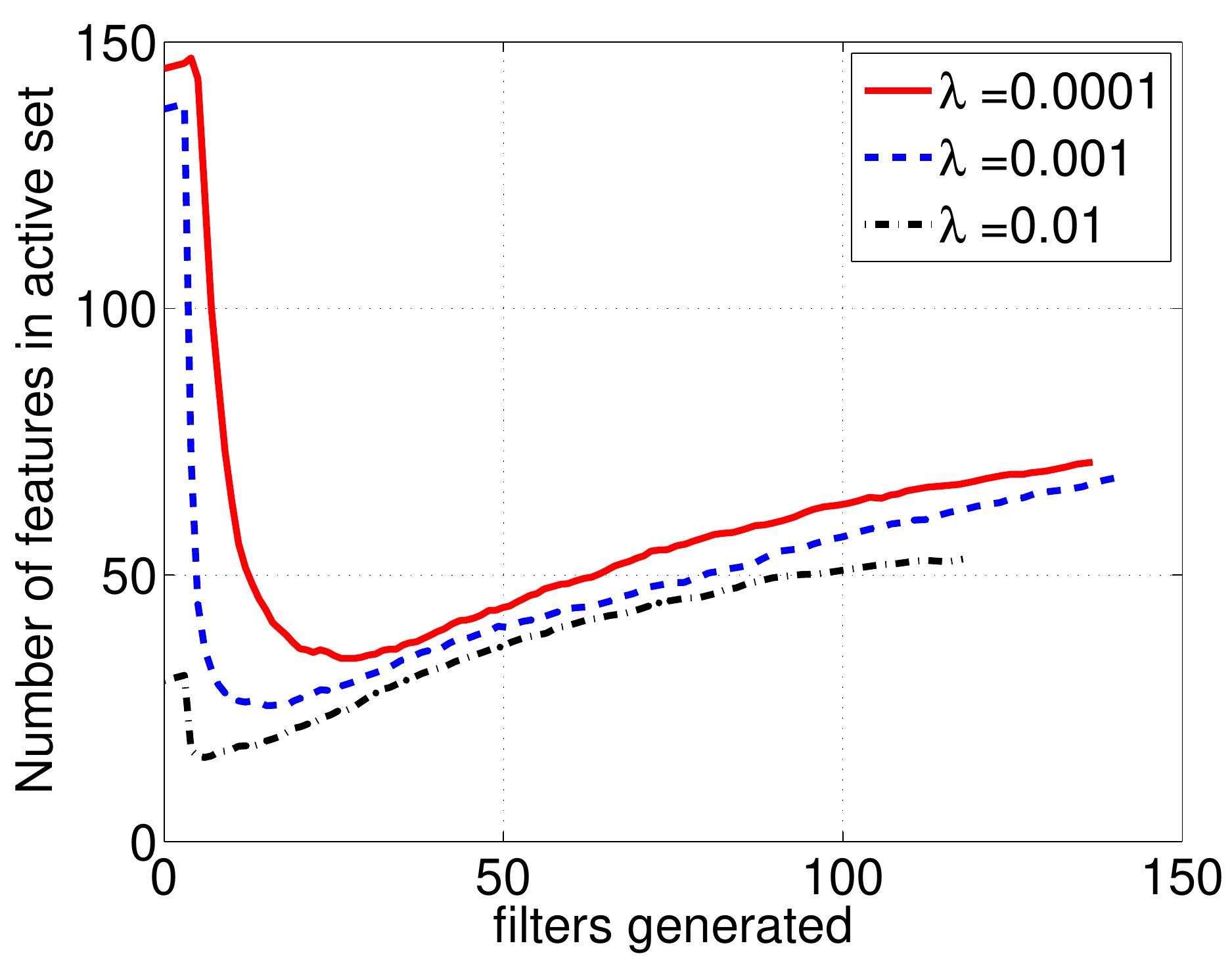}\\
\end{tabular}
\caption{Left: numerical performance (Kappa statistic) of \textsc{AS-Bands} for different degrees of regularization $\lambda$ and filtering the original bands. Right: number of active features during the iterations.}
\label{fig:resNum}

\end{figure}

\noindent \textbf{\textsc{AS-Bands}:} Numerical results for the three datasets in the \textsc{AS-Bands} (shallow) setting are provided in Fig.~\ref{fig:resNum}: the left column illustrates the evolution of the Kappa statistic~\citep{Foo04c}  along the iterations and for three levels of $\ell_1\ell_2$ regularization $\lambda$: the higher the $\lambda$ parameter, the sparser the model (and the harder to violate the optimality conditions). The right column of Fig.~\ref{fig:resNum} shows the evolution of the number of features in the active set.

For all the datasets, the iterative feature learning corresponds to a
continuous, almost monotonic, increase of the performance. This is
related to the optimality conditions of Eq.~\eqref{eq:MLC}: each time
the model adds one filter $\phi_{\theta_j^*}$ to $\varphi$, the MLC cost function decreases while the classifier performances raises. Overfitting is
 prevented by the group-lasso regularization: on the one hand this regularizer promotes sparsity through the $\ell_1$ norm, while on the other hand it limits the magnitude of the weight coefficients $\W$ and promotes smoothness of the decision function by the use of the $\ell_2$ norm. Note that for the \textsc{Houston
  2013B} dataset, the final classification performance is at the same
level as the one of the winners of the contest, thus showing the
ability of our approach to compete with state of the art methods.

For each case study, the model with the lowest sparsity ($\lambda = 0.0001$)
shows the initial best performance (it utilizes more features, as
shown in the right column) and then keeps providing the best
performances. However, the model with $\lambda = 0.001$ has an initial
sparser solution and shows a steeper increase of the curve in the
first iterations. When both models provide similar performance, they
are actually using the same number of features in all cases. The
sparsest model ($\lambda = 0.01$, black line) shows the worst  results in two out
of the three datasets and in general is related to less features
selected: 
our interpretation is that the regularization  ($\lambda = 0.01$) is too strong, leading to a model that discards  relevant features and is too biased for a good prediction (even when more features are added). 
 As a consequence, the learning rate may be steeper than for the other models, but the model does not converge to an optimal solution. 

\begin{figure}[!t]
\begin{tabular}{cc}
\multicolumn{2}{c}{\textsc{Indian Pines 1992}}\\
\includegraphics[width=.42\linewidth]{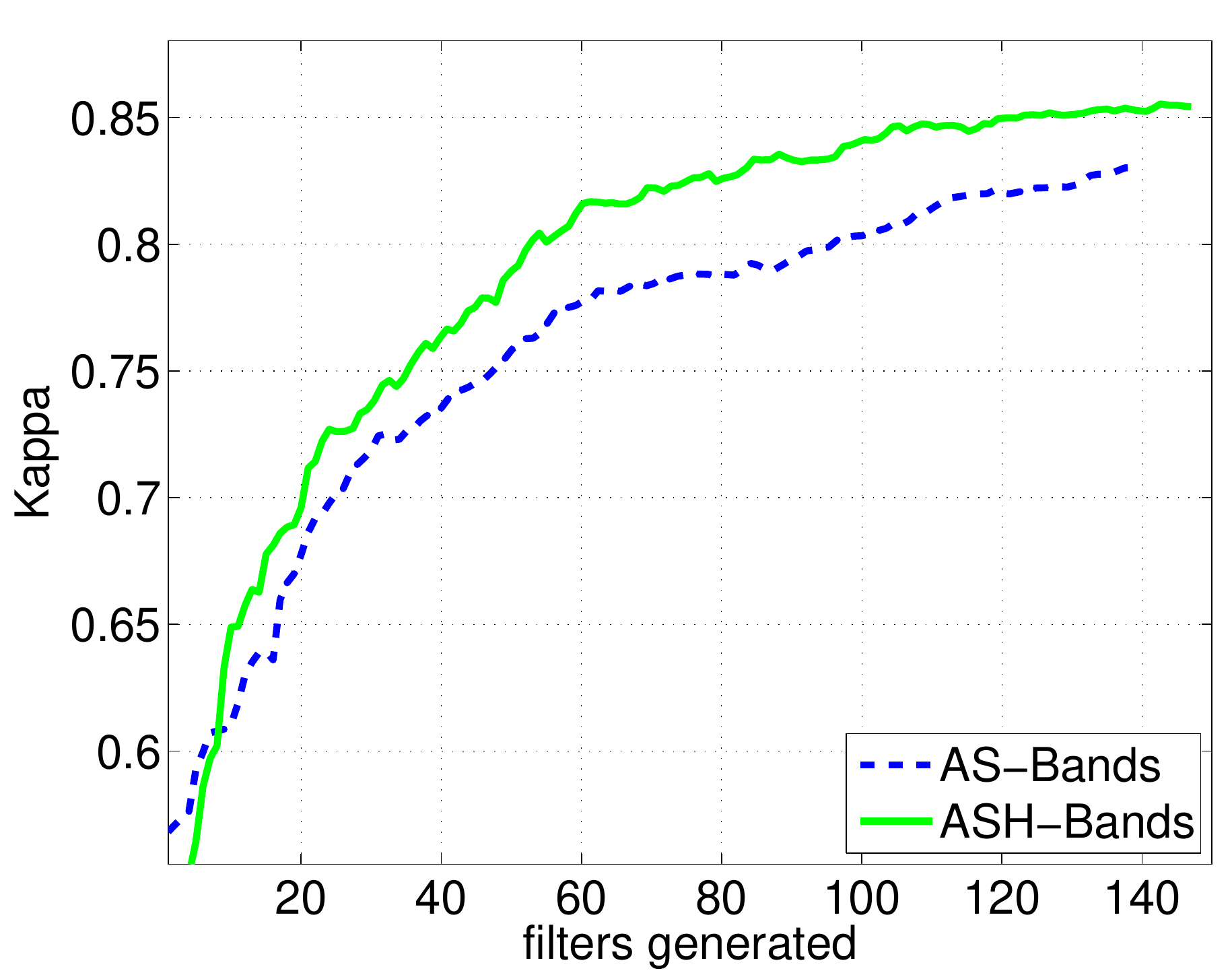}&
\includegraphics[width=.42\linewidth]{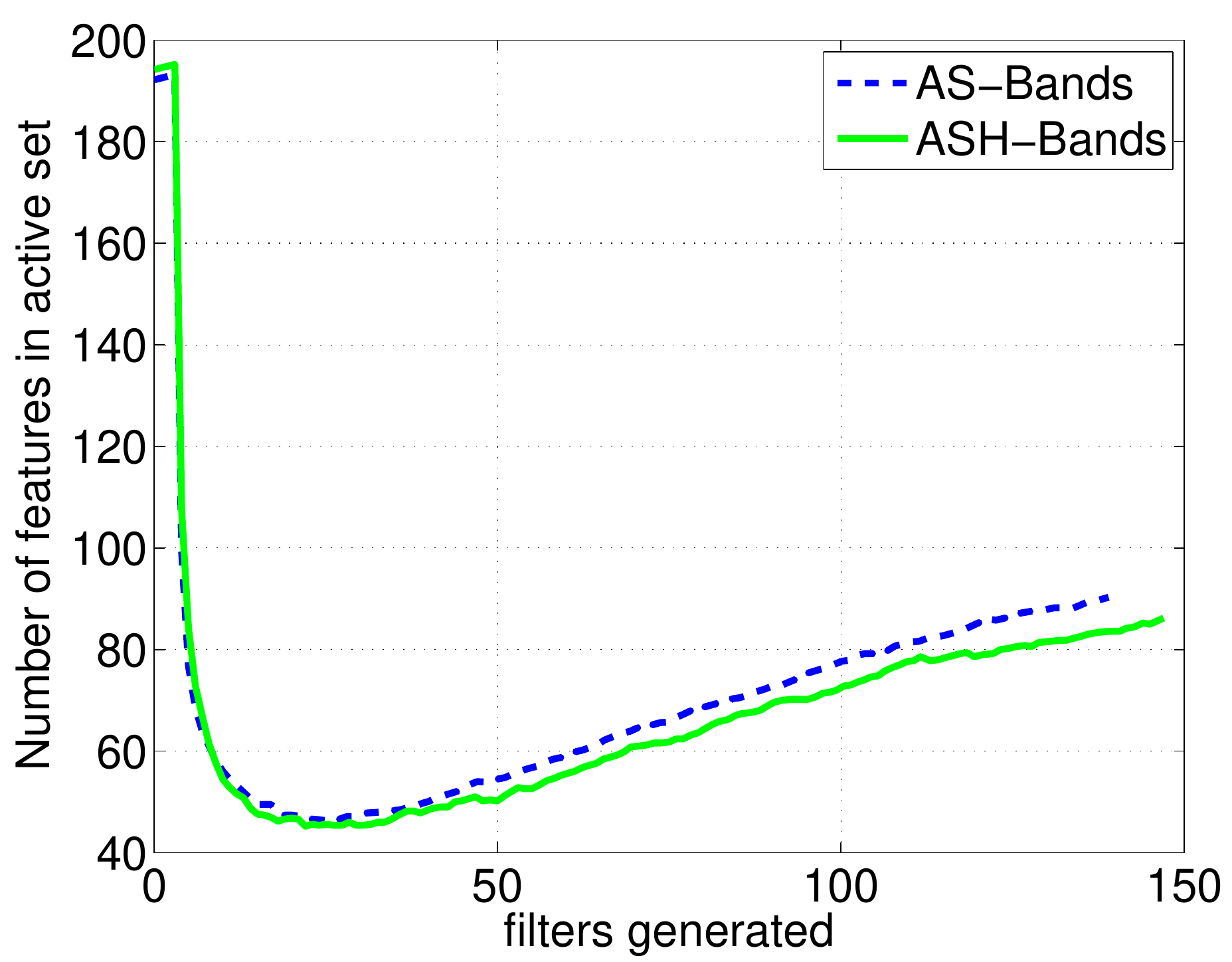}\\
\multicolumn{2}{c}{\textsc{Indian Pines 2010}}\\
\includegraphics[width=.42\linewidth]{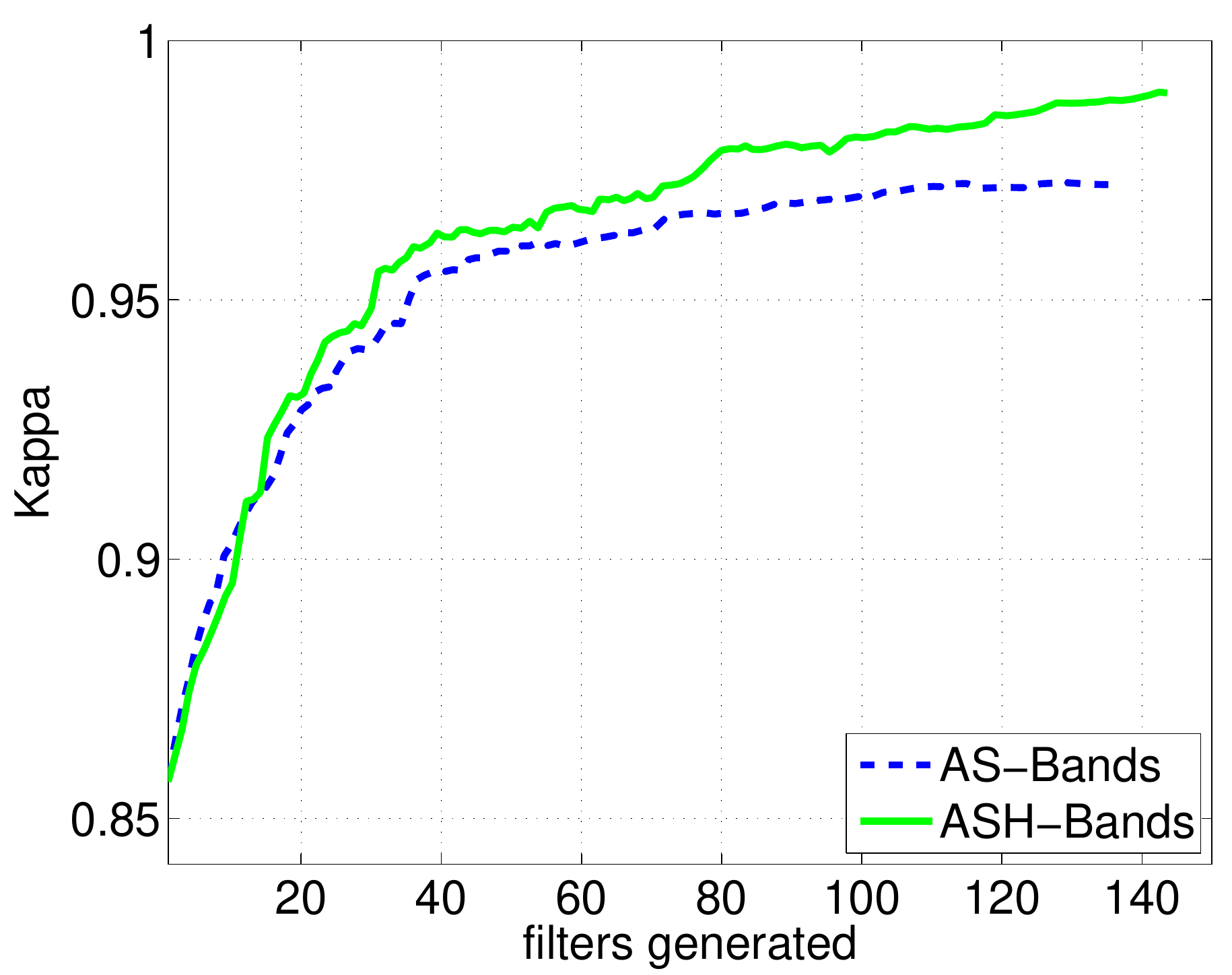}&
\includegraphics[width=.42\linewidth]{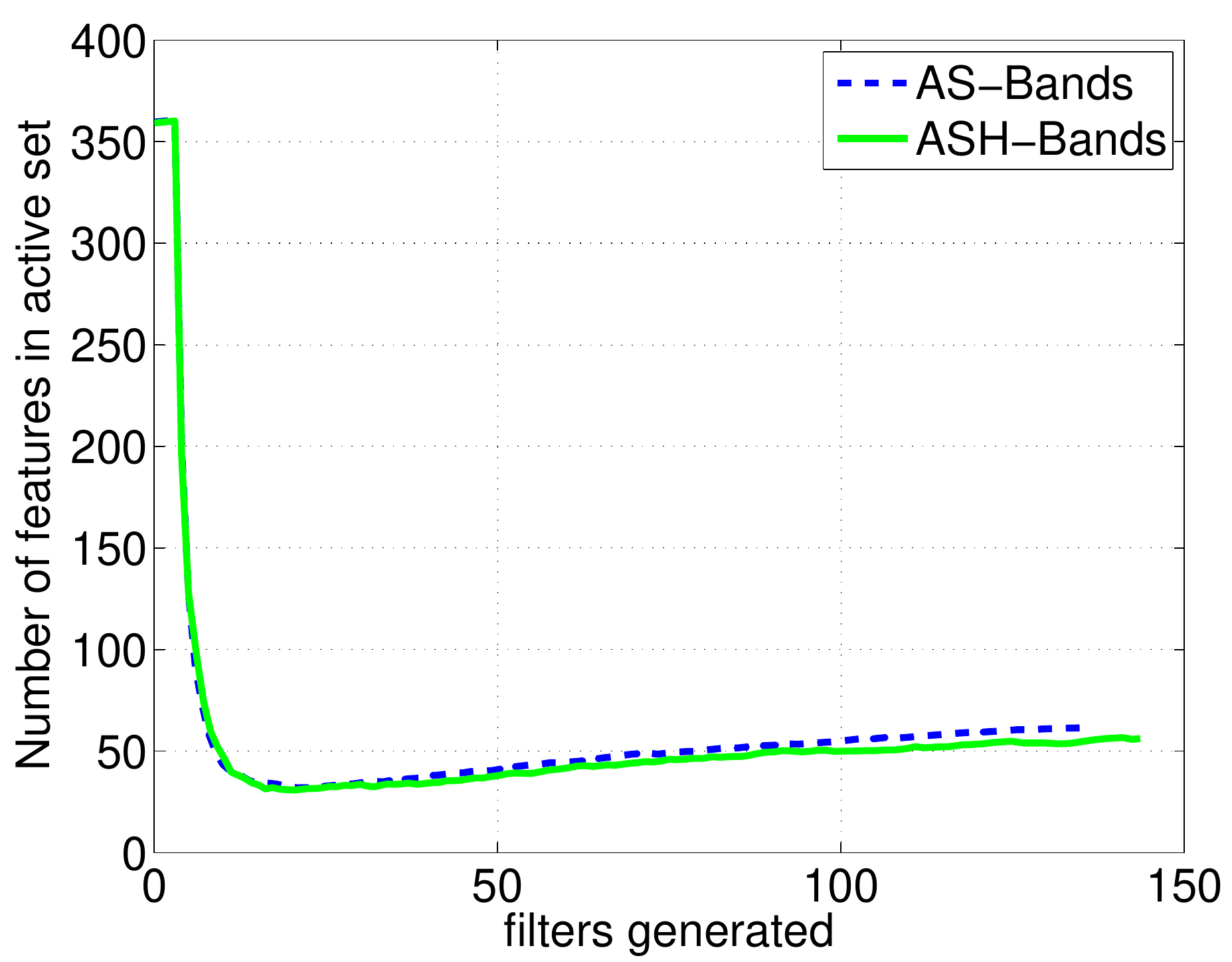}\\
\multicolumn{2}{c}{\textsc{Houston 2013A}}\\
\includegraphics[width=.42\linewidth]{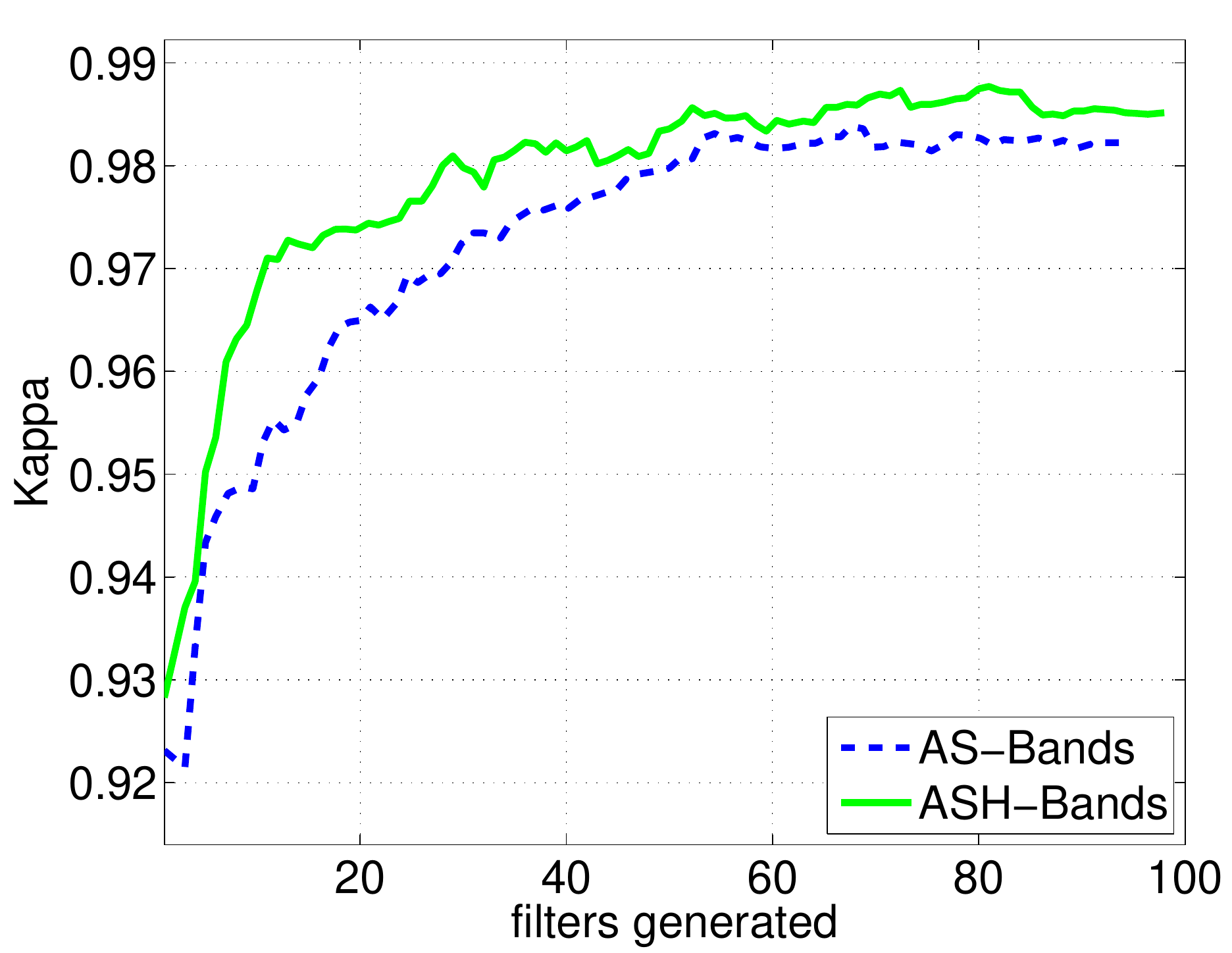}&
\includegraphics[width=.42\linewidth]{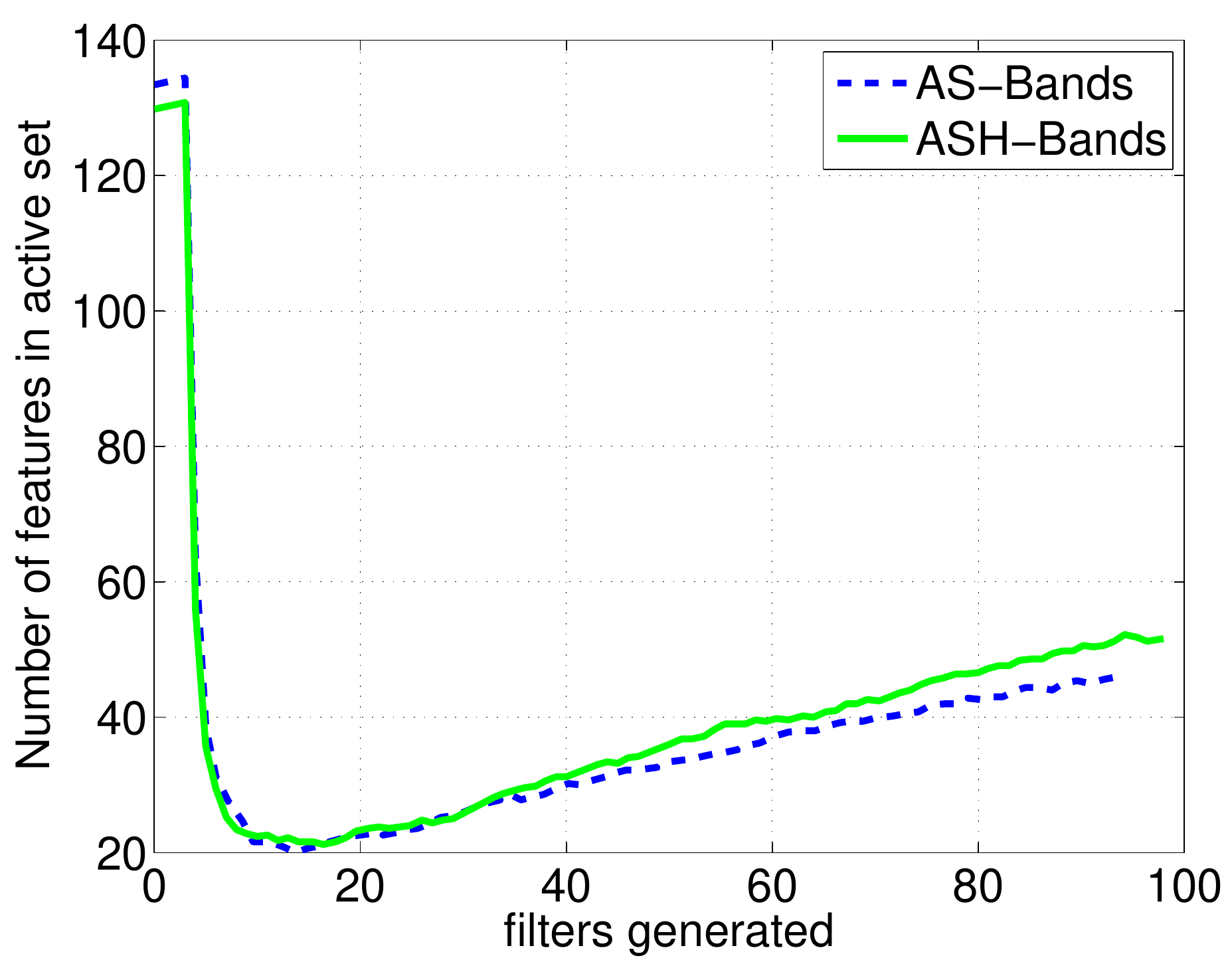}\\
\multicolumn{2}{c}{\textsc{Houston 2013B}}\\
\includegraphics[width=.42\linewidth]{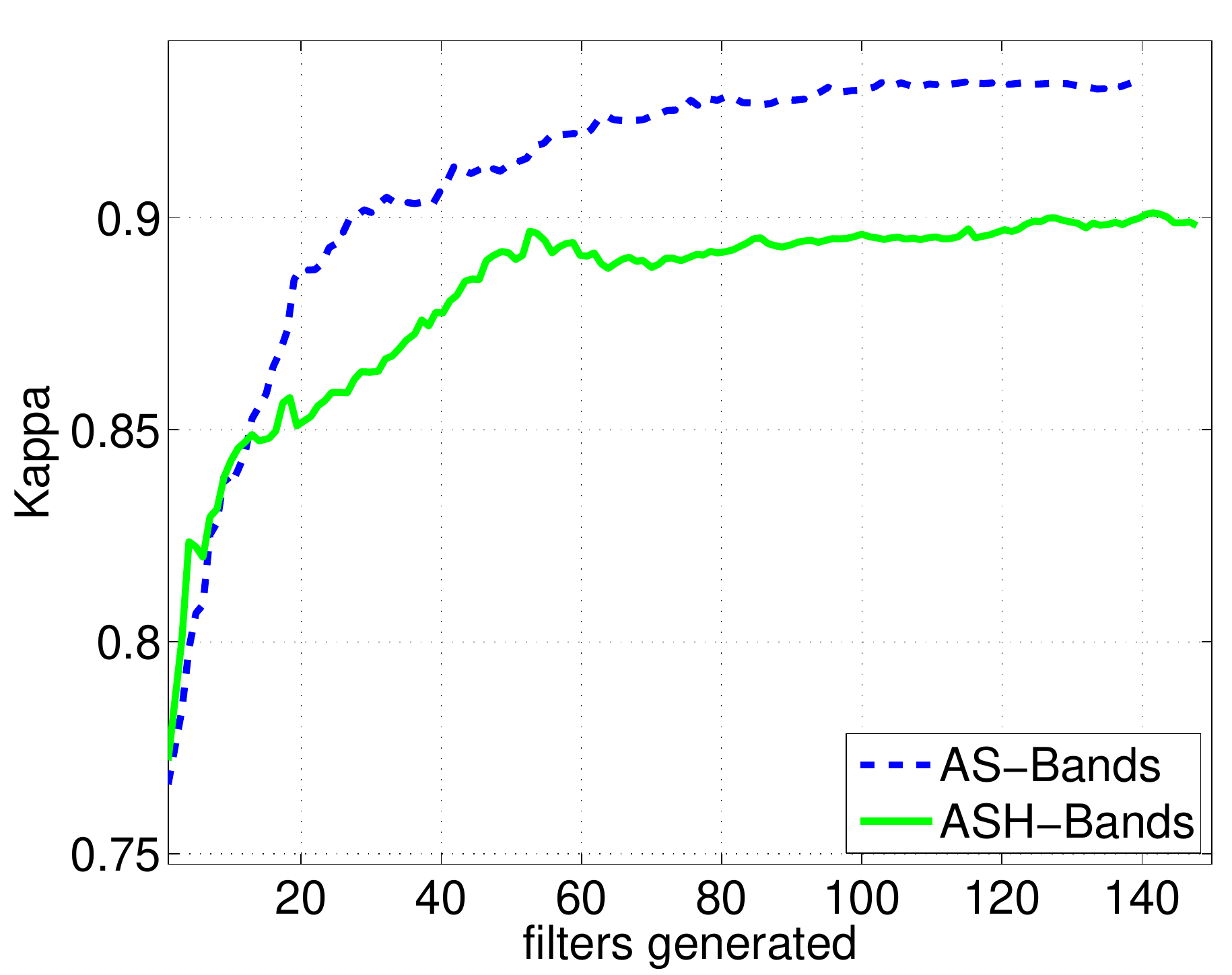}&
\includegraphics[width=.42\linewidth]{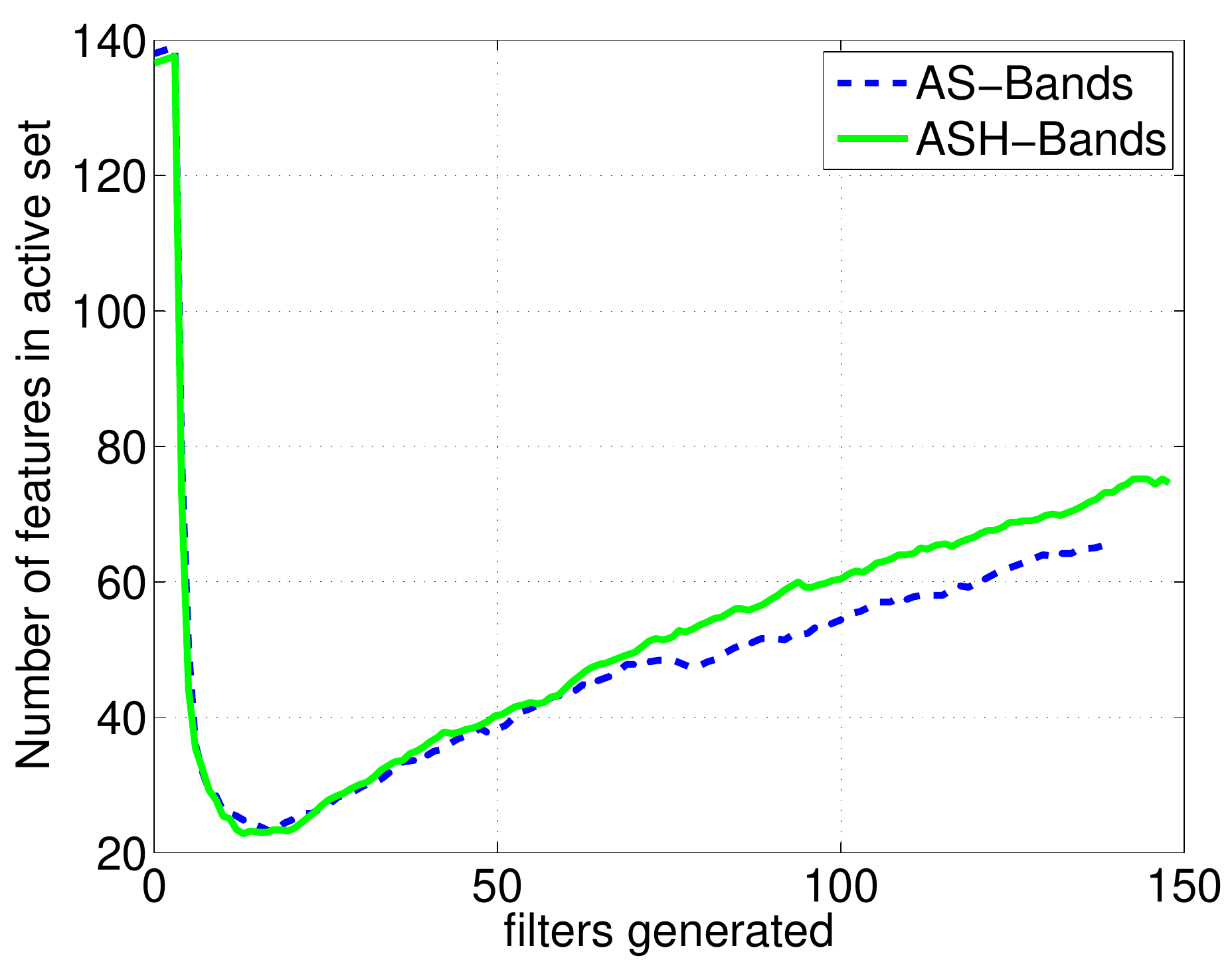}\\

\end{tabular}
\caption{Results of the \textsc{ASH-Bands} method. Left: numerical performance (Kappa statistic) for $\lambda = 0.001$. Right: number of active features during the iterations.}
\label{fig:compDeep}
\end{figure}

\begin{table*}[!t]
\caption{Results by MLC classifiers trained with the spectral bands ($\omega$), with spatial features extracted from the three first \blue{ principal components, PCs} ($s$, including morphological and attribute filters) or with the proposed active set (\textsc{AS-}). In the \textsc{Houston 2013A/B} cases, features extracted from the DSM have  been added to the input space of the baselines.}
\label{tab:compP}
\setlength{\tabcolsep}{5pt}\begin{tabular}{p{3cm}p{1.8cm}|c|c|c|c|c}
\hline
& Method & $\Omega$ & \textsc{Pines 1992} & \textsc{Pines 2010} & \textsc{Houston 2013A} & \textsc{Houston 2013B} \\
\hline\hline
No spatial info & MLC-$\omega$ & $\ell_1$ & $0.42 \pm 0.02$ & $0.58 \pm 0.01$ & $0.90 \pm 0.02$ & $0.61 \pm 0.01$\\
(baseline)&\multicolumn{2}{r|}{\# features} & $60 \pm 3$ & $107 \pm 9$ & $135 \pm 6$ & $54 \pm 3$    \\\cline{2-7}
&MLC-$\omega$ & $\ell_2$ &$0.59 \pm 0.03 $& $0.90 \pm 0.01$ &$0.92 \pm 0.02$& $0.80 \pm 0.01$\\
&\multicolumn{2}{r|}{\# features} & $200$ & $360$ & $145$& $145$ \\\hline 
Spatial info&\textsc{AS-bands}& $\ell_1\ell_2$ & $0.83\pm 0.02$ &$0.98\pm0.01$&$\mathbf{0.98\pm0.01}$&$\mathbf{0.93\pm0.01}$ \\
from bands& \multicolumn{2}{r|}{\# features} &  $96 \pm 5$& $68 \pm 5$& $46 \pm 4$ & $71 \pm 3$ \\\cline{2-7}
(proposed)&  \textsc{ASH-bands}& $\ell_1\ell_2$ & $0.85 \pm 0.03$ & $\mathbf{0.99 \pm 0.001}$ & $\mathbf{0.99\pm0.01}$& $0.90\pm0.03$ \\
& \multicolumn{2}{r|}{\# features} & $86 \pm 6$ &$56 \pm 3$& $52\pm5$& $75 \pm 2$ \\\hline\hline
Spatial info from &MLC-$s$ & $\ell_1$ & $0.85 \pm 0.02$ & $0.84 \pm 0.01$ &$0.97 \pm 0.01$&  $0.76 \pm 0.01$\\
three top \blue{ PCs} &\multicolumn{2}{r|}{\# features} & $85 \pm 7$ & $64.2 \pm 3$ &$122 \pm 12$ &$82 \pm 5$ \\\cline{2-7}
(baseline)&MLC-$s$&$\ell_2$& $0.85 \pm 0.01$ & $ 0.96 \pm 0.01$ &$0.97 \pm 0.01$& $0.87 \pm 0.01$\\
&\multicolumn{2}{r|}{\# features} & $217$ & $228$ &269& $273$  \\\hline
Spatial info from &  \textsc{AS-pcs}& $\ell_1\ell_2$  & $\mathbf{0.89 \pm 0.03}$   &$\mathbf{0.99 \pm 0.01}$ & $\mathbf{0.98 \pm 0.01}$& $\mathbf{0.92 \pm 0.01}$ \\
all \blue{ PCs}& \multicolumn{2}{r|}{\# features}  &$82 \pm 4$  & $83 \pm 8$  &$57 \pm 4$& $64\pm4$\\\cline{2-7}
(proposed)& \textsc{ASH-pcs}& $\ell_1\ell_2$  & $\mathbf{0.88 \pm 0.01}$ & $\mathbf{0.99 \pm 0.01}$ &  $\mathbf{0.99 \pm 0.01}$ & $\mathbf{0.92 \pm 0.02}$ \\
& \multicolumn{2}{r|}{\# features}  & $102 \pm7$ & $68 \pm 2$ & $59 \pm 3$ & $74 \pm 6$ \\\hline
\hline
\end{tabular}
\end{table*}

\noindent \textbf{\textsc{ASH-Bands}:} The performance of \textsc{ASH-Bands} are compared to those of \textsc{AS-Bands} in Fig.~\ref{fig:compDeep}. The case of $\lambda = 0.001$ is shown (the blue curves of Fig.~\ref{fig:compDeep} correspond to the blue curves of Fig.~\ref{fig:resNum}). From this comparison, two tendencies can be noticed: on the one hand, \textsc{ASH-Bands} shows better learning rates when the classification problem is fixed (i.e., no spectral shifts are observed between the training and test data: \textsc{Indian Pines 1992}, \textsc{Indian Pines 2010} and \textsc{Houston 2013A}): by constructing more complex features, \textsc{ASH-Bands} can solve the classification problem in a more accurate way and without increasing substantially the size of the model (both \textsc{AS-Bands} and \textsc{ASH-Bands} show similar number of active features during the process). On the other hand, in the \textsc{Houston 2013B} case \textsc{ASH-Bands} is outperformed by the shallow model \textsc{AS-Bands} by $0.03$ in $\kappa$. The variance of the single runs is also significantly higher (see, the \textsc{ASH-Bands} row for this dataset in Tab.~\ref{tab:compP}). We interpret this slower learning rate by an overfitting of the training data in the presence of dataset shift: since the test distribution is different that the one observed in training (by the projected cloud in the hyperspectral data), the spatial filters learned seem to become too specialized in explaining the training data and are then less accurate in the case of the (shifted) test distribution. Such behavior has been documented before in deep learning literature, especially when little training examples are used to learn the features~\citep{Ben12}. Note that the classification performance is still $\kappa=0.9$ on average.

\subsection{Numerical performances at the end of the feature learning}
Comparisons with competing strategies where the MLC classifier is learned on pre-defined feature sets are reported in Table~\ref{tab:compP}. First, we discuss the performance of our active set approach when learning the filters applied on the original bands (\textsc{AS-Bands} and \textsc{ASH-Bands}):  in the \textsc{Indian
  Pines 1992} case, the \textsc{AS-} methods obtain average Kappas of $0.83$ using $96$ features and $0.85$ using $86$ features, respectively. This is a good result if compared to the upper bound of $0.86$ obtained by a classifier using the complete set of $14`627$ morphological and attribute features extracted from each spectral band (result not reported in the table)\footnote{Only squared structuring elements were used and the filter size range was pre-defined by expert knowledge.}. On \blue{ both} the \textsc{Indian Pines 2010} and \textsc{Houston 2013A} datasets, the \textsc{AS-Bands} method provided average Kappa of $0.98$. \blue{ \textsc{ASH-Bands} provided comparable results, on the average $0.01$ more accurate, but still in the standard deviation range of the shallow model. The exception is} the last dataset, \textsc{Houston 2013B}, \blue{ for which the shallow model provides a Kappa of 0.93, while} the hierarchical model is $0.03$ less accurate, as discussed in the previous section.

We compared these results to those obtained by classifiers trained on fixed raw bands (MLC$-\omega$) or on sets of morphological and
attribute filters extracted form the three first principal components
(MLC-$s$). We followed the generally admitted hypothesis that
the first(s) principal component(s) contain most of the relevant information in
hyperspectral images~\citep{Ben05}. On all the datasets, the proposed \textsc{AS-bands} method
performs remarkably well compared with models using only the spectral
information (MLC-$\omega$) and compares at worse equivalently (and
significantly better in the \textsc{Indian Pines 2010} and
\textsc{Houston 2013B} cases) with models using $\ell_2$ classifiers
(thus without sparsity) and three to four times more features including spatial information (MLC-$s$). The good performance of the $\ell_2$ method on the \textsc{Indian
  Pines 1992} dataset (Kappa observed of $0.85$) is probably due to
the application of the PCA
transform prior to classification, which, besides allowing to decrease
the dimensionality of the data,  also decorrelates the signals and
isolates the bare soil reflectance, which is present for almost all
classes (cf. the data description in Section~\ref{sec:data}). 
For this reason, we also investigated a variant of our approach where, instead of working on the original spectral space, we used all the  principal components extracted from the original data (\textsc{AS-PCs} and \textsc{ASH-PCs}). In the \textsc{Indian Pines 1992} case, the increase in performance is striking, with a final Kappa of $0.89$. For the three other datasets, the results remain in the same range as for the \textsc{AS-bands} results.

\subsection{Multiclass selection}
\begin{figure}[!t]
\begin{tabular}{c}
\rotatebox{0}{\includegraphics[width=\linewidth]{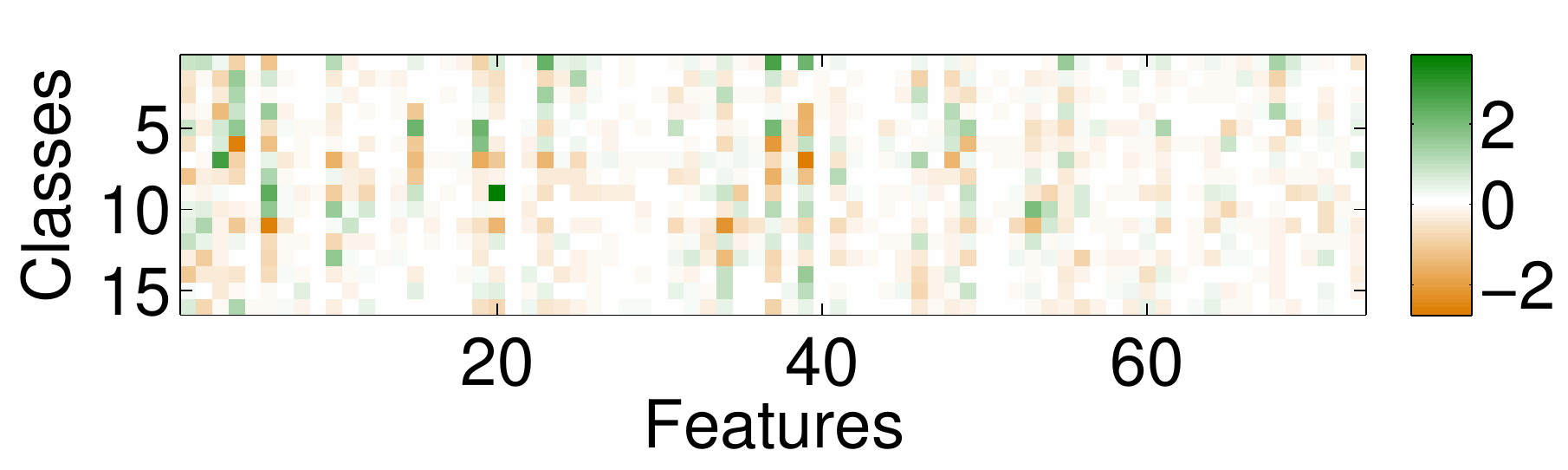}}\\
\rotatebox{0}{\includegraphics[width=\linewidth]{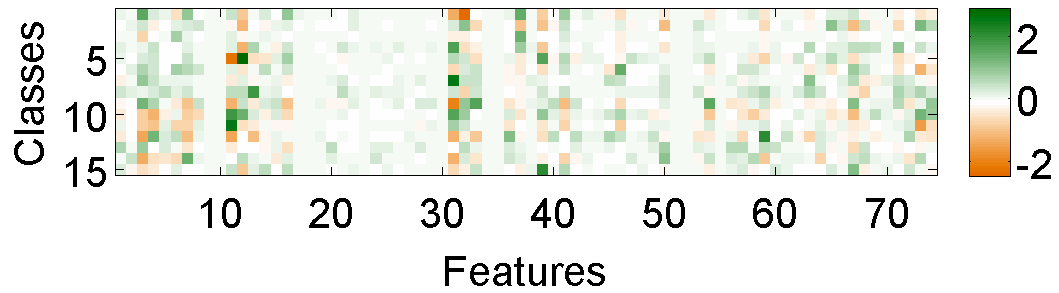}}\\
\end{tabular}
\caption{Final weight matrix for a run of the \textsc{Indian Pines 2010} (top) and \textsc{Houston 2013B} (bottom)  experiments.}
\label{fig:resW}
\end{figure}

For the four images, the active set models end up
with a maximum of $50-100$ features, shared by all classes. This model
is very compact, since it corresponds to only $30-50\%$ of the
initial dimensionality of the spectra. Due to the group-lasso regularization employed, the features selected are active for several
classes simultaneously, as shown in Fig.~\ref{fig:resW}, which
illustrates the  $\W^\top$ matrix for the \textsc{Indian Pines
  2010} and \textsc{Houston 2013B} experiments. The matrices correspond to those at the end of the feature learning, for one specific run of \textsc{AS-Bands} with $\lambda = 0.0001$. In both plots, each column corresponds to a feature selected by the proposed algorithm and each row to one class;  the color corresponds to the strength of the weight (positive or negative). One can  appreciate that the selected features (columns)  have large coefficients -- corresponding to strong green or brown tones in the figures -- for more than one class (the rows).

\subsection{Features visualization in \textsc{AS-Bands}}
\begin{figure*}[!t]
\setlength{\tabcolsep}{0pt}
\begin{tabular}{p{2.6cm}cccccccc}
		
				Class & \textcolor[rgb]{0.6745,0.4902,0.0431}{Soil} & 
		\textcolor[rgb]{0.0039,0.9412,0}{Tennis c.} & 
		\textcolor[rgb]{0.7843,0.0706,0.2196}{Run track} & 
		\textcolor[rgb]{0.7137,0.6824,0.3647}{Parking 2}  & 
		\textcolor[rgb]{0.4706,0,0.0039}{Residential} & 
		\textcolor[rgb]{0,0.4784,0.0039}{Str. grass} & 
		\textcolor[rgb]{0.4745,0.4745,0.4745}{Road} & 
		\textcolor[rgb]{0,0.7569,0.7765}{Water} \\
 \hline
 RGB &
 \includegraphics[width=1.85cm]{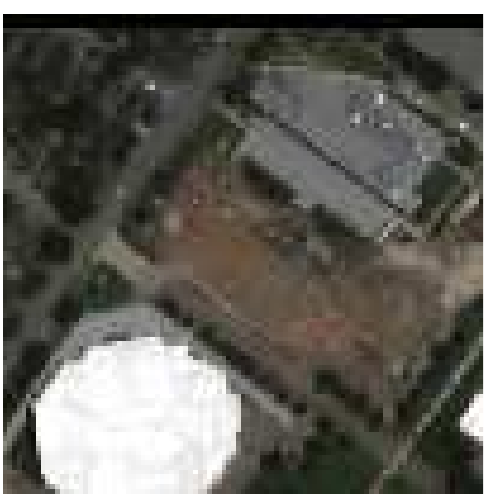}&
  \includegraphics[width=1.85cm]{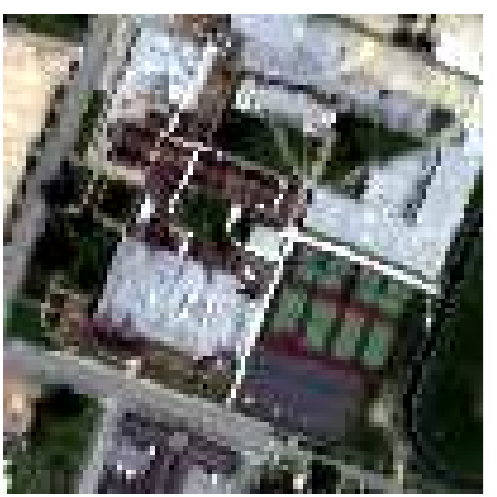}&
   \includegraphics[width=1.85cm]{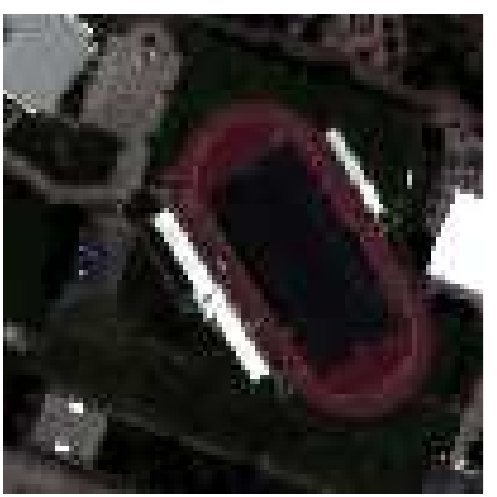}&
    \includegraphics[width=1.85cm]{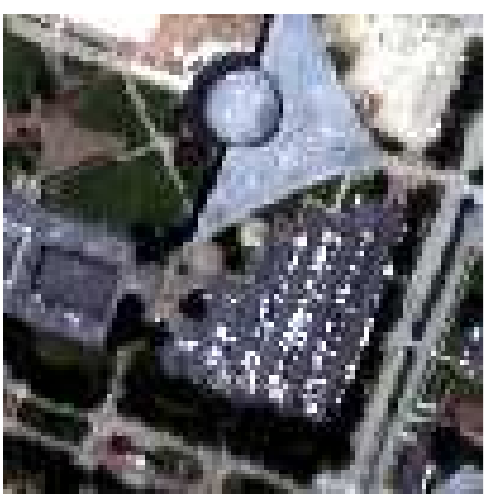}&
                \includegraphics[width=1.85cm]{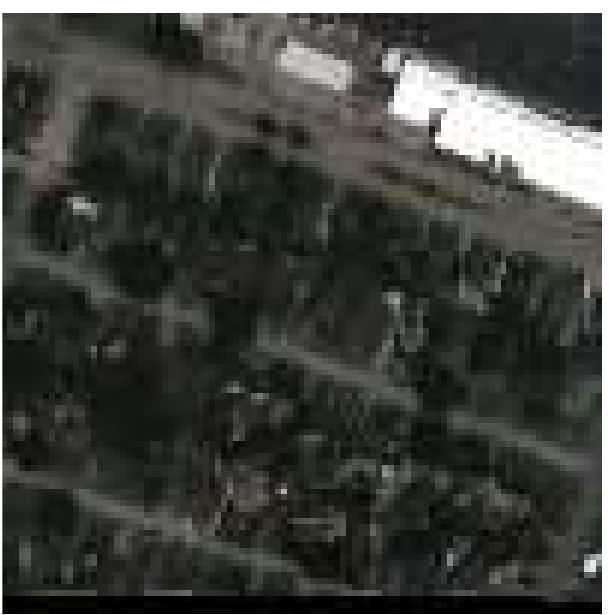}&
         \includegraphics[width=1.85cm]{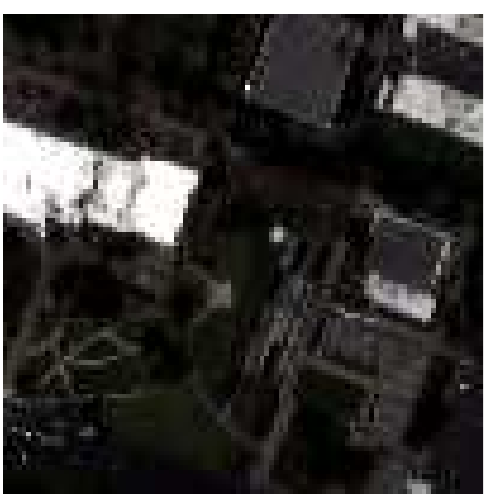}&
          \includegraphics[width=1.85cm]{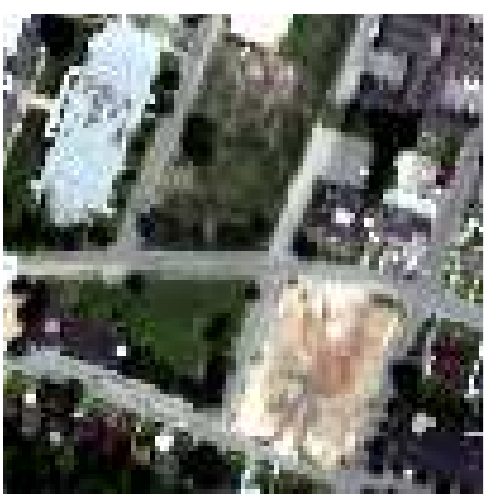}&
               \includegraphics[width=1.85cm]{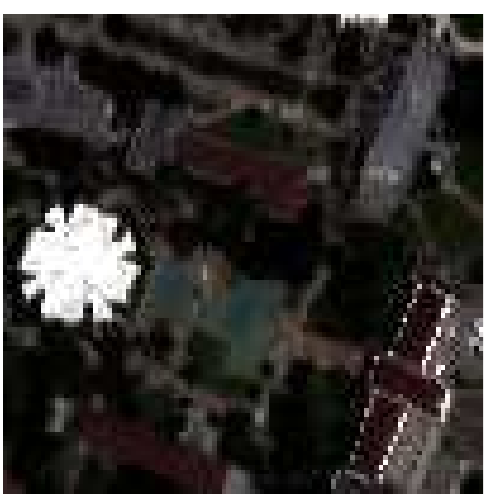}
          \\\hline
  GT &
 \includegraphics[width=1.85cm]{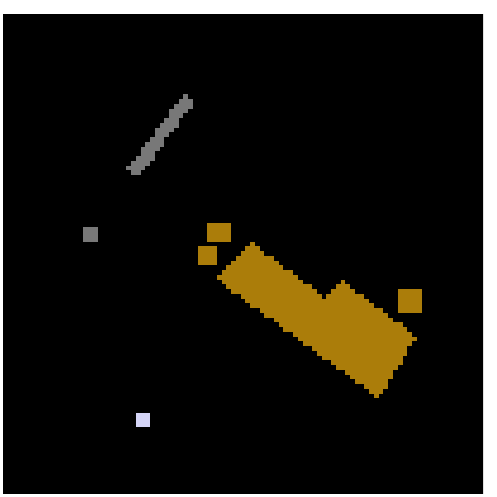}&
  \includegraphics[width=1.85cm]{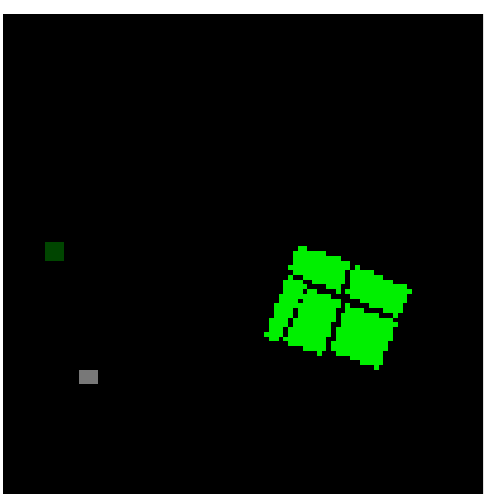}&
   \includegraphics[width=1.85cm]{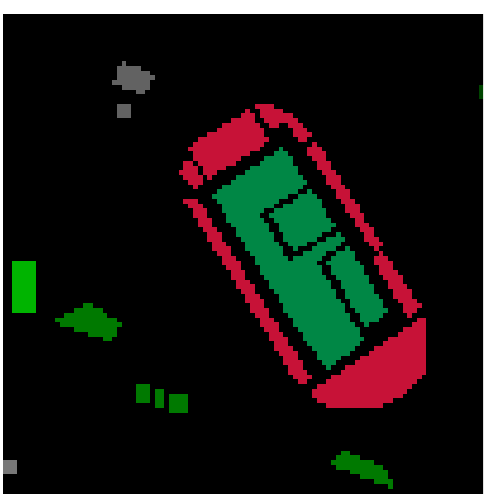}&
    \includegraphics[width=1.85cm]{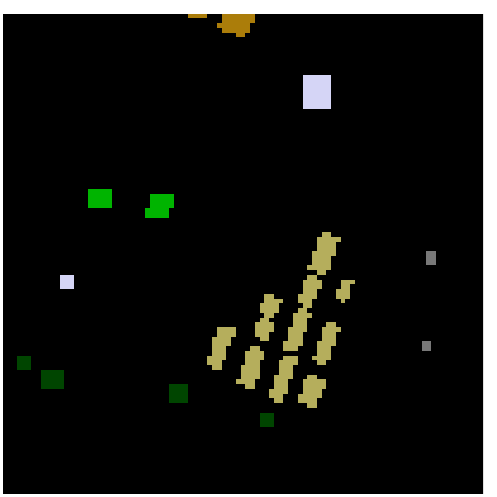}&
      \includegraphics[width=1.85cm]{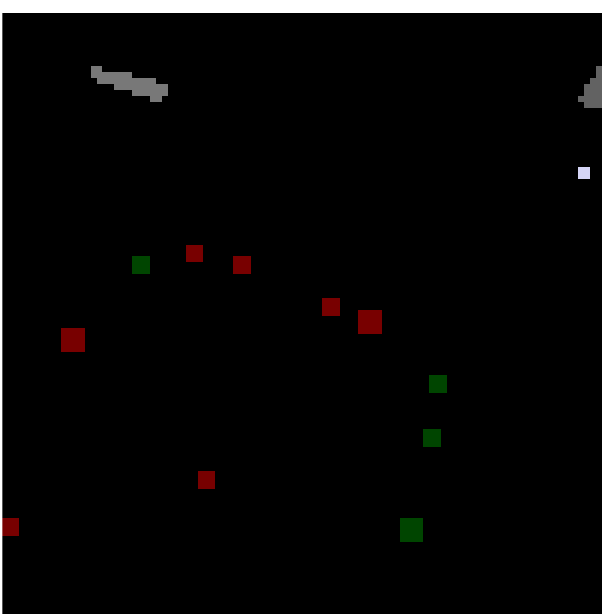}&
               \includegraphics[width=1.85cm]{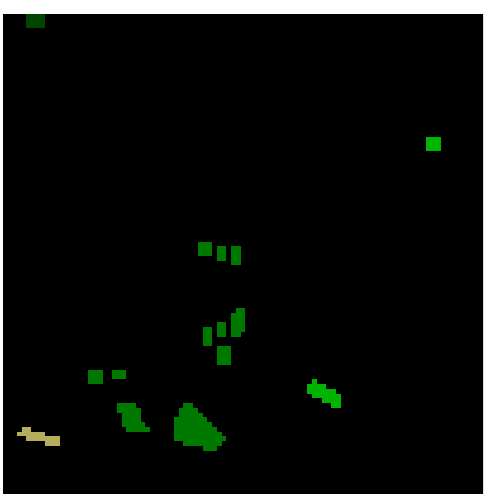}&         
          \includegraphics[width=1.85cm]{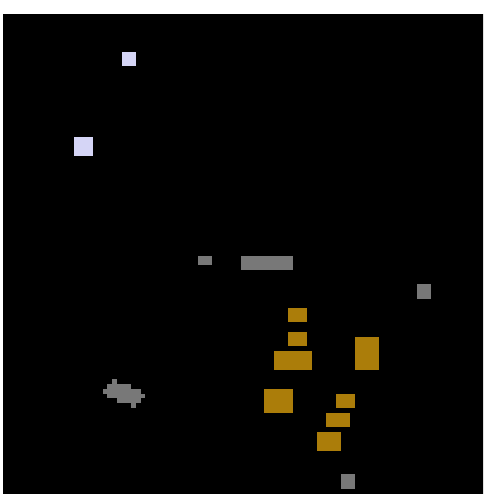}&
               \includegraphics[width=1.85cm]{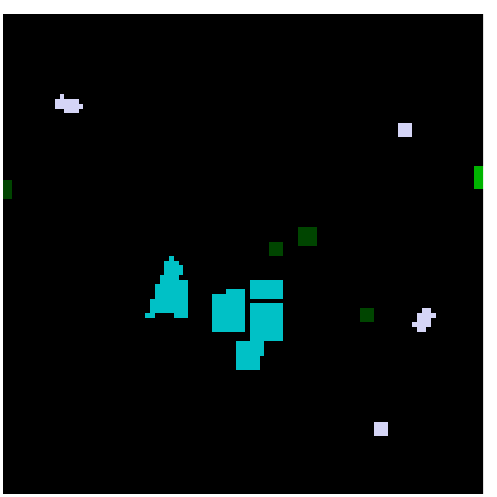}\\\hline
               
     \raisebox{.3cm}{\begin{tabular}{c}Classification\\(\textsc{AS-bands})\end{tabular}} &
 \includegraphics[width=1.85cm]{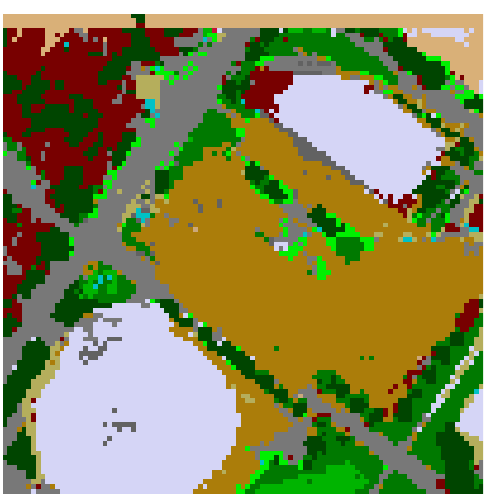}&
  \includegraphics[width=1.85cm]{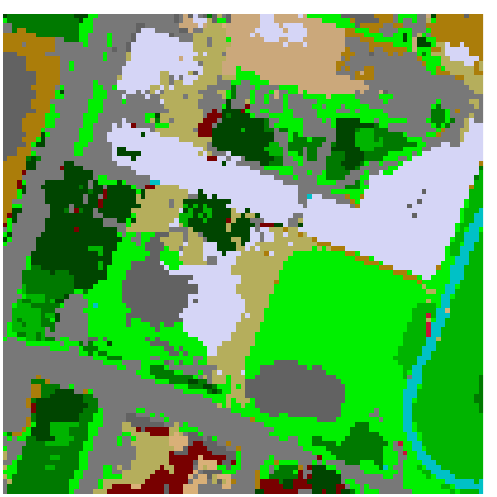}&
   \includegraphics[width=1.85cm]{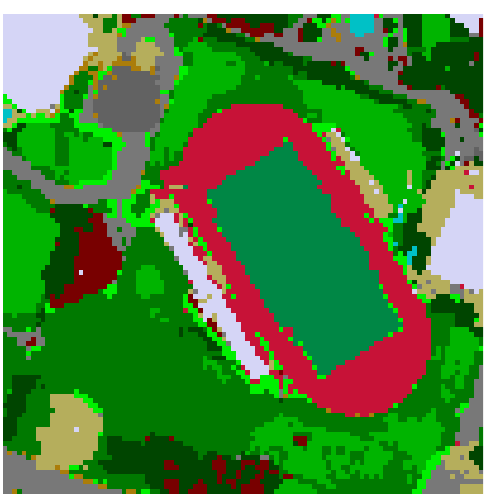}&
    \includegraphics[width=1.85cm]{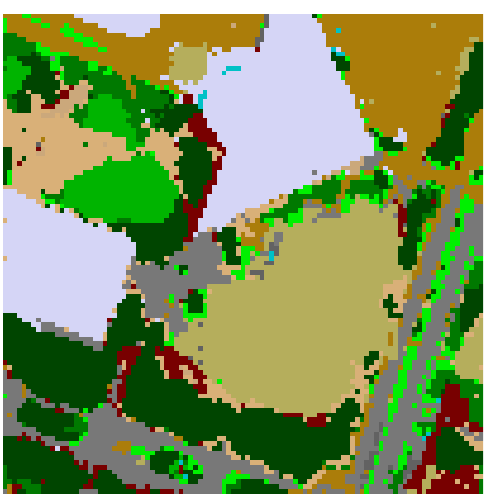}&
      \includegraphics[width=1.85cm]{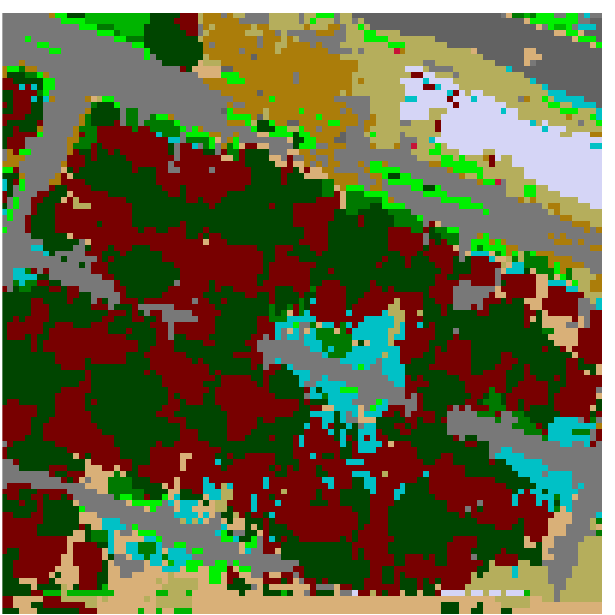}&
                    \includegraphics[width=1.85cm]{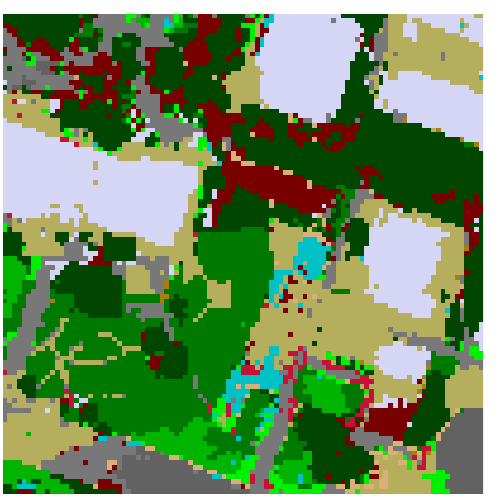}&
          \includegraphics[width=1.85cm]{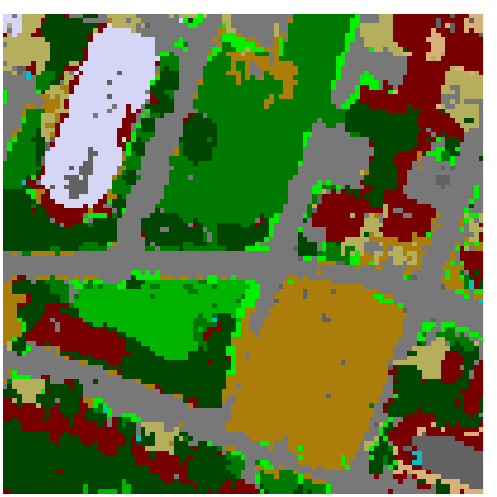}&
               \includegraphics[width=1.85cm]{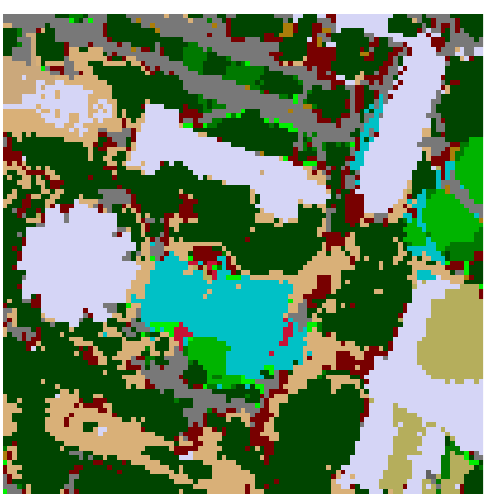}\\ 
     \hline
             \raisebox{.5cm}{{\scriptsize\begin{tabular}{l}
        Feature: 31\\ 
        Entropy, $15 \times 15$\\
         Band 145 (Lidar)\\
         Active in 11 classes
        \end{tabular}}}&
      \includegraphics[width=1.85cm]{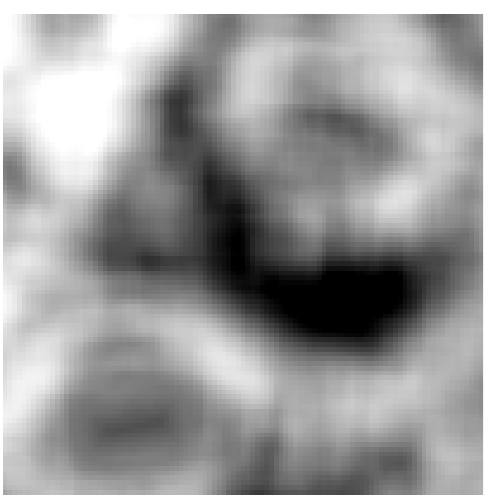}&
  \includegraphics[width=1.85cm]{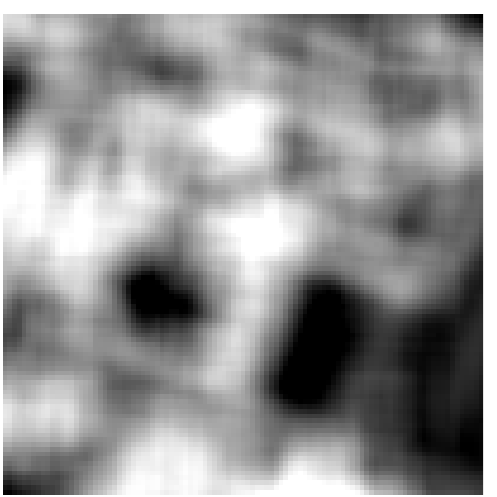}&
   \includegraphics[width=1.85cm]{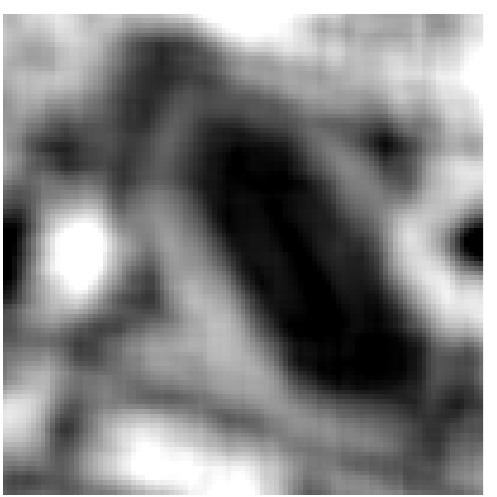}&
    \includegraphics[width=1.85cm]{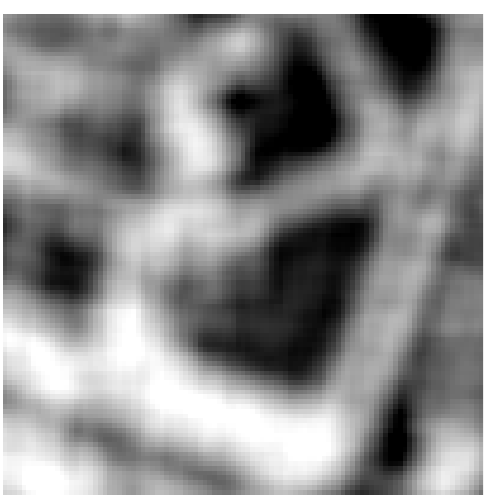}&
 \includegraphics[width=1.85cm]{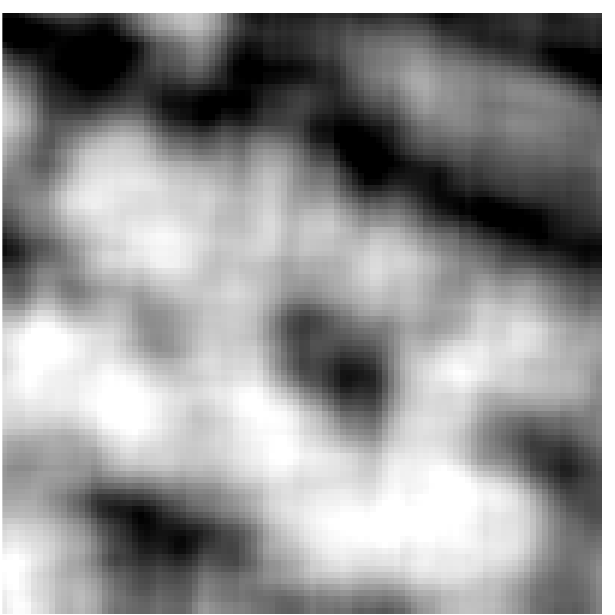}&
                 \includegraphics[width=1.85cm]{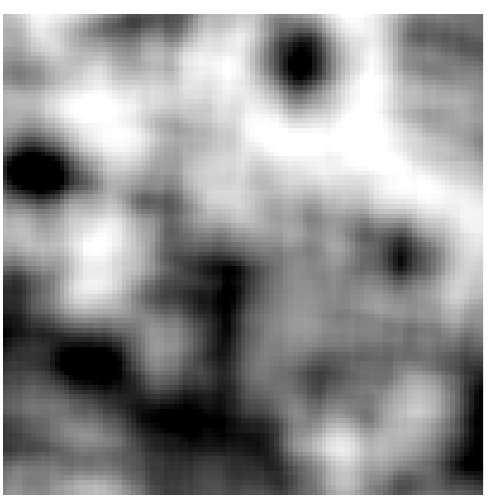}&
           \includegraphics[width=1.85cm]{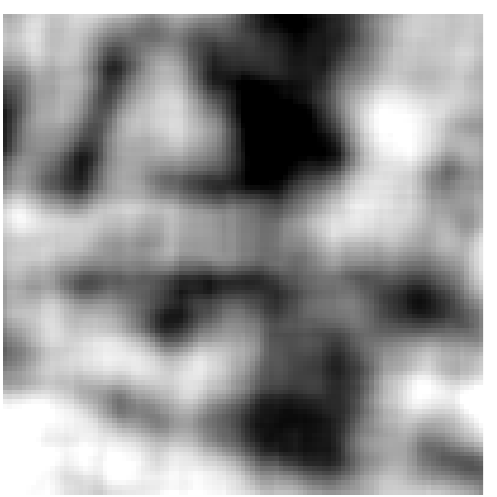}&
               \includegraphics[width=1.85cm]{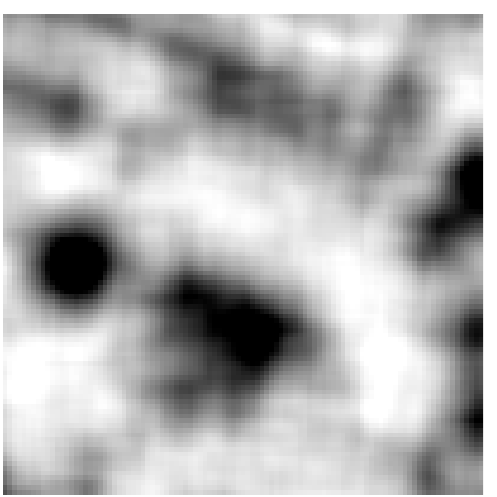}\\ 

               \raisebox{.5cm}{{\scriptsize\begin{tabular}{l}    
        Feature: 11\\ 
        Attribute area\\
         Band 145, 7010 pix.\\
         Active in 11 classes
        \end{tabular}}}&
      \includegraphics[width=1.85cm]{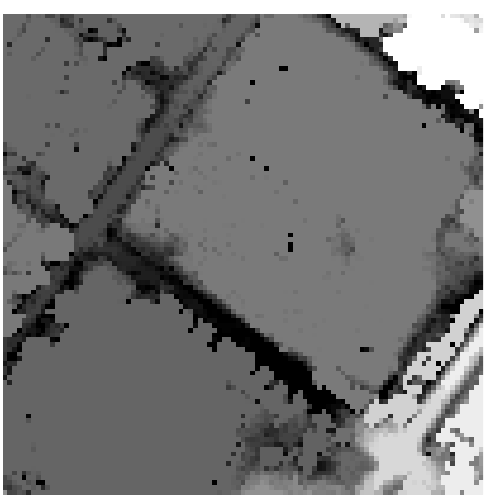}&
  \includegraphics[width=1.85cm]{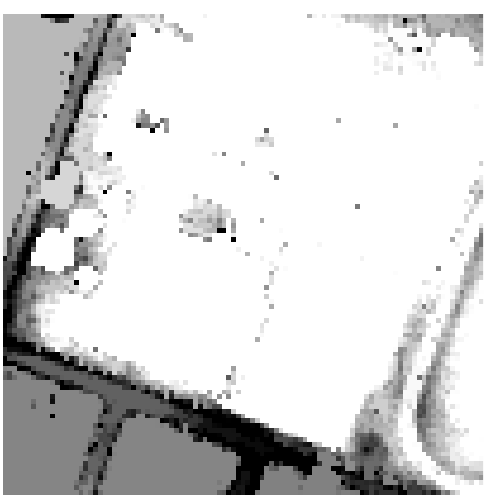}&
   \includegraphics[width=1.85cm]{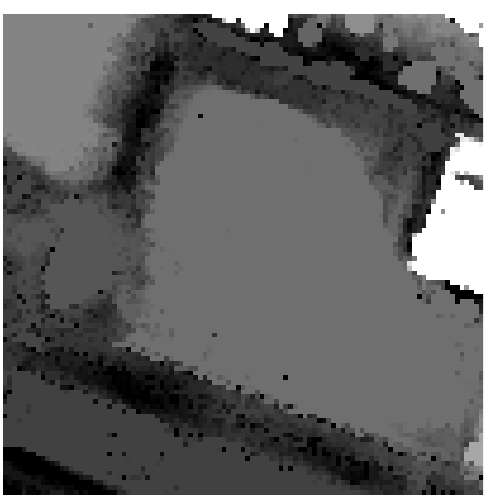}&
    \includegraphics[width=1.85cm]{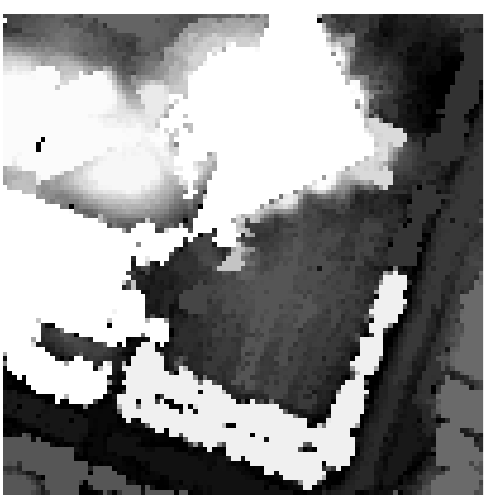}&
      \includegraphics[width=1.85cm]{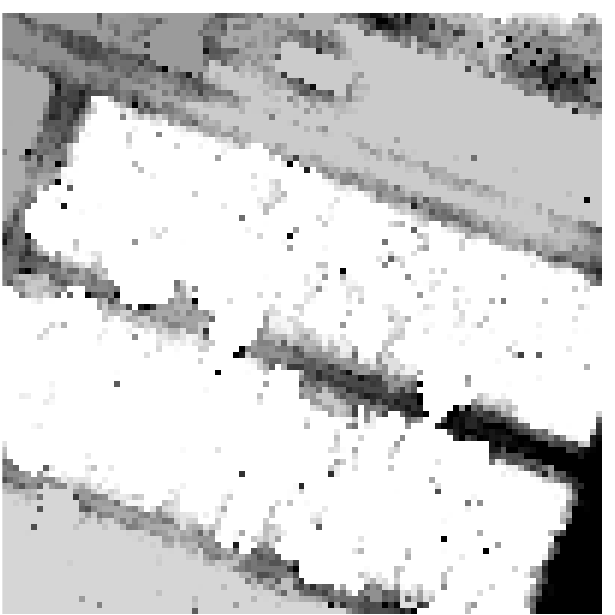}&
                    \includegraphics[width=1.85cm]{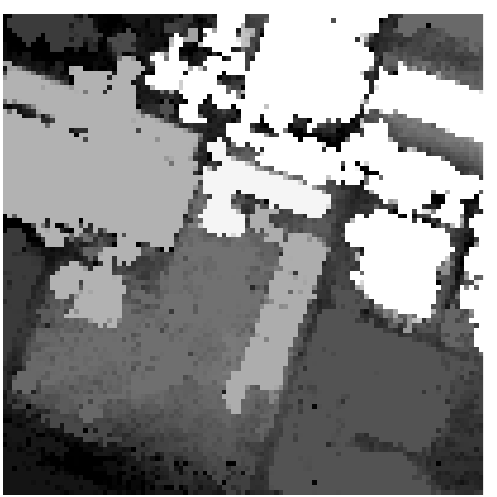}&
          \includegraphics[width=1.85cm]{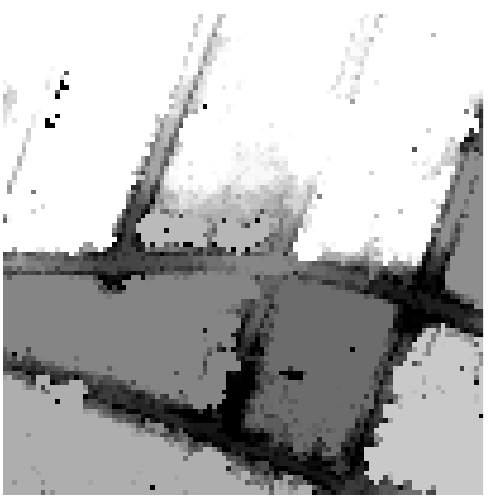}&
               \includegraphics[width=1.85cm]{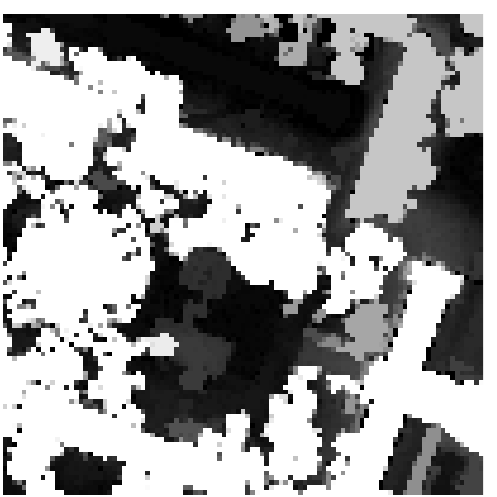}\\ 
     
  \raisebox{.5cm}{{\scriptsize\begin{tabular}{l}
        Feature: 12\\ 
     Attribute area\\
         Band 68, 2010 pix.\\
         Active in 12 classes
        \end{tabular}}}&
      \includegraphics[width=1.85cm]{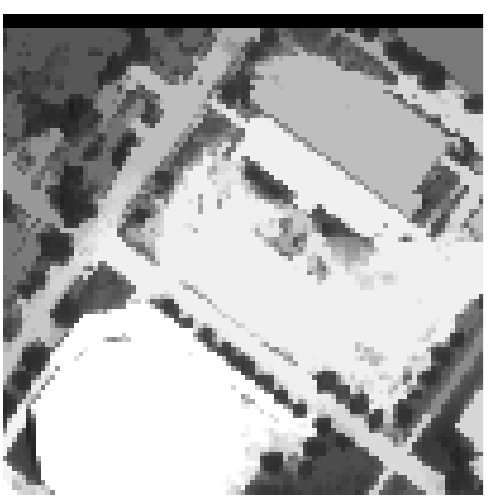}&
  \includegraphics[width=1.85cm]{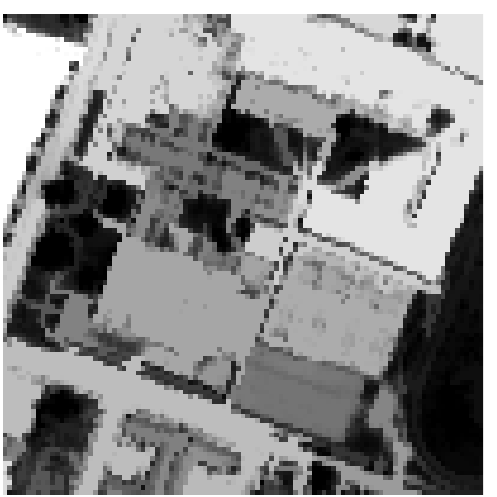}&
   \includegraphics[width=1.85cm]{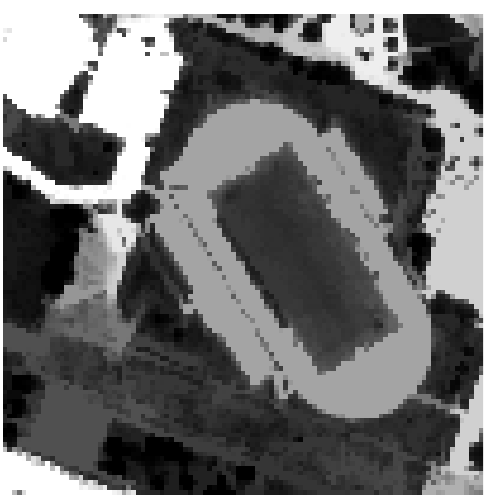}&
    \includegraphics[width=1.85cm]{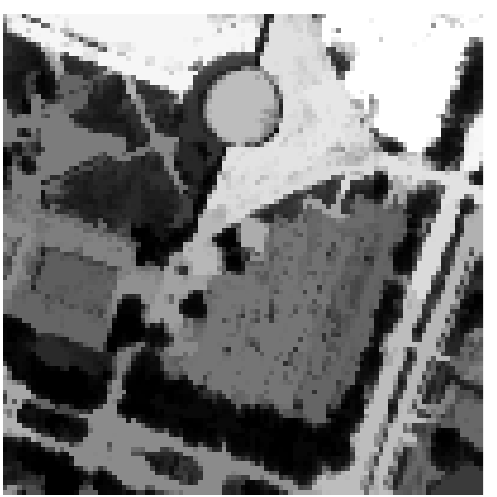}&
     \includegraphics[width=1.85cm]{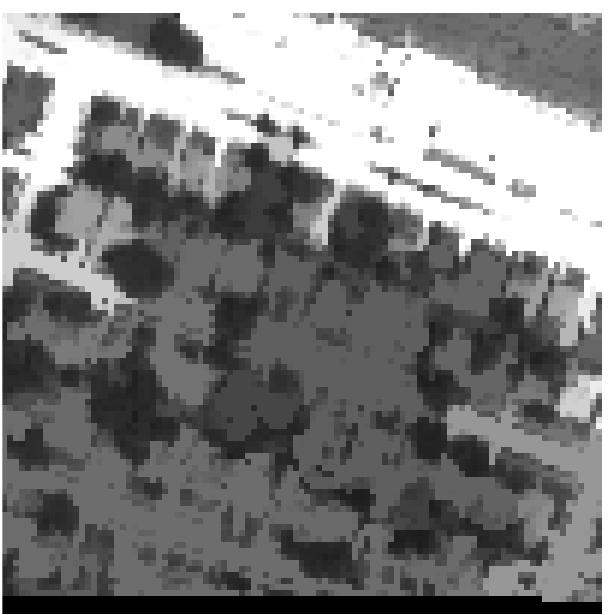}&
                   \includegraphics[width=1.85cm]{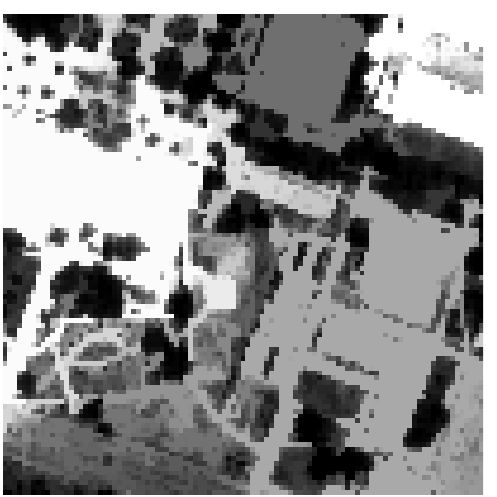}&
          \includegraphics[width=1.85cm]{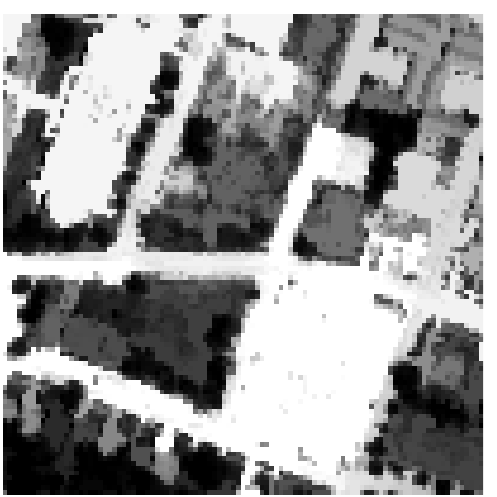}&
               \includegraphics[width=1.85cm]{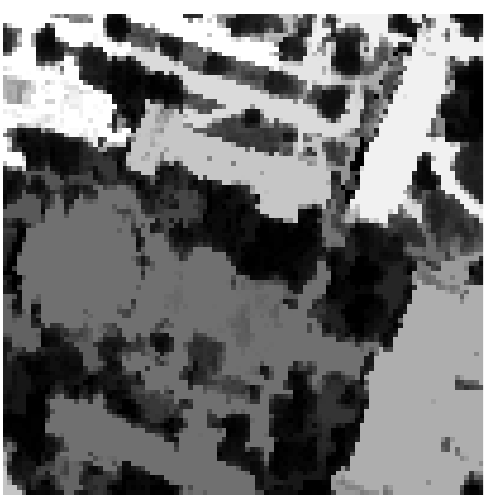}\\  
  \raisebox{.5cm}{{\scriptsize\begin{tabular}{l}
        Feature: 3\\ 
     Closing, diamond\\
         Band 110, $7 \times 7$\\
         Active in 11 classes
        \end{tabular}}}&
      \includegraphics[width=1.85cm]{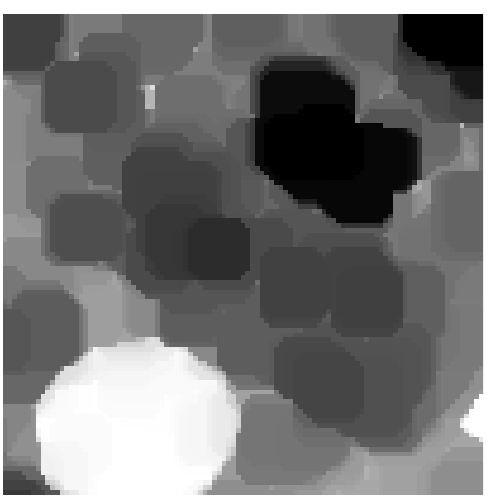}&
  \includegraphics[width=1.85cm]{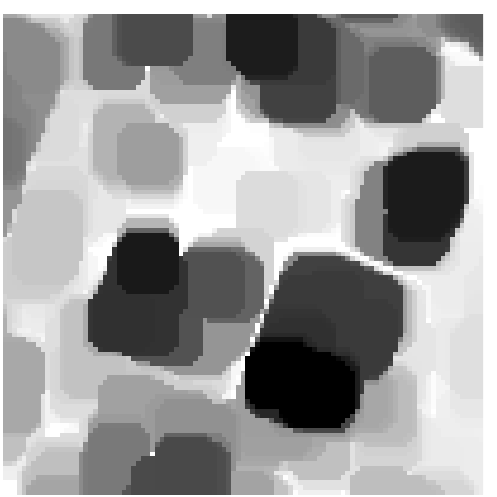}&
   \includegraphics[width=1.85cm]{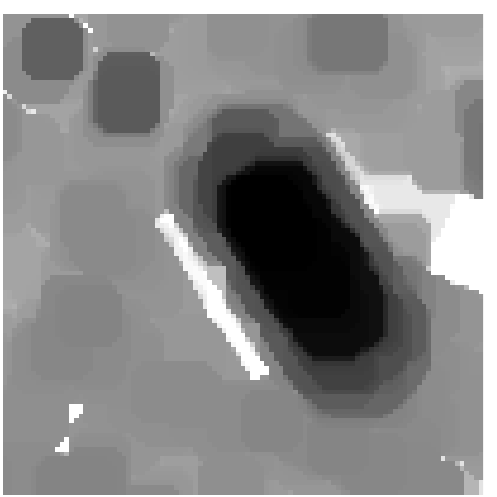}&
    \includegraphics[width=1.85cm]{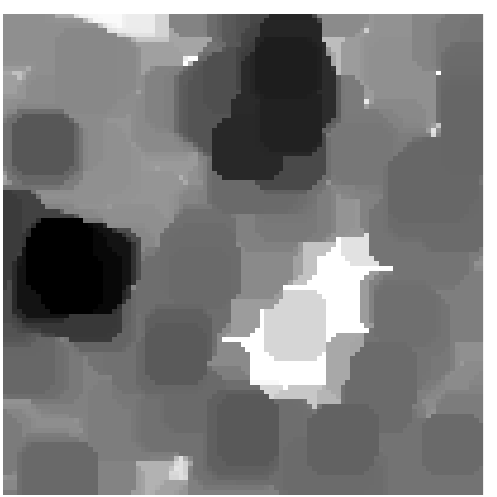}&
     \includegraphics[width=1.85cm]{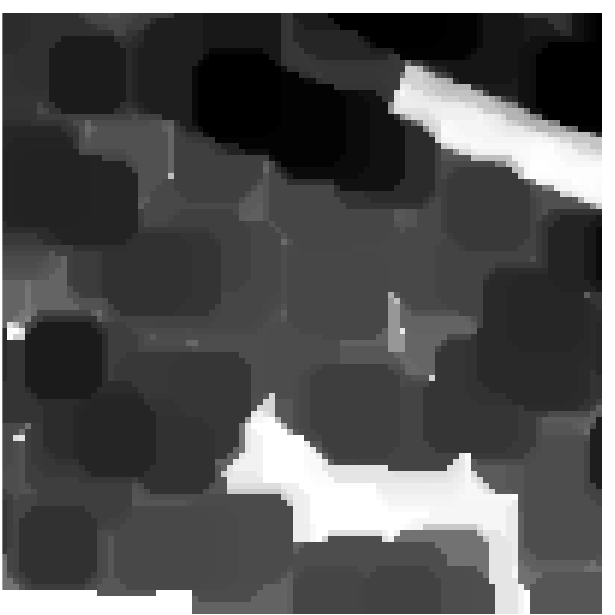}&
                     \includegraphics[width=1.85cm]{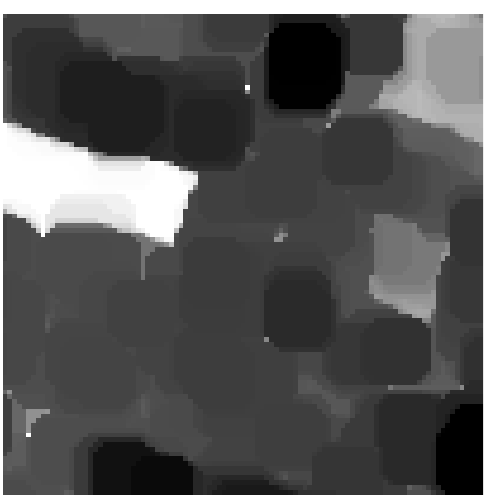}&
          \includegraphics[width=1.85cm]{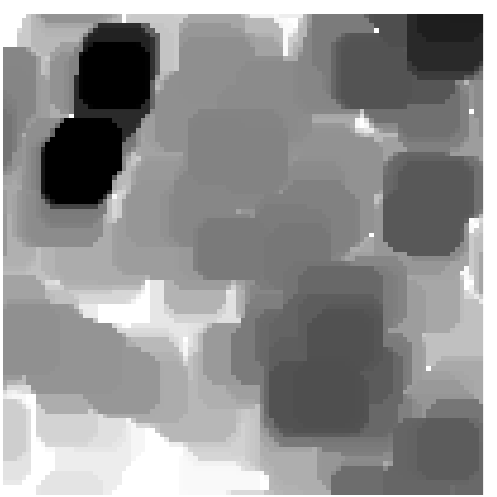}&
               \includegraphics[width=1.85cm]{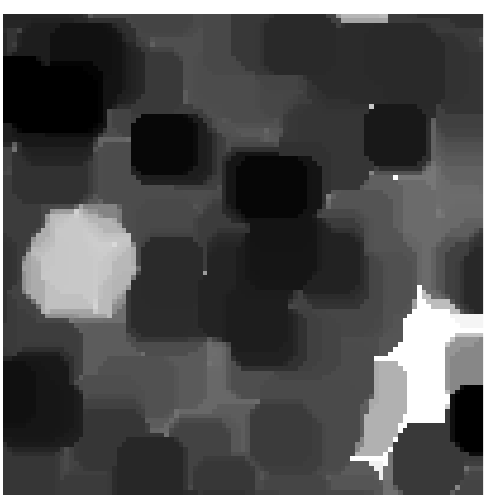}\\  
                 \raisebox{.5cm}{{\scriptsize\begin{tabular}{l}
        Feature: 46\\ 
     Closing rec. top hat\\
         Band 106, $15 \times 15$\\
         Active in 5 classes
        \end{tabular}}}&
      \includegraphics[width=1.85cm]{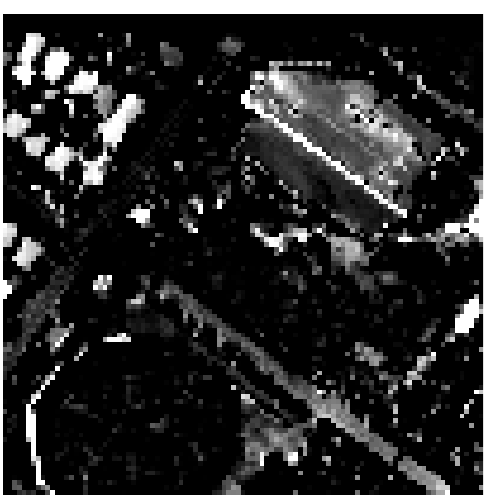}&
  \includegraphics[width=1.85cm]{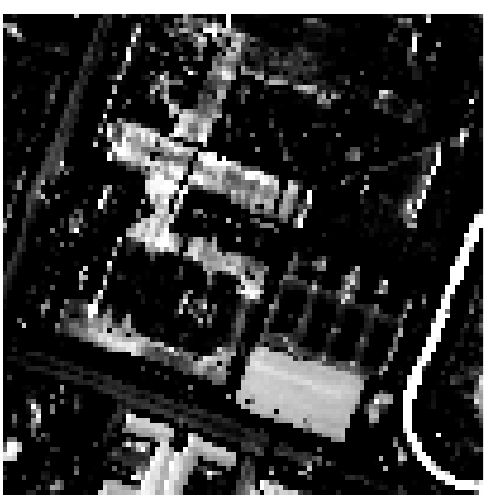}&
   \includegraphics[width=1.85cm]{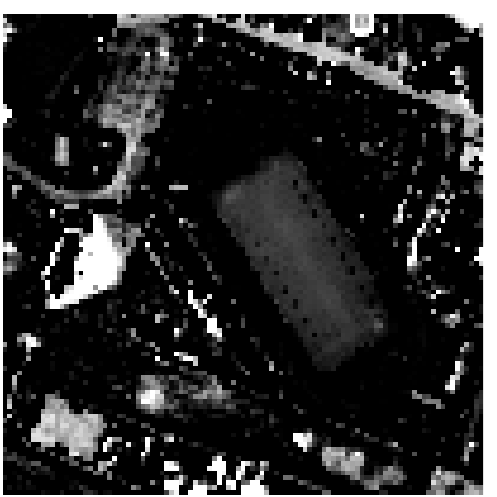}&
    \includegraphics[width=1.85cm]{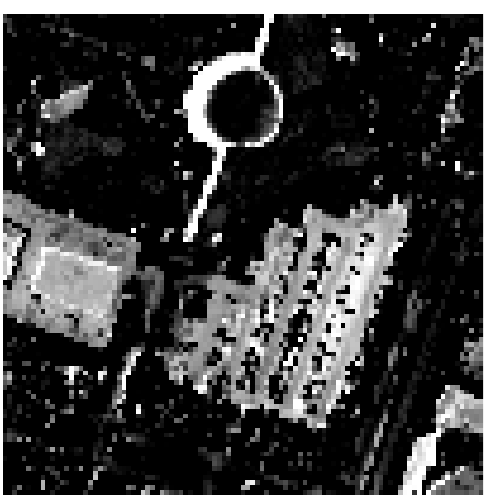}&
    \includegraphics[width=1.85cm]{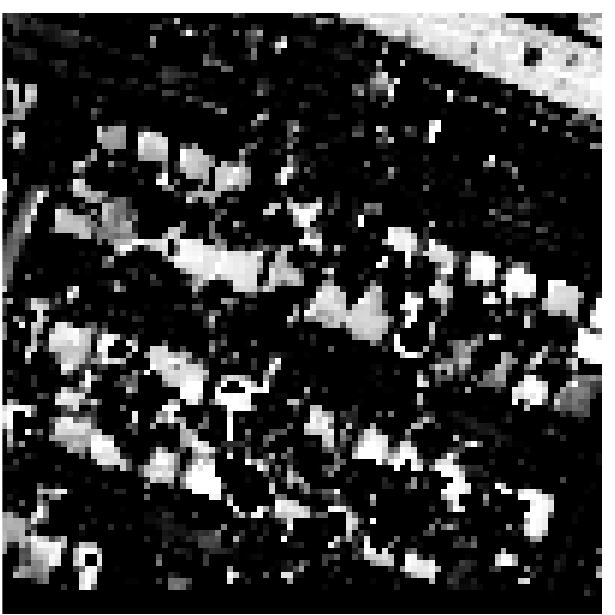}&
                   \includegraphics[width=1.85cm]{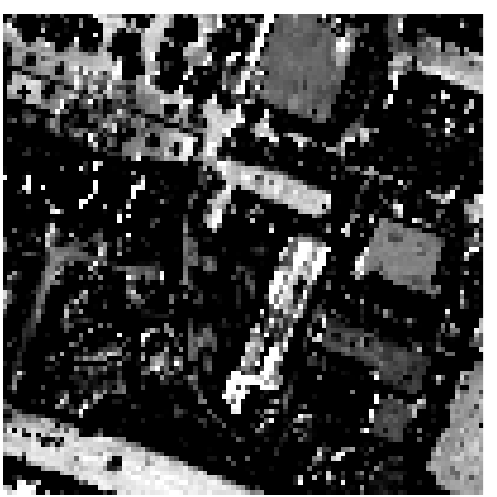}&
          \includegraphics[width=1.85cm]{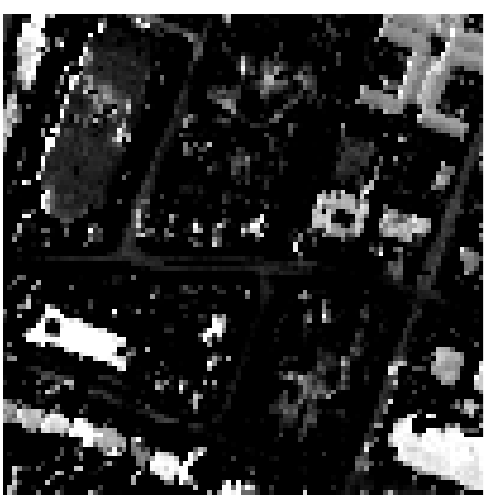}&
                            \includegraphics[width=1.85cm]{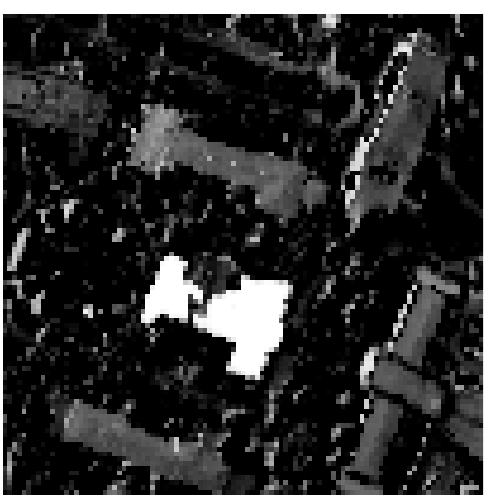}\\  
  \end{tabular}

  \caption{Visualization of the features with highest $||W_{j,\cdot}||_2$ for one run of the \textsc{Houston 2013B} results (cf. bottom matrix of Fig.~\ref{fig:resW}). First row: RGB subsets; second row: ground truth; third row: output of the classification with the proposed approach; fourth row to end: visualization of the six features with highest squared weights.}
    \label{fig:elsfeatures}
\end{figure*}

Figure~\ref{fig:elsfeatures} illustrates some of the features
selected by \textsc{AS-Bands} in the \textsc{Houston 2013B} case. Each column corresponds
to a different zoom in the area and highlights a specific class. We
visualized the features of the same run as the bottom row of
Fig.~\ref{fig:resW} and visualized the six features with highest
$||W_{j,\cdot}||_2$, corresponding to those active for most classes
with the highest squared weights. By analysis of the features learned,
one can appreciate that they clearly are discriminative for the specific
classification problem: this shows that, by decreasing the overall
loss, adding these features to the active set really improves
class discrimination.

\begin{figure}[!t]
\begin{tabular}{cc}
\includegraphics[width=.45\linewidth]{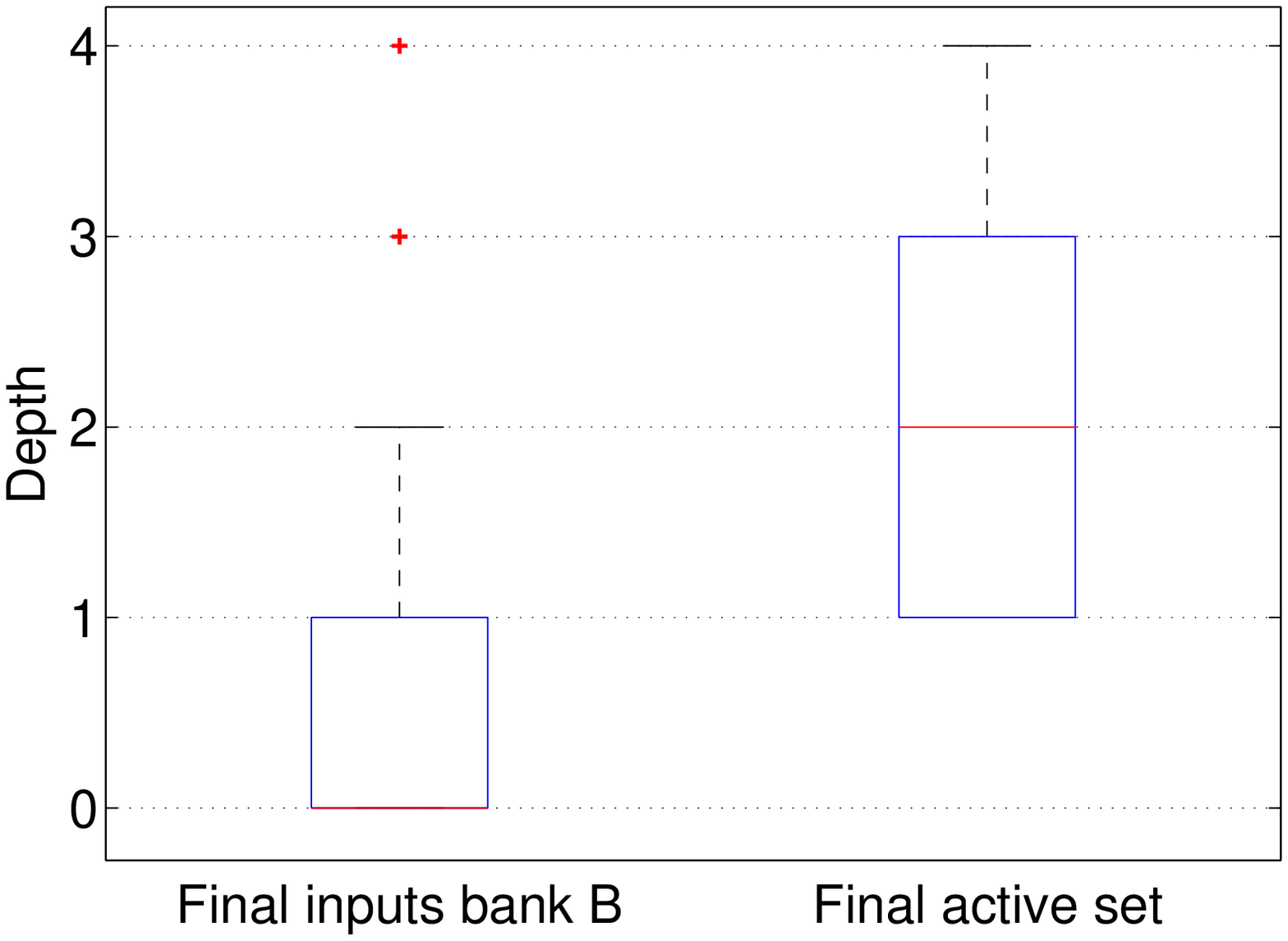}&
\includegraphics[width=.45\linewidth]{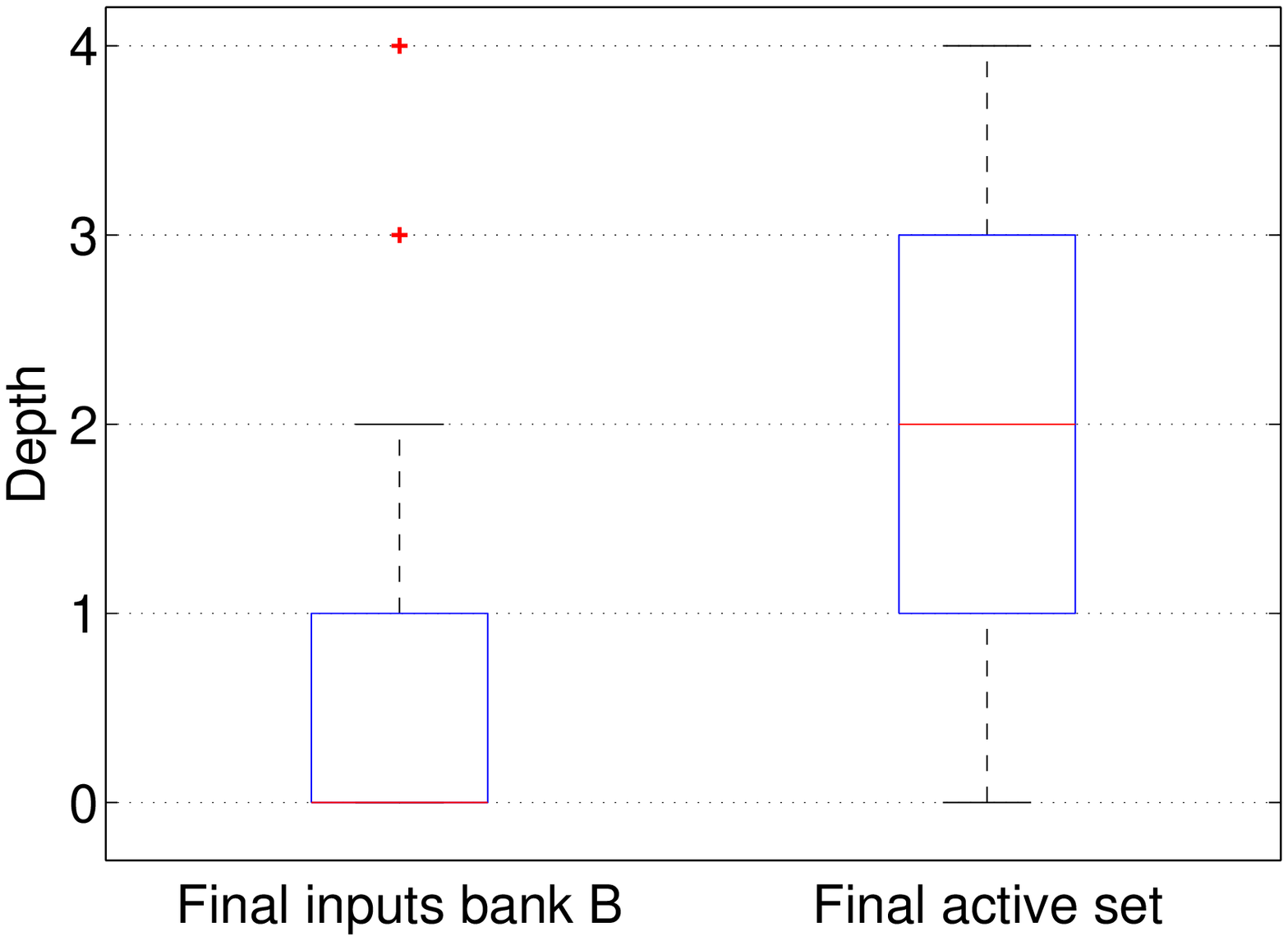}\\
\includegraphics[width=.45\linewidth]{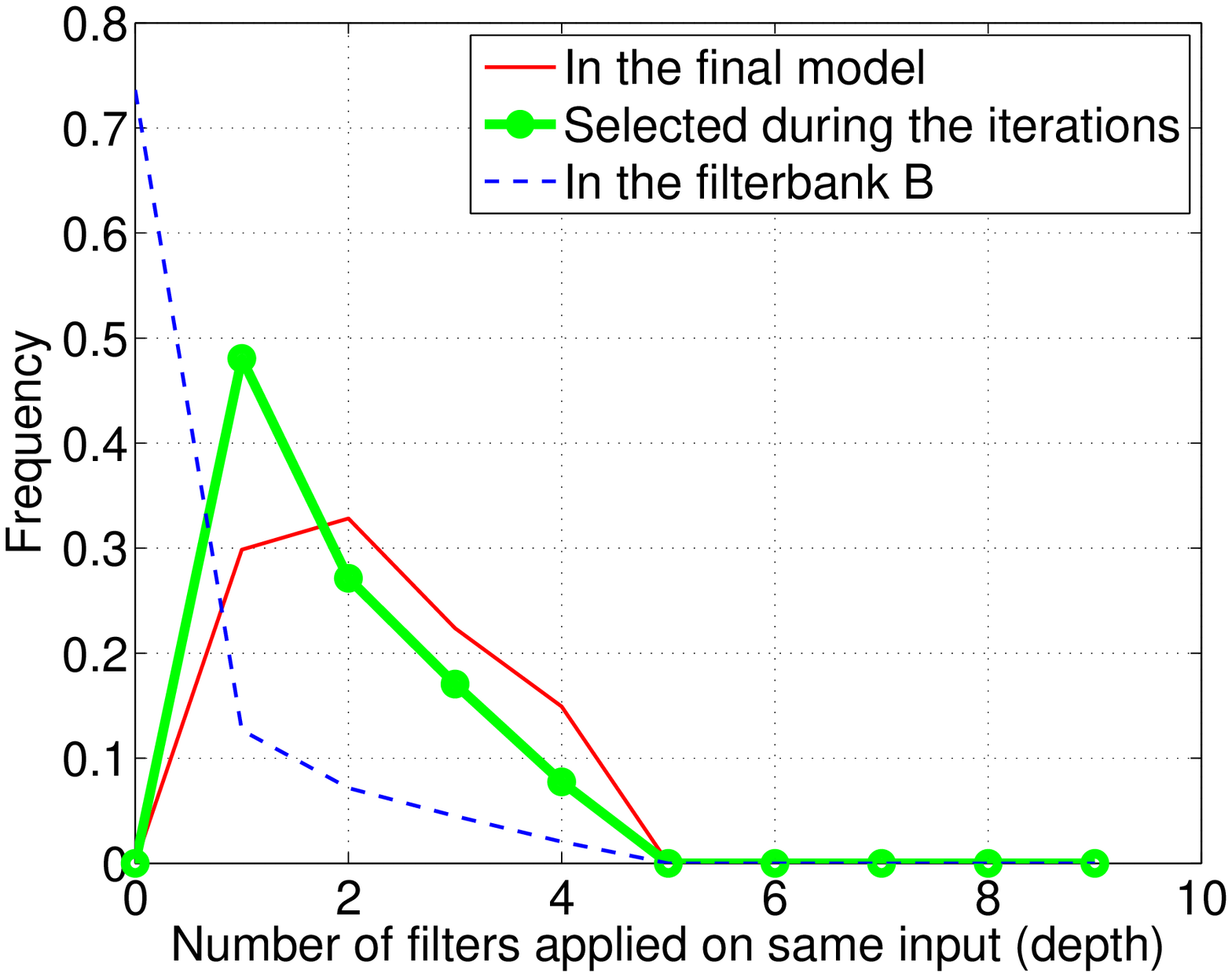}&
\includegraphics[width=.45\linewidth]{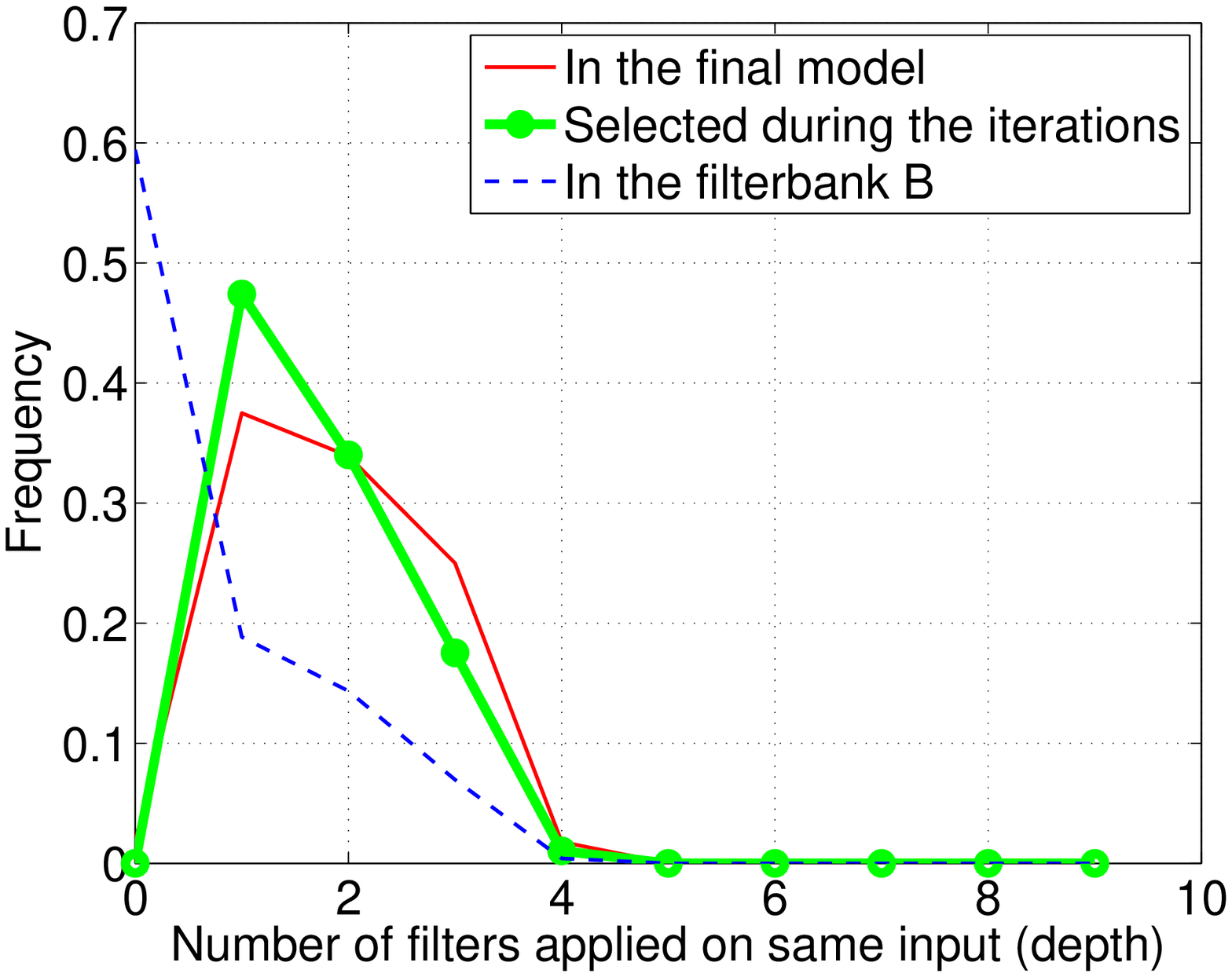}\\
(a) \textsc{Pines 2010} & (b) \textsc{Houston 2013A}\\
\end{tabular}
\caption{Analysis of the depth of the features in the final active set of one run of the \textsc{ASH-Bands} and $\lambda = 0.001$.}
\label{fig:depth}
\end{figure}

\subsection{Role of the features issued from the hierarchical model \textsc{ASH-Bands}}

Finally, we study in detail the hierarchical features that have
  been discovered by our method. First, we 
 discuss the distribution of the depth of features
in the active set in the \textsc{ASH-Bands} model. Top row of
Fig.~\ref{fig:depth} shows the distribution of the weights of the
features in both the inputs bank $B$ and in the active set $\varphi$ at the
end of the feature learning. Regarding the final bank $B$, which
contains $489$ features in the \textsc{Indian Pines 2010} and $244$ in
the \textsc{Houston 2013A} case, most of the features are of depth
$0$ (the original features), $1$ and $2$. But if we consider the final
active set $\varphi$, of size $67$ (\textsc{Indian Pines 2010}) and
$56$ (\textsc{Houston 2013A}), we see that the median depth is of $2$
in both cases: this means that no features of depth $0$ (no original
features) are kept in the final active set. The only exception is
provided by the LiDAR data in the \textsc{Houston 2013A} dataset,
which is kept in the final active set. These observations are
confirmed by the distributions illustrated in the bottom row of
Fig.~\ref{fig:depth}: the distribution of depths in the final bank $B$
(blue dashed line) has 60-70\% of features of depth $0$, while the
distribution of the features selected during the iterations (green
 line with circle markers) shows an average more towards a depth of $2$. The
features in the final active set $\varphi$ (red line) show a
distribution even more skewed towards higher depth levels, showing
that features of low depth (typically depths of $1$) are first added to $\varphi$ and then
replaced by features with higher depth issued from them. 

\begin{figure*}
\begin{tabular}{cl|cl}
\hline
\multicolumn{4}{c}{Four depth levels}\\
&\multicolumn{2}{c}{\includegraphics[width=.23\linewidth]{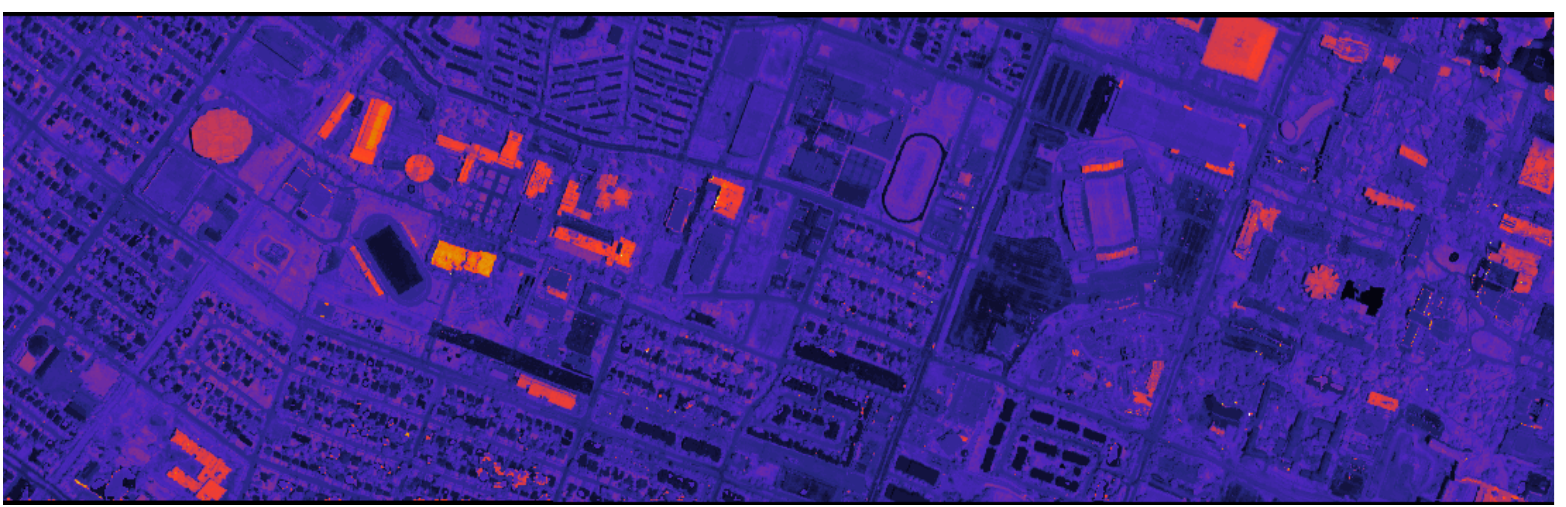} 
 \raisebox{.5cm}{{\scriptsize\begin{tabular}{p{3cm}}
        Depth 0\\
  	 Original band\\
	 Band 105 \\
        \end{tabular}}} 
        }
         \\
  \multicolumn{4}{c}{$\downarrow$}\\
  &  \multicolumn{2}{c}{\includegraphics[width=.23\linewidth]{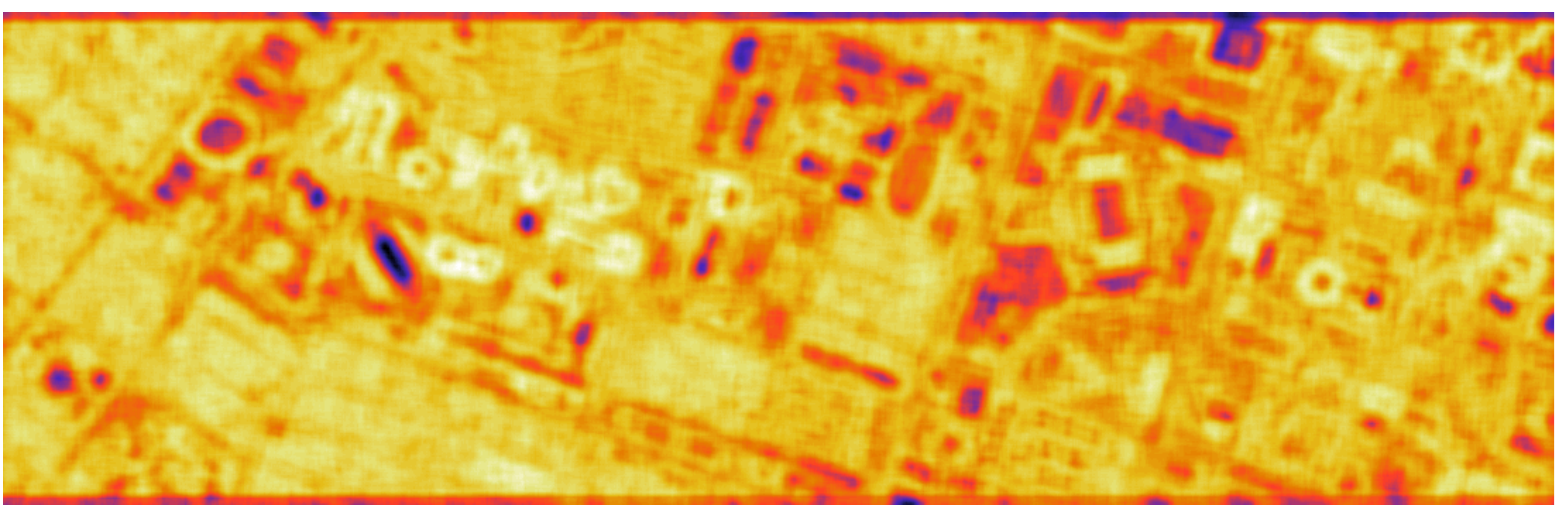} 
 \raisebox{.7cm}{\fbox{{\scriptsize\begin{tabular}{p{3cm}}
	\textbf{In the active set}\\
	Depth 1\\
  	Entropy filter\\
	 Local window, 15 pixels \\
        \end{tabular}}} }
        }
         \\
\multicolumn{4}{c}{$\swarrow \qquad \searrow$}\\
    \includegraphics[width=.23\linewidth]{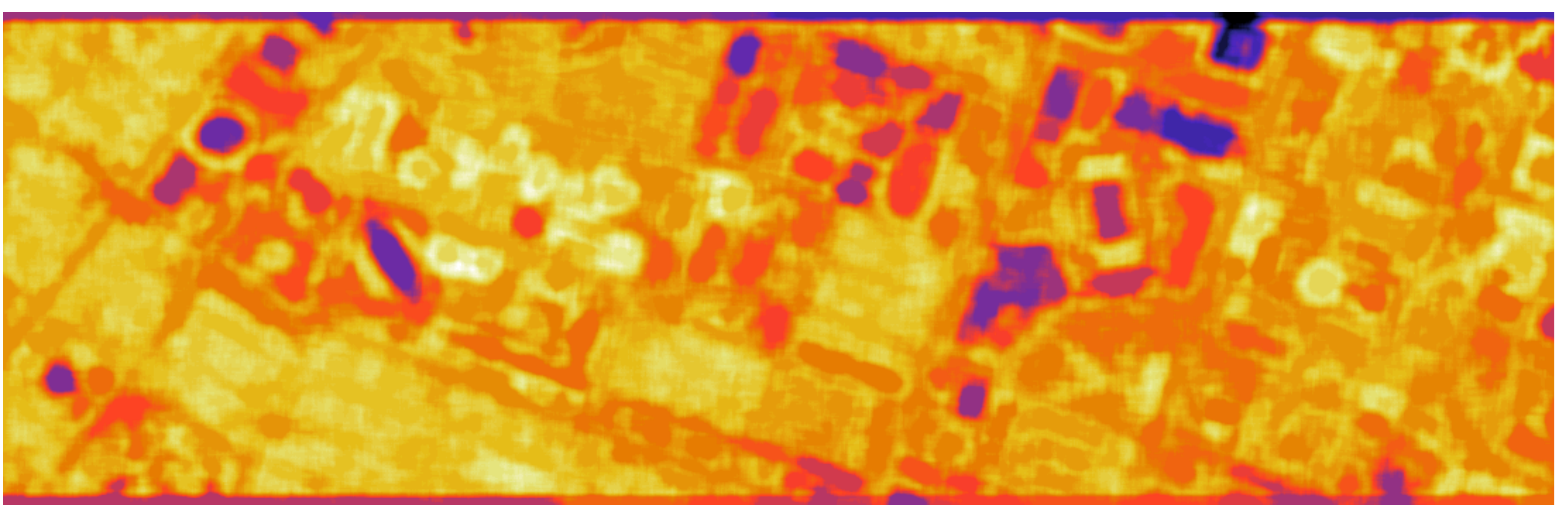} &
\raisebox{.7cm}{{\scriptsize\begin{tabular}{p{3cm}}
      	Depth 2\\
 	Closing by reconstruction\\
	Square SE, 13 pixels\\
        \end{tabular}}} 
&
  \includegraphics[width=.23\linewidth]{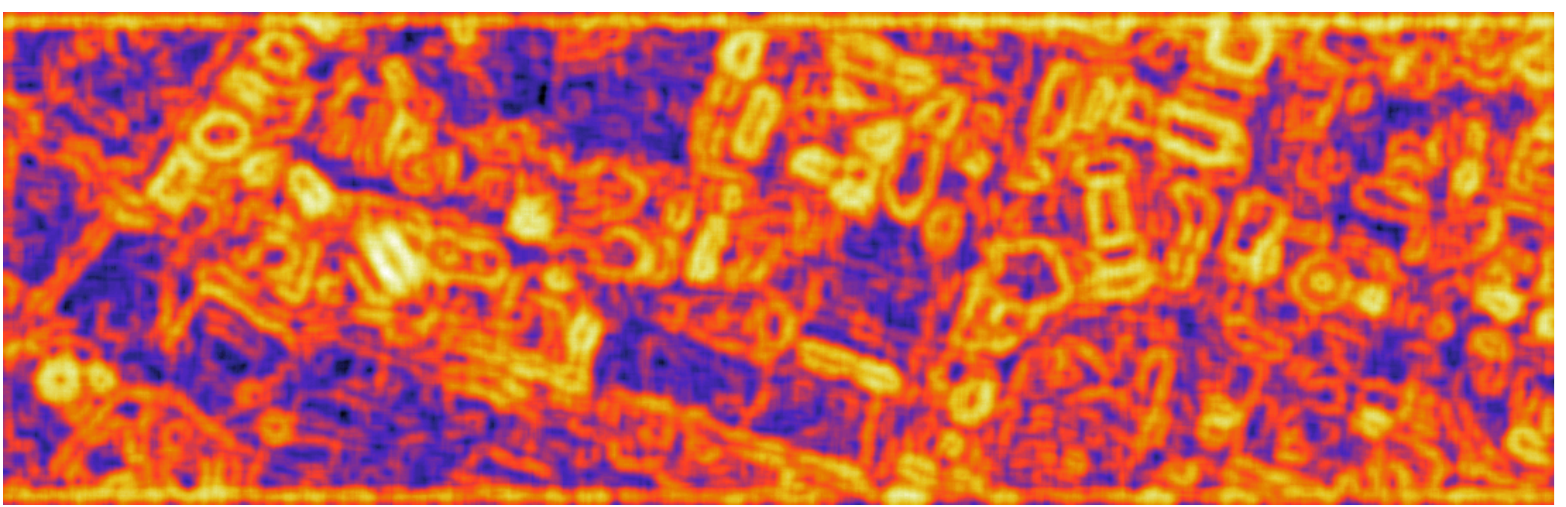} &
 \raisebox{.7cm}{{\scriptsize\begin{tabular}{p{3cm}}
      	Depth 2\\
 	Entropy filter\\
	 Local window 15 pixels \\

        \end{tabular}}} 
        \\
  $\downarrow$ & {} &  $\downarrow$ \\      

  \includegraphics[width=.23\linewidth]{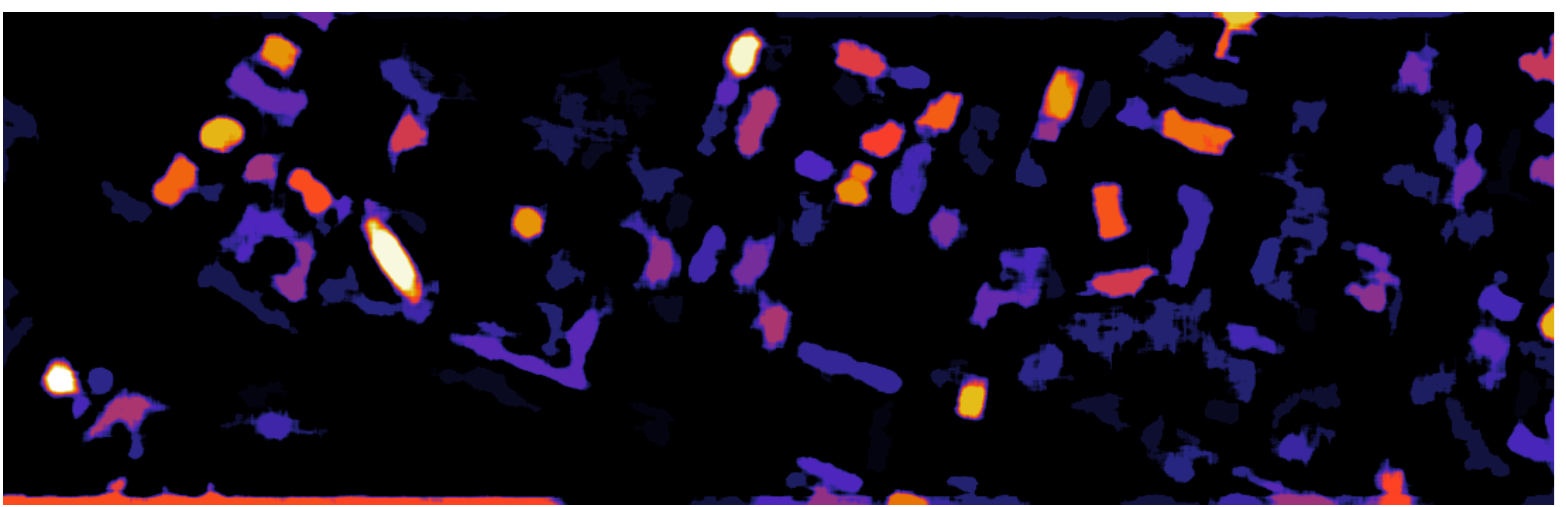} &
\raisebox{.8cm}{\fbox{{\scriptsize\begin{tabular}{p{3cm}}
      	\textbf{In the active set} \\
	Depth 3\\
 	Closing by reconstruction top-hat\\
	Square SE, 21 pixels\\
        \end{tabular}}} }
 &
         \includegraphics[width=.23\linewidth]{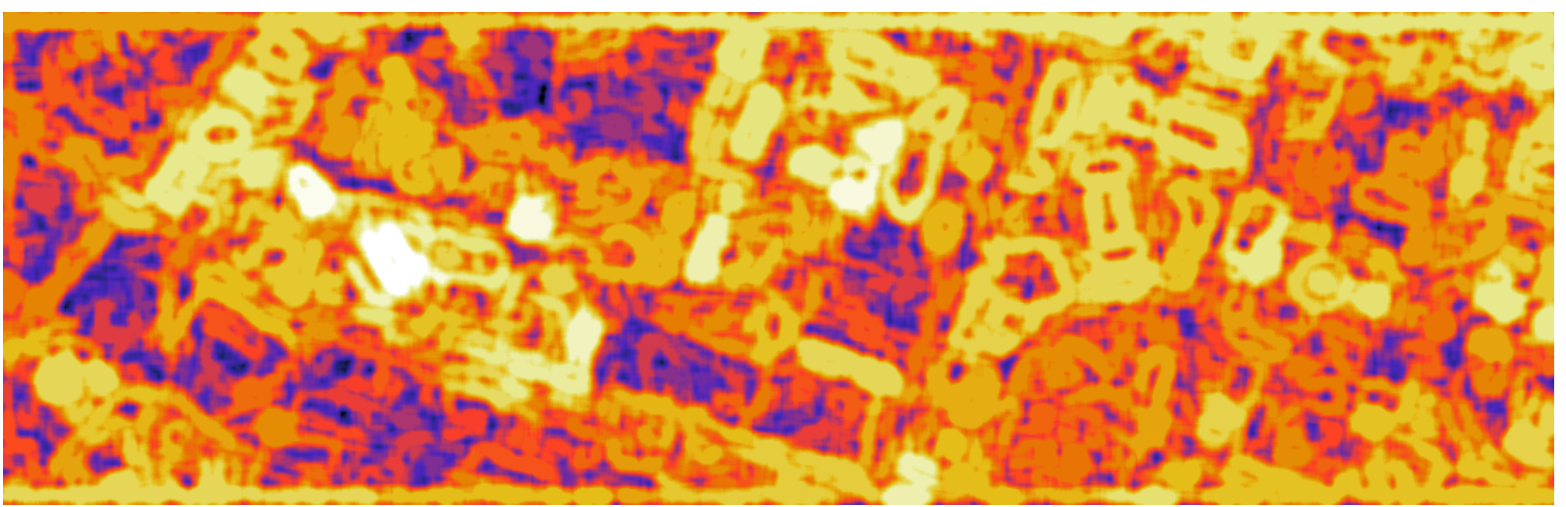} &
\raisebox{.8cm}{\fbox{{\scriptsize\begin{tabular}{p{3cm}}
        	\textbf{In the active set} \\
	Depth 3\\
 	Opening by reconstruction\\
	Square SE, 17 pixels\\
        \end{tabular}}} }
        \\

\hline
\hline
\multicolumn{4}{c}{Three depth levels}\\
\includegraphics[width=.23\linewidth]{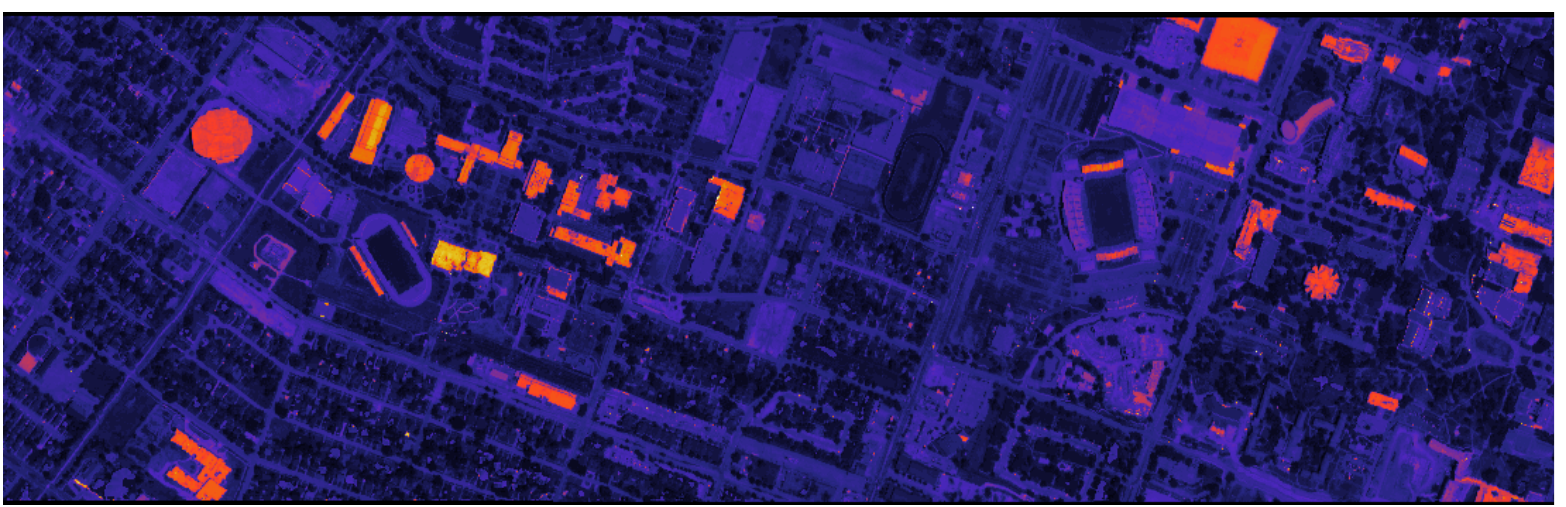} &
 \raisebox{1cm}{{\scriptsize\begin{tabular}{p{3cm}}
        Depth 0\\
  	 Original band\\
	 Band 72 \\
        \end{tabular}}} 
        &
  \includegraphics[width=.23\linewidth]{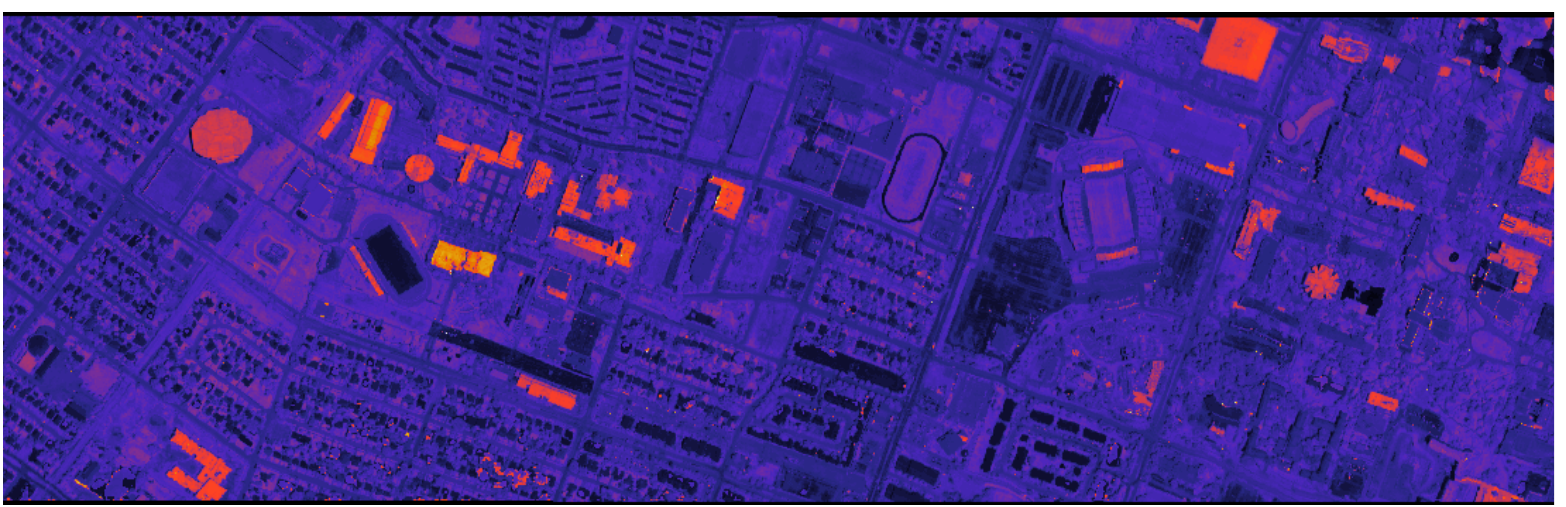} &
 \raisebox{1cm}{{\scriptsize\begin{tabular}{p{3cm}}
        Depth 0\\
  	 Original band\\
	 Band 93 \\
        \end{tabular}}} 
        \\
  $\downarrow$ & {} &  $\downarrow$ \\     

    \includegraphics[width=.23\linewidth]{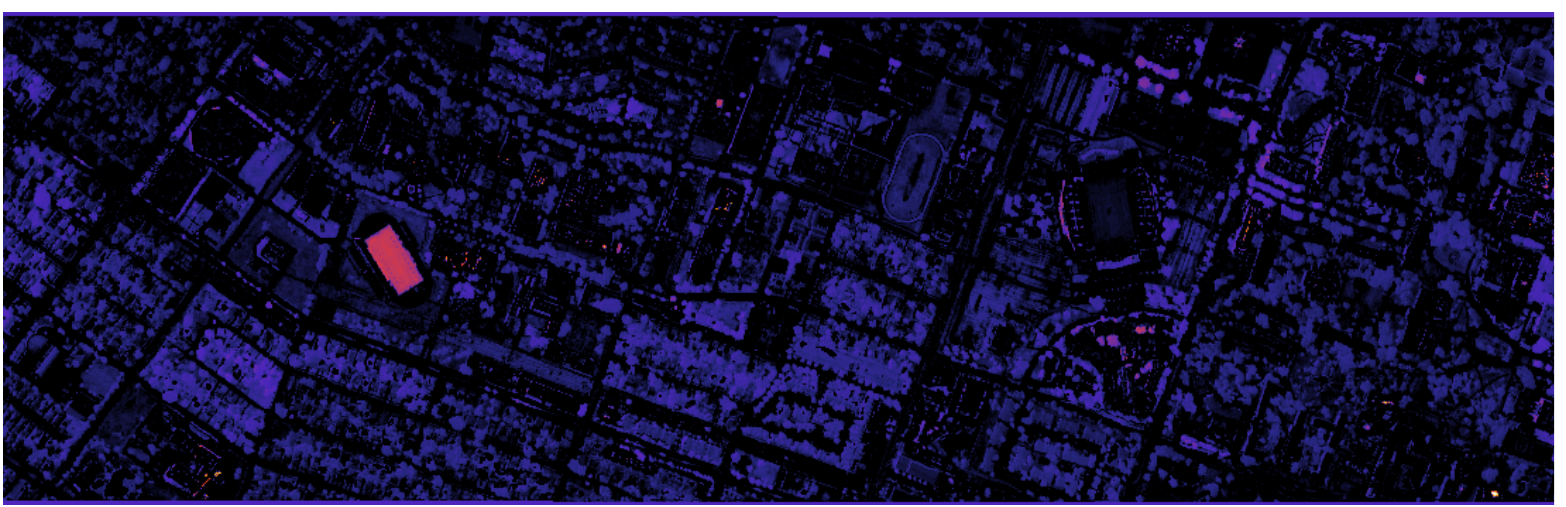} &
\raisebox{.9cm}{{\scriptsize\begin{tabular}{p{3cm}}
      	Depth 1\\
 	Closing by reconstruction top-hat\\
	Square SE, 19 pixels\\
        \end{tabular}}} 
&
  \includegraphics[width=.23\linewidth]{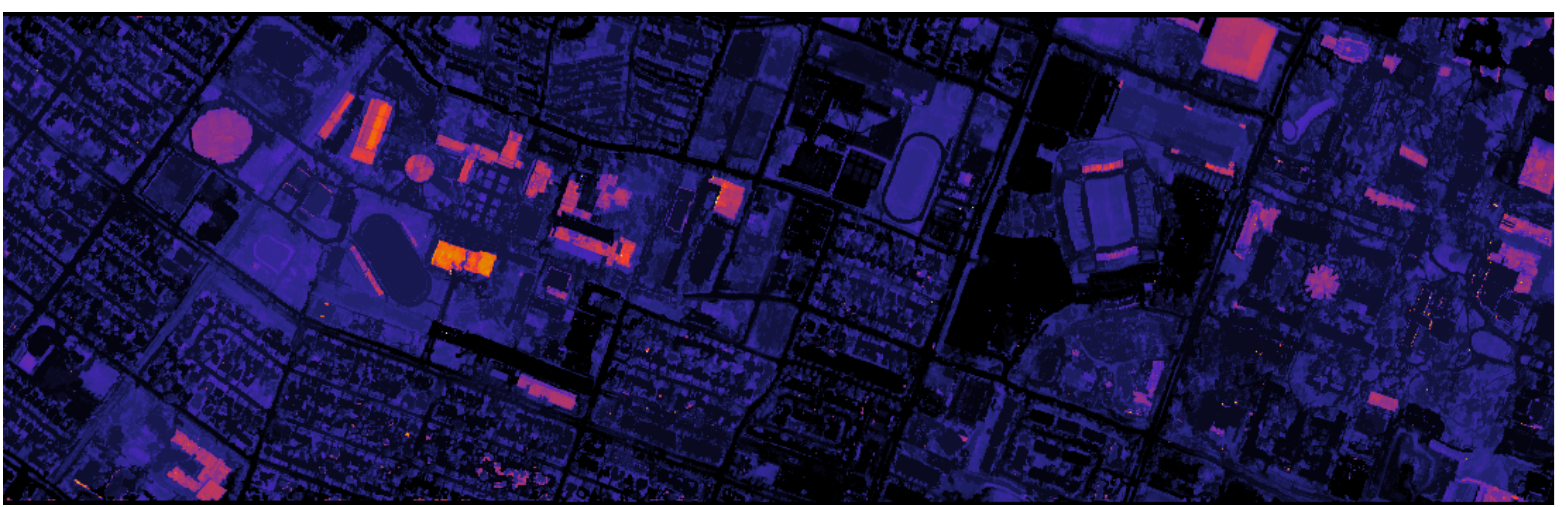} &
 \raisebox{.9cm}{{\scriptsize\begin{tabular}{p{3cm}}
      	Depth 1\\
 	Closing by reconstruction\\
	Disk SE, 15 pixels\\
        \end{tabular}}} 
        \\
  $\downarrow$ & {} &  $\downarrow$ \\      

  \includegraphics[width=.23\linewidth]{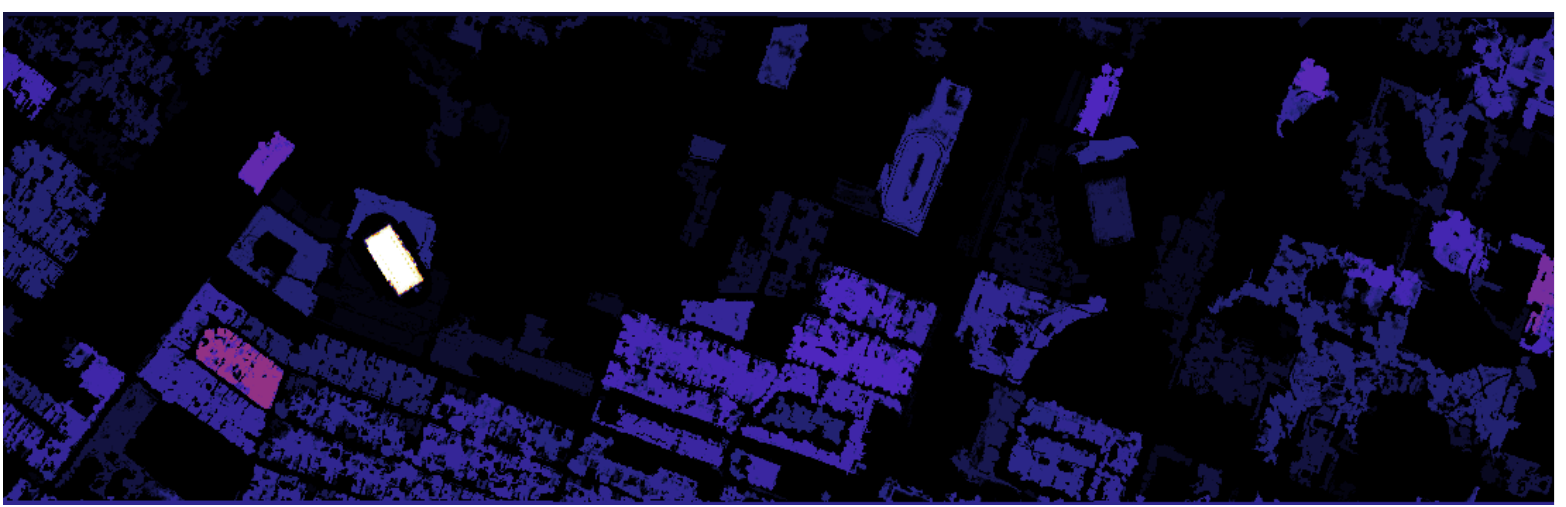} &
\raisebox{.8cm}{\fbox{{\scriptsize\begin{tabular}{p{3cm}}
      	\textbf{In the active set} \\
	Depth 2\\
 	Opening by reconstruction\\
	Square SE, 11 pixels\\
        \end{tabular}}}} 
 &
         \includegraphics[width=.23\linewidth]{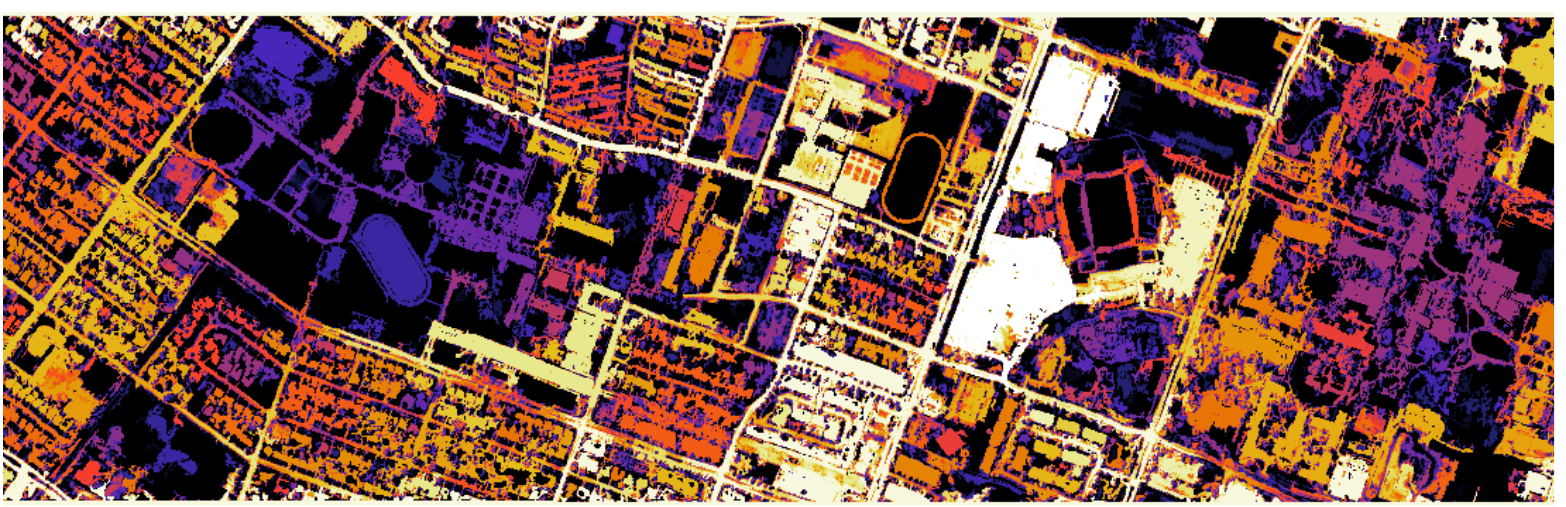} &
\raisebox{.8cm}{\fbox{{\scriptsize\begin{tabular}{p{3cm}}
        	\textbf{In the active set} \\
	Depth 2\\
 	Closing by reconstruction top-hat\\
	Disk SE, 21 pixels\\
        \end{tabular}}} }
        \\

\hline
\hline
\multicolumn{4}{c}{Two depth levels (as in the shallow setting)}\\
\includegraphics[width=.23\linewidth]{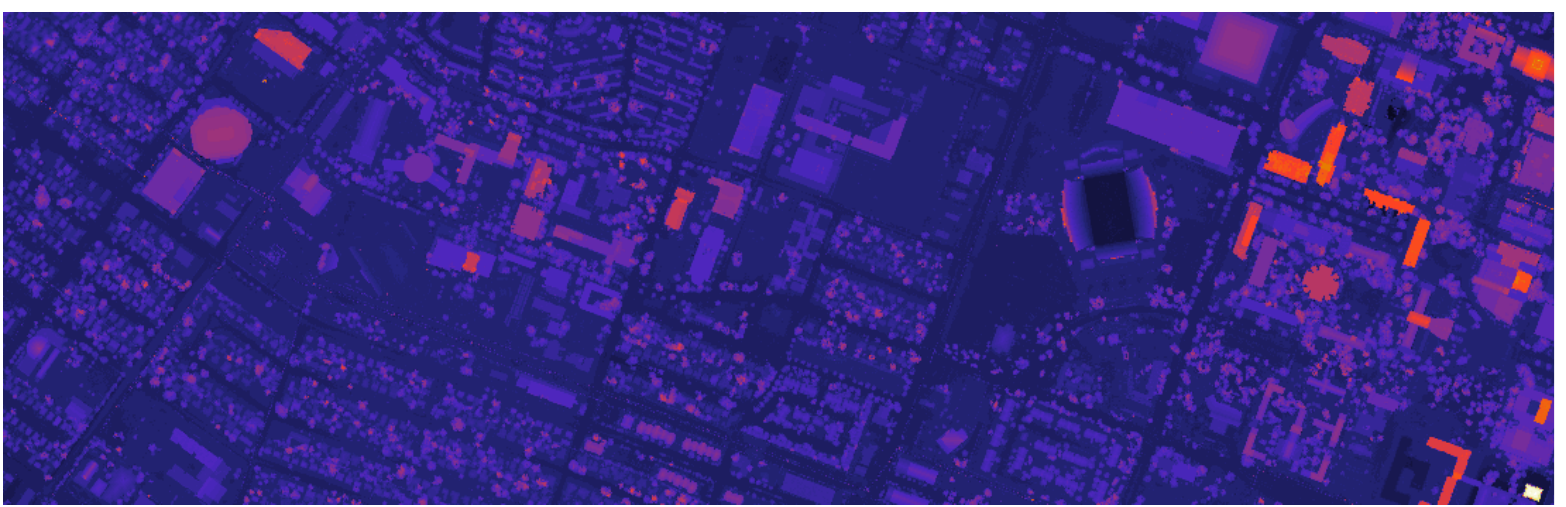} &
 \raisebox{.7cm}{{\fbox{\scriptsize\begin{tabular}{p{3cm}}
        	\textbf{In the active set}\\
	Depth 0\\
  	 Original band\\
	 Band 145 (LiDAR) \\
        \end{tabular}}} }
        &
  \includegraphics[width=.23\linewidth]{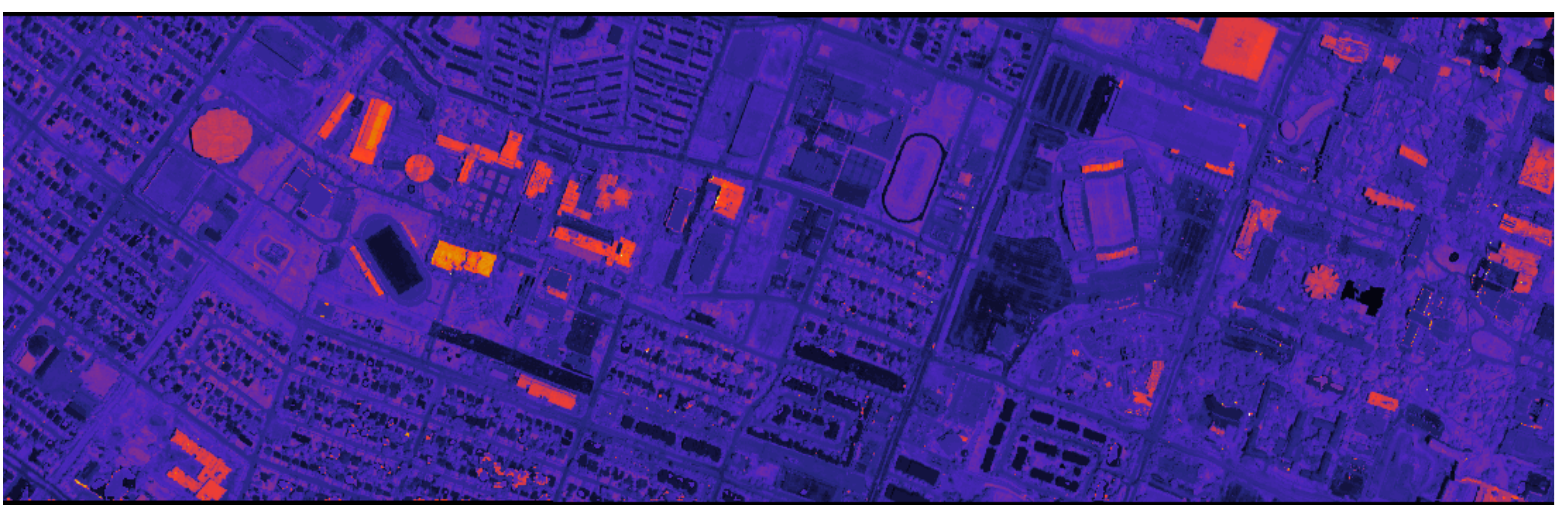} &
 \raisebox{1cm}{{\scriptsize\begin{tabular}{p{3cm}}
        Depth 0\\
  	 Original band\\
	 Band 110 \\
        \end{tabular}}} 
        \\
  $\downarrow$ & {} &  $\downarrow$ \\     

    \includegraphics[width=.23\linewidth]{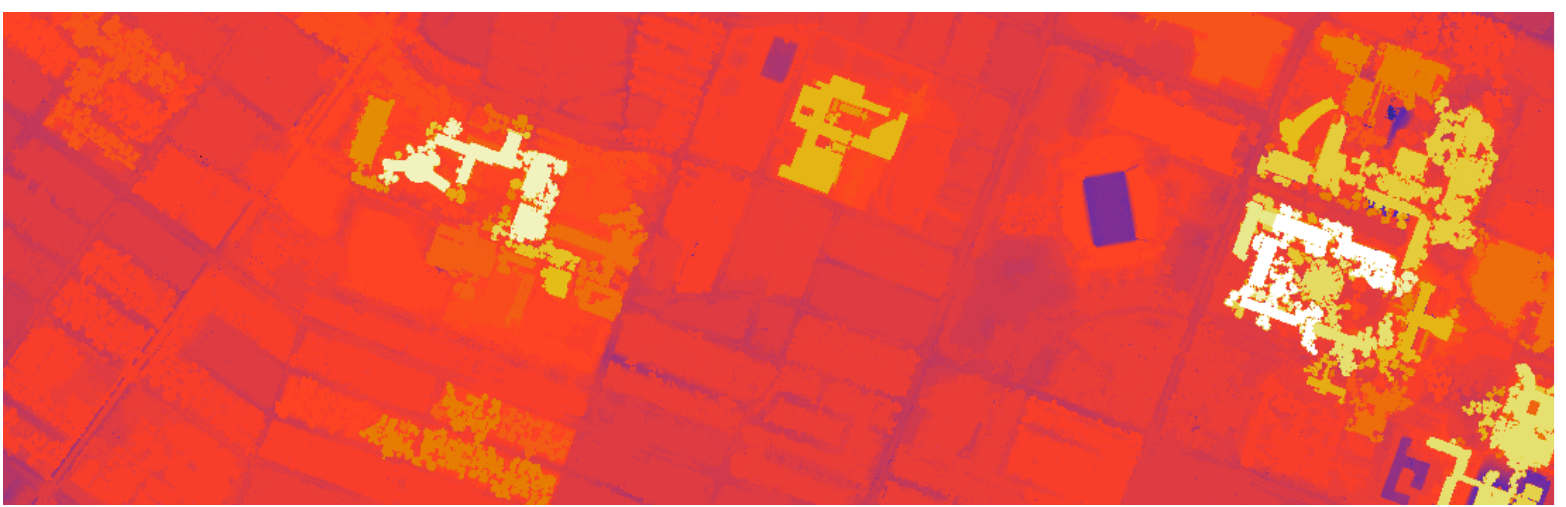} &
\raisebox{.9cm}{\fbox{{\scriptsize\begin{tabular}{p{3cm}}
 	\textbf{In the active set}\\
	Depth 1\\
 	Attribute filters - area\\
	Area of 3010 pixels\\
        \end{tabular}}} }
&
  \includegraphics[width=.23\linewidth]{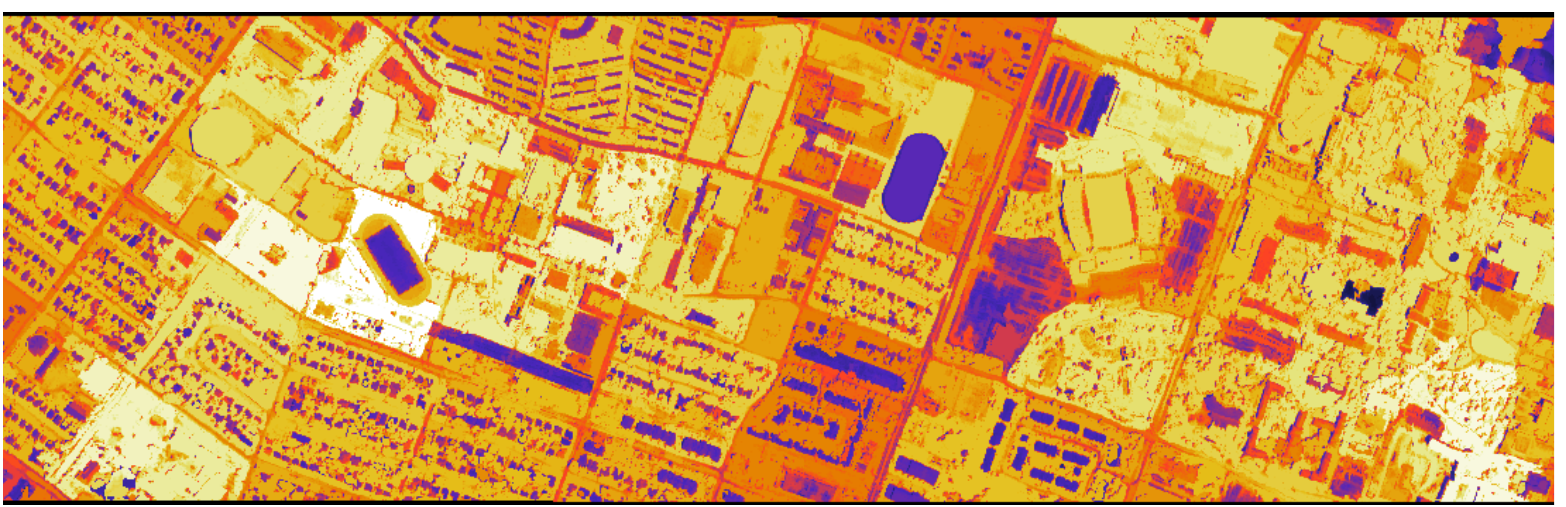} &
 \raisebox{.9cm}{\fbox{{\scriptsize\begin{tabular}{p{3cm}}
      	\textbf{In the active set}\\
	Depth 1\\
 	Attribute filters - area\\
	Area of 3010 pixels\\
        \end{tabular}}} }
        \\

\hline
\end{tabular}
\caption{Examples of the bands retrieved by the hierarchical feature learning for one specific run of the experiments on the \textsc{Houston 2013A} dataset. Highlighted are bands that are included in the final active set (after 100 iterations).}\label{fig:deep}
\end{figure*}

To confirm 
this hypothesis even further,  we study some of the features in the
final active set, illustrated in Fig.~\ref{fig:deep}: when considering
features of higher depth, we can appreciate the strong nonlinearity
induced by the hierarchical feature construction, as well as the fact
that intermediary features (the original band 105 or the features
of depth 2) are discarded from the final model, meaning that they
became uninformative during the process, but were used as basis to
generate other features that were relevant. Another interesting
behavior is the bifurcation observed in these features: the entropy
filter on band 105 was re-filtered in two different ways, and ended up
providing two very complementary, but informative filters to solve the
problem.

\section{Conclusions}\label{sec:conc}

In this paper, we proposed an active set algorithm to learn relevant
features for spatio-spectral hyperspectral image classification.
Confronted to a set of filters randomly generated from the bands of
the hyperspectral image, the algorithm selects only those that will
improve the classifier if added in the current input space. To do so,
we exploit the optimality conditions of the optimization
  problem with a regularization promoting group-sparsity. We also propose a hierarchical extension,
where active features (firstly bands and then also previously selected filters) are used as
inputs, thus allowing for the generation of more complex, nonlinear
filters. Analysis of four hyperspectral classification scenarios
confirmed the efficiency (we use a fast and linear classifier) and
effectiveness of the approach. The method is fully automatic, can
include the user favorite types of spatial or frequency filters and
can accommodate multiple co-registered data modalities.

In the future, we would like to extend the hierarchical algorithm to
situations, where a datasets shift has occurred between the training
and testing distribution: we observed that the proposed hierarchical
algorithm yields lower performances on data with spectral distortion between training and test data, as in the \textsc{Houston 2013B}
dataset. Moreover, connections to deep neural nets can be better
formalized and lead to more principled way of exploring and choosing
the features.

\section*{Acknowledgements}

This work has been supported by the Swiss National Science Foundation (grant PP00P2\_150593) and by a visiting professor grant from EPFL. We would like to thank the Image Analysis and Data fusion Technical Committee of the IEEE Geoscience and Remote Sensing Society, as well as Dr. S. Prasad, for providing the Houston data.
\section*{References}

\bibliographystyle{elsarticle-harv}
\biboptions{authoryear}

\end{document}